\documentclass{article}

    \usepackage[preprint]{neurips_2024}

\usepackage[utf8]{inputenc} 
\usepackage[T1]{fontenc}    
\usepackage{hyperref}       
\usepackage{url}            
\usepackage{booktabs}       
\usepackage{amsfonts}       
\usepackage{nicefrac}       
\usepackage{microtype}      
\usepackage{xcolor}         
\usepackage{algorithm}
\usepackage{algorithmic}
\usepackage{amsthm}
\usepackage{multirow}
\usepackage{subfigure}
\usepackage{graphicx}
\usepackage{caption}

\usepackage{amsmath,amsfonts,bm}

\def\eqref#1{equation~\ref{#1}}

\def\1{\bm{1}}

\def\eps{{\epsilon}}

\def\vzero{{\bm{0}}}
\def\vone{{\bm{1}}}

\def\vg{{\bm{g}}}

\def\vm{{\bm{m}}}

\def\vu{{\bm{u}}}
\def\vv{{\bm{v}}}
\def\vw{{\bm{w}}}

\def\mI{{\bm{I}}}

\def\mP{{\bm{P}}}

\DeclareMathAlphabet{\mathsfit}{\encodingdefault}{\sfdefault}{m}{sl}
\SetMathAlphabet{\mathsfit}{bold}{\encodingdefault}{\sfdefault}{bx}{n}

\def\gD{{\mathcal{D}}}

\def\gI{{\mathcal{I}}}

\def\gO{{\mathcal{O}}}

\def\gT{{\mathcal{T}}}

\newcommand{\E}{\mathbb{E}}
\newcommand{\Ls}{\mathcal{L}}
\newcommand{\R}{\mathbb{R}}

\theoremstyle{plain}
\newtheorem{theorem}{Theorem}[section]
\newtheorem{proposition}[theorem]{Proposition}
\newtheorem{lemma}[theorem]{Lemma}

\theoremstyle{definition}
\newtheorem{definition}[theorem]{Definition}
\newtheorem{assumption}[theorem]{Assumption}
\theoremstyle{remark}

\title{Communication-Efficient and Privacy-Preserving Decentralized Meta-Learning}

\author{%
  Hansi Yang, James T. Kwok \\
  Department of Computer Science and Engineering\\
  Hong Kong University of Science and Technology\\
  Hong Kong SAR, China \\
  \texttt{\{hyangbw, jamesk\}@cse.ust.hk} \\
}

\begin{document}

\maketitle

\begin{abstract}
Distributed learning, 
which does not require gathering training data in a central location, has become increasingly
important in the big-data era.  In particular, random-walk-based
decentralized algorithms are flexible in that they
do not need
a central server trusted by all clients and do not require
all clients to be active in all iterations.
However, existing distributed learning algorithms assume that all learning clients share
the same task.  In this paper, we consider the more difficult meta-learning
setting, in which different clients perform different (but related) tasks with
limited training data.
To reduce communication cost and allow better privacy protection,
we propose LDMeta (Local
Decentralized Meta-learning)
with the use of local auxiliary optimization parameters and random perturbations
on the model parameter.
Theoretical results 
are provided
on both
convergence 
and privacy analysis.
Empirical results on a number of 
few-shot learning data sets
demonstrate that LDMeta has similar
meta-learning accuracy as centralized meta-learning algorithms, but does not require gathering
data from each client and is able to better protect data privacy for each client. 
\end{abstract}

\section{Introduction}

Modern machine learning relies on increasingly large models 
trained on increasingly large amount of data. 
However, real-world data often come from diverse sources, 
and collecting these data to a central server can lead to large communication cost and high privacy risks.
As such, distributed learning
\cite{balcan_distributed_2012, yuan2022decentralized},
which
does not require gathering training data together, 
has received increasing attention in recent years. 
Existing methods for distributed learning can be classified as (i) 
centralized
distributed learning~\cite{predd2009collaborative, balcan_distributed_2012}, which
assumes the presence of a central server to coordinate the computation and communication for model training, 
and (ii) decentralized learning
\cite{mao_walkman_2020, lu_optimal_2021, yuan_decentlam_2021,
sun_adaptive_2022},
which does not involve a central server, 
thus is more preferable when it is hard to find a central server trusted by all clients. 
Decentralized learning
methods
can be further subdivided as: 
(i) gossip methods~\cite{koloskova_unified_2020, yuan_decentlam_2021},
which let all clients communicate with their neighbors to jointly learn models;  and
(ii) random-walk (or incremental) methods~\cite{mao_walkman_2020, sun_adaptive_2022, triastcyn_decentralized_2022},
which activate only one client in each round. 
While many works consider gossip methods, 
it requires most clients to be active during training, 
which can be difficult in practice.
For example,
in IoT applications (especially when
clients are placed in the wild), 
clients can be offline due to energy or communication issues.
In such cases, random-walk methods may be more preferable.

Most distributed learning methods assume all clients
perform the same task
and 
share a global model.
However, in many applications, different clients may have different (but related)
tasks.
For example, consider bird classification in the wild, 
different clients (camera sensors) at different locations  may target
different kinds of birds. 
On the other hand, the naive approach of training a separate model for each client
is not practical, 
as each client typically has only very limited data, 
and directly training a model can lead to bad generalization performance. 

In a centralized setting, 
meta-learning~\cite{hospedales2021survey}  has been a popular approach for
efficient learning 
of a diverse set of related tasks
with limited training data.
It has been successfully used in many applications, 
such as few-shot learning 
\cite{ravi2017meta, finn2017maml} 
and learning with label noise~\cite{shu2019mwn}.
Recently, 
meta-learning is extended to the centralized distributed setting in the context of
personalized federated learning (PFL)~\cite{marfoq_personalized_2022, pillutla_federated_2022, collins_exploiting_2021, singhal_federated_2021}. 
The central server 
updates the meta-model,
while each client 
obtains its own personalized model from the meta-model. 
However, 
PFL,
as in standard federated learning, 
still requires the use of a central server to coordinate learning. 
Some works have also considered generalizing meta-learning to decentralized settings. 
For example, Dif-MAML~\cite{Kayaalp2020DifMAMLDM} combines gossip algorithm with MAML~\cite{finn2017maml}, 
and DRML~\cite{Zhang2022DistributedRA} combines gossip algorithm with Reptile~\cite{nichol2018reptile}. 
Another example is L2C~\cite{li2022learning}, 
which also uses gossip algorithm 
and proposes to dynamically update the mixing weights 
for different clients. 
Also, methods based on decentralized bi-level optimization~\cite{NEURIPS2022_01db36a6, pmlr-v202-liu23az, pmlr-v202-chen23n, NEURIPS2023_686a3f32} 
may also be used to solve the meta-learning problem. 
Nevertheless, these works are all based on gossip algorithm, 
and share a common disadvantage that they need most clients to be always active during the learning process to achieve good performances. 
Furthermore, these methods only learn a model that can be used for all training clients, 
and the final model cannot be adapted to unseen clients that are not present during training.

Motivated by the above limitations, 
we propose a novel decentralized learning algorithm for the setting
where each client has limited data for different tasks. 
Based on random-walk decentralized optimization methods, 
the proposed method removes additional communication cost of directly using adaptive optimizers.  
We also introduce random perturbations to protect data privacy for each client. 
We prove that the proposed method
achieves the same convergence rate as existing centralized meta-learning methods, 
and provide theoretical justifications on how it can protect data privacy for each client. 
Empirical results demonstrate that the proposed method achieves similar
performances with centralized settings. 
Our contributions are listed as follows:
\begin{itemize}
    \item We propose a novel decentralized meta-learning algorithm based on random walk. 
    Compared with existing decentralized learning algorithms, 
        it has a smaller communication cost and can protect client privacy. 
    \item Theoretically, we prove that the proposed method achieves the same asymptotic convergence rate with existing decentralized learning algorithms, 
    and analyze how the perturbation variance affects privacy protection. 
    \item Extensive empirical results on various data sets and communication networks demonstrate that the proposed method can reduce the communication cost and protect client privacy, without sacrificing model performance. 
\end{itemize}

\section{Related works}





\subsection{Random-Walk Decentralized Optimization}
Given a set of 
$n$ 
clients,
random-walk (incremental) decentralized optimization algorithms~\cite{mao_walk_2019, mao_walkman_2020,
sun_adaptive_2022, triastcyn_decentralized_2022} 
aim to minimize the total loss over all clients:
\begin{align} 
\min_{\vw} \Ls(\vw) = \sum_{i=1}^n \ell(\vw, \xi_i)
\label{eq:dec}
\end{align} 
in a decentralized manner 
by performing random walk in the communication network.
Here,
$\vw$ is the model parameter,
$\xi_i$ is the training data on client $i$, and $\ell(\vw, \xi_i)$ is 
client $i$'s 
loss on 
its local data.  
In each iteration, one client is activated, 
receives the current model from the previously activated client, 
updates the model parameter with its own training data, 
and then sends the updated model to the next client. 
The active client is selected from a Markov chain with transition probability matrix $\mP 
=[\mP_{ij}]
\in \R^{n \times n}$, 
where $\mP_{ij}$
is the 
probability 
$P(i_{t+1} = j \;|\; i_t = i)$ 
that the next client $i_{t+1}$ is $j$ given that the current client
is $i$. 

The pioneering work on random-walk decentralized optimization is in
\cite{bertsekas1997}, which focuses only on the least squares problem. 
A more general algorithm is proposed 
in \cite{johansson2010}, which
uses (sub)gradient descent with Markov chain sampling. 
More recently, the Walkman algorithm~\cite{mao_walkman_2020} 
formulates problem (\ref{eq:dec}) as a linearly-constrained optimization problem, 
which is then solved by the alternating direction method of multipliers 
(ADMM)~\cite{boyd2011distributed}. 
However, these works are all based on the simple SGD for decentralized optimization. 
Very recently,
adaptive 
optimizers
are also used in random-walk decentralized optimization~\cite{sun_adaptive_2022}.
However, its communication cost is three times that of 
SGD,
as both the momentum and preconditioner
(which are of the same size as the model parameter)
need to be transmitted.
Moreover, existing works in random-walk decentralized learning assume that all clients perform the same task, 
which is not the case in many real-world applications. 


\subsection{Privacy in Distributed Learning}

Privacy is a core issue in distributed machine learning. 
Among various notations for privacy, one of the most well-known is
differential privacy (DP)~\cite{dwork2014algorithmic}.
The idea is to add noise to the model updates
so that the algorithm output does not reveal sensitive information about any individual data sample.
Although it is originally proposed for centralized machine learning algorithms~\cite{mcmahan2018dp}, 
DP has also found wide applications in centralized
distributed learning, 
particularly the federated learning setting
where a central server coordinates model training 
on distributed data sources without data ever leaving each client. 
An example is FedDP~\cite{wei2020federated}, 
where DP is directly combined with the FedAvg algorithm~\cite{mcmahan2017communication}.
Later, 
\cite{hu2020personalized} 
generalizes DP
to the personalized federated learning, 
where different clients have non-i.i.d. training data. 

There have been limited progress on privacy in decentralized learning without a central server. 
One prominent work is~\cite{cyffers2022privacy}, 
which considers random-walk algorithms on rings and fully-connected graphs,
but not 
communication networks with diverse topological structures
as is often encountered in the real world.
Another 
decentralized learning algorithm with privacy guarantees
is
Muffliato
\cite{cyffers2022muffliato},
which is based on gossip 
methods but not 
random walk.
Moreover, both cannot be used for decentralized meta-learning,
in which different clients perform different tasks. 

\begin{algorithm}[ht]
\caption{Adaptive Random Walk Optimizer.} \label{alg:adam}
\begin{algorithmic}[1]
\STATE {\bfseries Input:} hyper-parameters $\eta > 0, 0 \le \theta < 1, \lambda >0$. 
	\STATE initialize $\vm_{-1} = \vzero, \vv_{-1} = \vzero$ for all client $i$ and set the first client $i_0$;
   \FOR{$t=0$ {\bfseries to} $T-1$}
   
   \STATE initialize $\vu_0 = \vw_t$;
   
   \COMMENT{$K$ steps of SGD for base learner}
   \FOR{$k=0$ {\bfseries to} $K-1$}
   \STATE compute $\vg_{k} = \nabla \ell(\vu_k;\xi^s_{i_t})$ with support data $\xi^s_{i_t}$ of client $i_t$;
   \STATE update $\vu_{k+1} = \vu_k - \alpha \vg_k$;
\ENDFOR

\COMMENT{Update by meta learner}
\STATE compute $\vg_{t} = \nabla_{\vw_t} \ell(\vu_K;\xi^q_{i_t})$ with query data $\xi^q_{i_t}$ of client $i_t$;
\STATE $\vm_t = \theta \vm_{t-1} + (1-\theta)\vg_t$;
   \STATE $\vv_t = \beta \vv_{t-1} + (1-\beta) [\vg_t]^2$;
   \STATE $\vw_{t+1} = \vw_t - \eta \frac{\vm_t}{(\vv_t+\lambda \vone)^{1/2}}$;
   \STATE Select next client $i_{t+1}$ from the Markov chain with transition probability matrix $\mP 
=[\mP_{ij}]
\in \R^{n \times n}$. 
    \STATE transmit $(\vw_{t+1}, \vm_t, \vv_t)$ to next client $i_{t+1}$;
   \ENDFOR
   \STATE transmit final model $\vw_T$ to unseen clients
\end{algorithmic}
\end{algorithm}

\begin{algorithm}[ht]
\caption{LoDMeta: Local
Decentralized Meta-learning.} \label{alg:meta}
\begin{algorithmic}[1]
\STATE {\bfseries Input:} hyper-parameters $\eta > 0, 0 \le \theta < 1, 0 \le \beta <1, \lambda >0, 0<\eps<1, 0<\delta < 1/2$. 
	\STATE initialize $\vm^i_{-1} = \vzero, \vv^i_{-1} = \vzero$ for all client $i$ and set the first client $i_0$;
   \FOR{$t=0$ {\bfseries to} $T-1$}
   \STATE initialize $\vu_0 = \vw_t$;
   
   \COMMENT{$K$ steps of SGD for base learner}
   \FOR{$k=0$ {\bfseries to} $K-1$}
   \STATE compute $\vg_{k} = \nabla \ell(\vu_k;\xi^s_{i_t})$ with support data $\xi^s_{i_t}$ of client $i_t$;
   \STATE update $\vu_{k+1} = \vu_k - \alpha \vg_k$;
\ENDFOR

\COMMENT{Update by meta learner}
\STATE compute $\vg_{t} = \nabla_{\vw_t} \ell(\vu_K;\xi^q_{i_t})$ with query data $\xi^q_{i_t}$ of client $i_t$;
\STATE $\vm^{i_t}_t = \theta \vm^{i_t}_{t-1} + (1-\theta)\vg_t$;
   \STATE $\vv^{i_t}_t = \beta \vv^{i_t}_{t-1} + (1-\beta) [\vg_t]^2$;
   \STATE generate Gaussian perturbation $\bm{\epsilon}_t$ where each element has variance $\sigma^2=\frac{8 M_{meta}^2 \ln(1.25/\delta)}{\eps^2}$;
   \STATE $\vw_{t+1} = \vw_t - \eta \frac{\vm^{i_t}_t+\bm{\epsilon}_t}{(\vv^{i_t}_t+\lambda \vone)^{1/2}}$;
   \STATE select next client $i_{t+1}$ from the Markov chain with transition probability matrix $\mP 
=[\mP_{ij}]
\in \R^{n \times n}$;
    \STATE transmit $\vw_{t+1}$ to next client $i_{t+1}$;
   \ENDFOR
   \STATE transmit final model $\vw_T$ to unseen clients
\end{algorithmic}
\end{algorithm}

\section{Proposed Method}




\subsection{Problem Formulation}

Following the formulation in (\ref{eq:dec}), 
we consider the setting where each client has its own task, 
and new clients may join the network with limited data. 
We propose to use meta-learning~\cite{hospedales2021survey} to jointly learn from different tasks. 
Denote the set of all tasks (which also corresponds to all clients) as $\gI$, 
we have the following bi-level optimization problem:
\begin{align*}
& \min_{\vw} & \!\!\!\! \Ls(\vw) = 
\frac{1}{| \gI |}\sum_{i \in \gI} L((\vw, \vu^i(\vw));
\gD^i_{\text{vald}}), \text{s.t. }
\vu^i \equiv
\vu^i(\vw) = \arg\min_{\vu} L((\vw, \vu); \gD^i_{\text{tr}}), \forall i
\end{align*}
where 
$\vw$ is the 
meta-parameter shared by all tasks,
$\vu^i(\vw)$ is the parameter specific to task $i$,  and
$\gD^i_{\text{tr}}$
(resp. 
$\gD^i_{\text{vald}}$)
is 
task $i$'s
meta-training  or support
(resp. 
meta-validation or query)
data.
$L((\vw, \vu); \gD)
= \E_{\xi \sim \gD}[\ell((\vw, \vu); \xi)]$
is the loss 
of task $i$'s model on 
data $\gD$,
 where $\ell((\vw, \vu); \xi)$ is the loss on a
stochastic sample $\xi$.
As in most works on meta-learning~\cite{finn2017maml, nichol2018reptile, zhou2022task},
we use the meta-parameter as meta-initialization.
This can be used on both training clients and unseen clients, 
as any new client $i$ can simply use the learned meta-parameter to initialize its model $\vu^i$.  
The outer loop  finds  a suitable
meta-initialization $\vw$,
while the inner loop adapts it
to each client $i$ as $\vu^i(\vw)$. 
An example algorithm for such adaptation is shown in Algorithm~\ref{alg:adp} of Appendix~\ref{app:alg}.

While existing works on random-walk decentralized optimization~\cite{sun_adaptive_2022,triastcyn_decentralized_2022} 
can also be easily extended
to the meta-learning setting
(an example is shown in Algorithm~\ref{alg:adam}), 
they often have high communication cost
as the
adaptive optimizer's
auxiliary parameters (momentum $\vm_t$ and pre-conditioner $\vv_t$) 
need to be passed
to the next client. 
Moreover, sending more auxiliary parameters can possibly lead to high privacy risk, 
as adversarial clients have more information to attack. 



\subsection{Reducing Communication Cost }


Since the high communication cost and privacy leakage both come from sending auxiliary parameters to the other clients, 
we propose to use \textit{localized} auxiliary parameters for each client. 
Specifically, 
the meta-learner  of
each client $i$ keeps its own momentum 
$\vm^{i}_t$ 
and pre-conditioner
$\vv^{i}_t$. They
are no longer sent to the next client, and only the model parameter needs to be
transmitted.
The proposed algorithm, called LoDMeta
(Local
Decentralized Meta-learning), 
is shown 
in Algorithm~\ref{alg:meta}.
At step~2, 
we initialize the local auxiliary parameters $\vm^i_{-1}, \vv^i_{-1}$ for each client $i$. 
During learning, 
each client then uses its local auxiliary parameters $\vm^i_{t}$ and $ \vv^i_{t}$. 
Without the need to transmit auxiliary parameters, 
its communication cost is reduced to only one-third of that in 
Algorithm~\ref{alg:adam}.
Moreover, as will be shown theoretically in the next section, 
Algorithm~\ref{alg:meta} can achieve the same asymptotic convergence rate as 
Algorithm~\ref{alg:adam} 
even only with localized auxiliary parameters. 

While LoDMeta in
Algorithm~\ref{alg:meta}
is based on the MAML algorithm and Adam optimizer, it can be easily used with other meta-learning algorithms (e.g., ANIL~\cite{Raghu2020Rapid} or BMG~\cite{flennerhag_bootstrapped_2022}) 
by simply replacing the update step with steps in the corresponding meta-learning algorithm. 
Similarly, LoDMeta can also be easily used with 
    other adaptive optimizers 
    that need transmission of auxiliary parameters 
    (e.g., AdaGrad~\cite{duchi2011adaptive}, AdaBelief~\cite{zhuang2020adabelief} and Adai~\cite{pmlr-v162-xie22d}) by again replacing 
    the global auxiliary parameters 
    with local copies.

\subsection{Protecting Privacy}
Sharing the model parameter 
can still incur privacy leakage. 
For privacy protection, 
we propose to add random Gaussian perturbations to the model
parameters~\cite{dwork2014algorithmic, cyffers2022privacy}. 
There have been works on privacy-preserving adaptive optimizers~\cite{li2022private, li2023differentially}. 
While they achieve remarkable performance under the centralized setting, 
they cannot be directly generalized to the decentralized setting. 
For example, AdaDPS~\cite{li2022private} requires additional side information (e.g., public training data without privacy concerns) to estimate the momentum or preconditioner, 
which is hard to obtain in practice even in the centralized setting. 
DP$^2$-RMSprop~\cite{li2023differentially} requires accumulating gradients across different clients.
This needs additional communication and computation
in the decentralized setting.

In contrast, as in Algorithm~\ref{alg:meta},
the proposed method protects privacy by first removing communication of the auxiliary parameters.
We then only need to add random perturbations to the model parameters, 
which is the only source of privacy leakage.

%

\section{Theoretical Analysis}
\label{sec:ana}


\subsection{Analysis on Convergence Rate and Communication Cost}

Denote the total communication cost as $C$, which can be expressed by $C = C_T T$, 
where $C_T$ denotes the per-iteration communication cost and $T$ denotes the number of iterations. 
Then to compare the total communication cost for different methods, we need to consider their per-iteration communication costs and the total number of iterations. 
For comparison fairness, we consider the relative per-iteration communication cost, 
which can neglect other affecting factors such as model size and parameter compression techniques. 
We take the per-iteration communication cost of LoDMeta as 1 unit, 
as the active client only sends model parameters to another client. 
LoDMeta(basic) then requires three
times the communication cost of LoDMeta in each iteration, 
as it needs to also transmit momentum and preconditioner 
to the next client. 
Centralized methods (i.e., MAML and FedAlt)
require twice the communication cost for each active client, 
as each client requires downloading and uploading the current meta-parameter to the central server. 

\begin{table}[ht]
\vspace{-10px}
\caption{Relative per-iteration communication costs for the various methods.}
\label{tab:comm}
\begin{center}
\begin{tabular}{c c c}
\toprule
MAML/FedAlt 
& L2C/LoDMeta(basic) & LoDMeta (SGD)/LoDMeta \\
(Centralized, $n$ denotes number of active clients) &   (Decentralized) &  (Decentralized) \\
\midrule
2$n$ & 3 & 1 \\ 
\bottomrule
\end{tabular}
\vspace{-10px}
\end{center}
\end{table}

Then we compare the number of iterations by deriving the convergence rate for LoDMeta. 
Under the meta-learning setting, the objective in (\ref{eq:dec}) takes the following form: 
\begin{align}
\min_{\vw} \Ls(\vw) = 
\frac{1}{n}\sum_{i=1}^n \ell(\vu^i_K(\vw);\xi^q_i), \label{eq:meta}
\end{align}
where $\vu^i_k(\vw)$, the local model parameter for client $i$ computed from $\vw$, is computed in the inner loop of
Algorithm~\ref{alg:meta}:
\begin{align*}
\vu^i_0(\vw) = \vw, \; \vu^i_{k+1}(\vw) = \vu^i_k(\vw) - \alpha \nabla \ell(\vu^i_k(\vw); \xi^s_i).
\end{align*}
The meta-gradient 
for client $i$ is then computed as~\cite{ji2020maml}:
$G_i(\vw) =\prod_{k=0}^{K-1}(\mI - \alpha \nabla^2 \ell(\vu_k(\vw); \xi^s_i))\nabla
\ell(\vu_K(\vw); \xi^q_i)$,
where
 $\mI$ is the identity matrix. 

We make the following assumptions, which are commonly used in the convergence
analysis of
meta-learning~\cite{pmlr-v108-fallah20a, ji2020maml, pmlr-v162-yang22g} 
and random-walk decentralized optimization~\cite{sun_adaptive_2022,triastcyn_decentralized_2022}.
\begin{assumption}\label{assum:smoothoff}
For data $\xi$, the loss $\ell(\cdot; \xi)$ satisfies: (i) 
{\bf bounded loss}: $ \inf_{\vw}
\ell(\vw; \xi) > -\infty$; 
(ii) {\bf Lipschitz gradient}: 
$\|\nabla \ell(\vu; \xi) -\nabla \ell(\vw; \xi) \| \leq M\|\vu - \vw \|$
For any $\vu, \vw$;
(iii) {\bf Lipschitz Hessian}: 
$\|\nabla^2 \ell(\vu; \xi) -\nabla^2 \ell(\vw; \xi) \|_{sp} \leq \rho \| \vu - \vw \|$
for any $\vu, \vw$, 
where $\| \cdot\|_{sp}$ is the spectral norm;
(iv) {\bf bounded
gradient variance}: 
For any $\vw$, $\mathbb{E}_{i}\|\nabla \ell(\vw; \xi_i^q) -\mathbb{E}_{i} [
\ell(\vw; \xi_i^q)] \|^2 \leq \sigma^2$; 
(v) {\bf bounded differences for support/query data}: for each $i\in \mathcal{I}$, there exists a constant
		$b_i>0$ such that $\|\nabla \ell(\vw; \xi^s_i) - \nabla \ell(\vw; \xi^q_i)\|
		\leq b_i$ for any $\vw$. 
\end{assumption}

The following Proposition shows that the expected meta-gradient 
$\nabla \mathcal{L}(\vw) = \frac{1}{n} \sum_{i=1}^n G_i(\vw)$
is also Lipschitz.
This is useful in analyzing the convergence 
and privacy properties
of Algorithm~\ref{alg:meta}. 
\begin{proposition}
\label{prop:lip} 
For any $\vu, \vw \in\mathbb{R}^d$, we have 
	$\|\nabla \mathcal{L}(\vu) - \nabla \mathcal{L}(\vw)\| \leq M_{meta}\|\vu -
	\vw\|$,
	where
$M_{meta} = (1+\alpha M)^{2K}M + C (b + \mathbb{E}_{i}\|\nabla
\ell(\vw; \xi^q_i)\|)$,
$b= \frac{1}{n} \sum_{i=1}^n b_i$ and $C = \big( \alpha \rho + \frac{\rho}{M}  (1+\alpha M)^{K-1} \big)
(1+\alpha M)^{2K}$. 
\end{proposition}
\begin{theorem}
\label{thm:conv} 
Set 
the inner- and outer-loop
learning rates in 
Algorithm~\ref{alg:meta} 
to
$\alpha = \frac{1}{8KM}$, 
	and $\eta = \frac{1}{80 M_{meta}}$, respectively. 
For any 
$\eps>0$,
with $T = O\left( \max \{ \frac{n}{\eps^{2} [\log(1/\sigma_2(\mP))]^{2}},
\frac{n}{\eps^{2}} \} \right)$, 
where $\sigma_2(\mP)$ is the second largest eigenvalue of the transition probability matrix $\mP$, 
we have
$\min_{0 \le t \le T} \E \| \nabla \Ls(\vw_t) \|^2 = O(\eps)$.
\end{theorem}
Proof is in Appendix~\ref{app:lip}, 
where we need to make different bounds with local auxiliary parameters. 
Compared with the convergence of MAML in centralized setting~\cite{ji2020maml},
Theorem~\ref{thm:conv} has the same dependency on $\epsilon$.
This also agrees with previous work on random-walk
algorithms~\cite{sun_adaptive_2022,triastcyn_decentralized_2022},
though their analysis requires auxiliary parameters to be synchronized across all clients, 
while Algorithm~\ref{alg:meta} uses localized ones. 
The impact of communication network is reflected by the $\log(1/\sigma_2(P))$, 
which also matches previous analysis on random-walk algorithms~\cite{sun_adaptive_2022,triastcyn_decentralized_2022}. 
Then since LoDMeta has the same convergence rate (same number of iterations $T$) 
but significantly smaller per-iteration communication cost $C_T$ (as in Table~\ref{tab:comm}), 
it has much smaller communication cost than existing methods. 

\subsection{Privacy Analysis}

Let the (private) data on client $i$ be $D_i$, 
and the union of all client data be $D = \cup_{i=1}^n D_i$. 
For two such unions $D$ and $D'$, 
we use $D \sim_i D'$ to indicate that $D$ and $D'$ have the same 
number of clients
and differ only on client $i$'s data, 
which defines a \emph{neighboring relation} over these unions. 
Following existing works on privacy in decentralized algorithms~\cite{cyffers2022privacy}, 
we consider any decentralized algorithm $A$ 
as a (randomized) mapping that takes the union of client data $D$ as input
and 
outputs all messages exchanged between two clients
over the network. 
We denote all these messages as
$A(D) = \{ (i,m,j) : \text{ user } i \text{ sent message with
content } m \text{ to user } j \}$.
A key difference between centralized and decentralized algorithms is that in the decentralized setting, 
a given client does not~have access to 
all messages in $A(D)$, 
but only to the messages it 
is involved in. 
As such, to analyze the privacy property of a decentralized algorithm, 
we need to consider separate view of each client. 
Mathematically, we denote 
client $i$'s 
view 
of algorithm $A$ 
as:
$\mathcal{O}_{i}(A(D))
=\{ (i,m,j)\in A(D), j \in \gI\} 
\cup \{ (j,m,i)\in A(D), j \in \gI \}$.

\begin{definition}[Network Differential Privacy~\cite{cyffers2022privacy}]
A decentralized algorithm $A$ satisfies $(\epsilon, \delta)$-network DP if for all pairs of distinct clients $i, j$ and all neighboring unions of data $D \sim_{i} D'$, we have:
$P(\gO_j(A(D))) \le \exp(\epsilon) P(\gO_j(A(D'))) + \delta$.
\end{definition}
In other words, network DP requires that for any two users $i$ and $j$, 
the information gathered by user $j$ from algorithm $A$ 
should not depend too much on user $i$’s data. 
Under this definition, we can now prove 


\begin{theorem}
\label{thm:iteration}
Let $\eps<1$, $\delta < 1/2$.
Suppose $\eta\leq 2/M_{meta}$, 
and $\bm{\epsilon}_t$ is generated from the normal distribution with variance $\sigma^2=\frac{8 M_{meta}^2 \ln(1.25/\delta)}{\eps^2}$
in Algorithm~\ref{alg:meta}, 
then Algorithm~\ref{alg:meta} achieves $
(\eps',\delta+ \hat{\delta})$-network
DP for all $\hat{\delta}>0$ with
\begin{equation}
\label{eq:completegd_bigO}
\eps' = \sqrt{2q \ln (1/\delta)}\eps/\sqrt{\ln(1.25/\delta)},
\end{equation}
where
$q = \max \big(2 N_u, 2 \ln(1/\delta) \big)$ and $N_u = \frac{T}{n} + \sqrt{
\frac{3}{n} T \ln (1/\hat{\delta})}$.
\end{theorem}

Algorithm~\ref{alg:meta} have similar
dependencies on $\epsilon$ and $\delta$ 
as in
\cite{cyffers2022privacy}. 
As $\epsilon'$ is proportional to $\epsilon$,
a smaller $\epsilon$ leads to better protection of privacy.
Recall
that a smaller $\epsilon$ leads to a larger perturbation $\bm{\epsilon}_t$ in Algorithm~\ref{alg:meta} (step 12).
Thus, a larger perturbation leads to better privacy protection, 
which agrees with our intuition. 
Compared with \cite{cyffers2022privacy}, 
our analysis is applicable to
networks of any topology, 
while the analysis in
\cite{cyffers2022privacy}
is only applicable to rings and fully-connected networks.
Moreover, 
\cite{cyffers2022privacy}
only considers learning a single specific task 
(namely, mean estimation or stochastic
gradient descent on convex objectives), while
we consider the more sophisticated and general meta-learning setting.

The recent work MetaNSGD~\cite{zhou2022task} also considers
private meta-learning. However,
we consider a decentralized setting while MetaNSGD assumes all the data to be stored in a centralized server. 
Moreover, MetaNSGD assumes that the loss for each task/client 
is convex
(which does not hold for deep networks),
while our analysis does not require such strong assumption.

\section{Experiments}

\subsection{Setup}
\label{ssec:set}

{\bf Datasets}.
We conduct experiments on few-shot learning 
using two 
standard benchmark data
sets:
(i) \textit{mini-ImageNet},  
which is a coarse-grained image classification data set 
popularly used in meta-learning~\cite{finn2017maml, nichol2018reptile}; 
(ii) \textit{Meta-Dataset}~\cite{Triantafillou2020Meta-Dataset:},
which is a collection of 
fine-grained image classification data sets. 
As in~\cite{yao2019hsml},
we use four data sets in \textit{Meta-Dataset}: (i) \textit{Bird}, (ii) \textit{Texture}, (iii) \textit{Aircraft}, and (iv) \textit{Fungi}.
We consider two few-shot settings: 
5-way 1-shot and 5-way 5-shot. 
\footnote{$N$-way $K$-shot refers to doing classification with $N$ classes, 
and each client has $K$ samples for each class $KN$ samples in total. }
The number of query samples is always set to 15. 
Following 
standard practice in meta-learning
\cite{finn2017maml, yao2019hsml}, 
some classes are used for meta-training, while the rest is for meta-testing. 

\noindent
{\bf Baselines}.
Our proposed method
LoDMeta 
is compared
with the following baselines:
(i) two popular methods from personalized federated learning, 
including MAML under the federated learning setting~\cite{fallah2020maml} 
and FedAlt~\cite{marfoq_personalized_2022}
(ii) L2C~\cite{li2022learning}, 
which is the only known decentralized meta-learning algorithm and uses
the gossip algorithm instead of the random-walk algorithm,
and (iii) the basic MAML extension 
to decentralized learning
in Algorithm~\ref{alg:adam}, denoted as LoDMeta(SGD) and LoDMeta(basic). 
Both of them do not perform communication cost reduction.
For MAML, since its communication cost depends on the number of clients, 
we consider two settings: 
the original setting~\cite{finn2017maml} where it samples 4 clients in each iteration, 
referred as MAML, 
and another setting where it only samples 1 client to reduce communication cost, 
referred as MAML (1 client). 

\noindent
{\bf Communication network}.
For centralized methods (MAML and FedAlt), 
the communication network is essentially a star,
with the server at the center.
For decentralized methods (L2C, LoDMeta(basic) and LoDMeta),  we use two networks: 
the popular Watts-Strogatz small-world network
\cite{watts1998collective}, and
the 3-regular expander network, in which each client has 3 neighbors. 
The number of clients for each data set is 
in Table~\ref{tab:stats} in Appendix~\ref{app:dex}. 

The 
clients in the
network are divided into two types:
(i) training clients, with
data coming from the meta-training classes; and
(ii) unseen
clients, which join the network after meta-training. Their
data are from the meta-testing classes,
and they use the trained meta-model for adaptation.


\begin{table}[ht]
\vspace{-10px}
\caption{Testing accuracies (in percentage) on training clients with different $\epsilon$'s and $\delta$'s.}
\label{tab:prt}
\begin{center}
\begin{tabular}{c c c c c c c}
\toprule
& $\delta=0.4$  & $\delta=0.3$  & $\delta=0.2$ & $\delta=0.1$ & $\delta=0.05$ & $\delta=0.02$ \\ \midrule
$\epsilon=0.8$ & \textbf{49.8} & \textbf{49.8} & 49.7         & 49.4         & 48.2          & 47.3          \\
$\epsilon=0.7$ & 49.7          & 49.7          & 49.6         & 49.3         & 47.6          & 46.4          \\
$\epsilon=0.6$ & 49.7          & 49.6          & 49.5         & 49.1         & 47.1          & 45.2          \\
$\epsilon=0.5$ & 49.6          & 49.6          & 49.4         & 48.8         & 46.7          & 44.6          \\
\bottomrule
\end{tabular}
\vspace{-.2in}
\end{center}
\end{table}


\begin{table}[ht]
\vspace{-10px}
\caption{Testing accuracies (in percentage) on unseen clients with different $\epsilon$'s and $\delta$'s.}
\label{tab:pru}
\begin{center}
\begin{tabular}{c c c c c c c}
\toprule
               & $\delta=0.4$  & $\delta=0.3$  & $\delta=0.2$ & $\delta=0.1$ & $\delta=0.05$ & $\delta=0.02$ \\ 
              \midrule
$\epsilon=0.8$ & \textbf{48.0} & \textbf{48.0} & 47.8         & 47.2         & 46.6          & 45.9          \\
$\epsilon=0.7$ & \textbf{48.0} & \textbf{48.0} & 47.6         & 47.0         & 45.5          & 44.6          \\
$\epsilon=0.6$ & \textbf{48.0} & 47.9          & 47.4         & 46.7         & 44.7          & 43.2          \\
$\epsilon=0.5$ & \textbf{48.0} & 47.9          & 47.1         & 46.3         & 43.8          & 42.3          \\
\bottomrule
\end{tabular}
\vspace{-.2in}
\end{center}
\end{table}

\begin{figure}[h]
	\centering
	\subfigure[1-shot. Small-world network.]
	{\includegraphics[width=0.23\columnwidth]{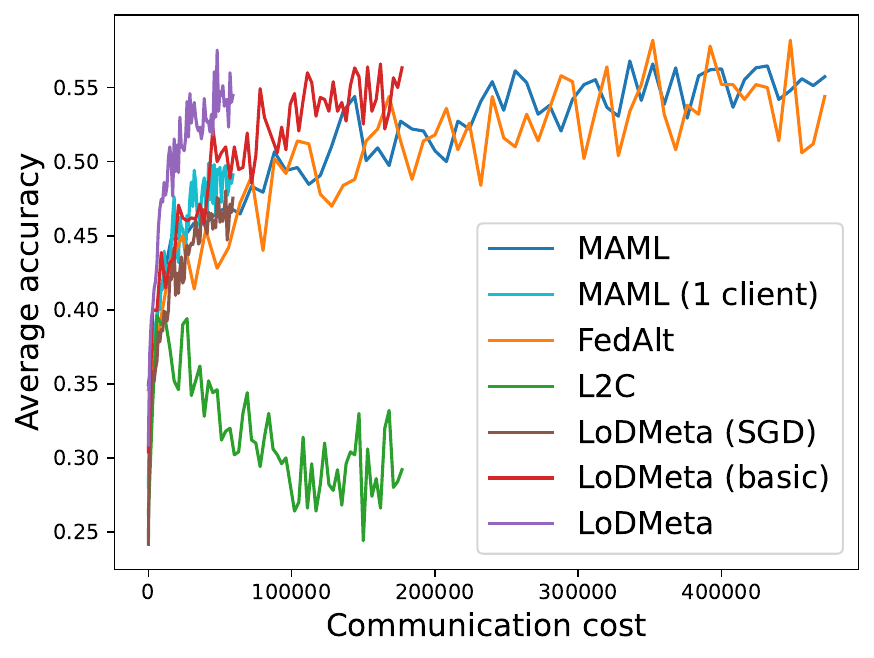}}	
	\subfigure[5-shot. Small-world network.]
	{\includegraphics[width=0.23\columnwidth]{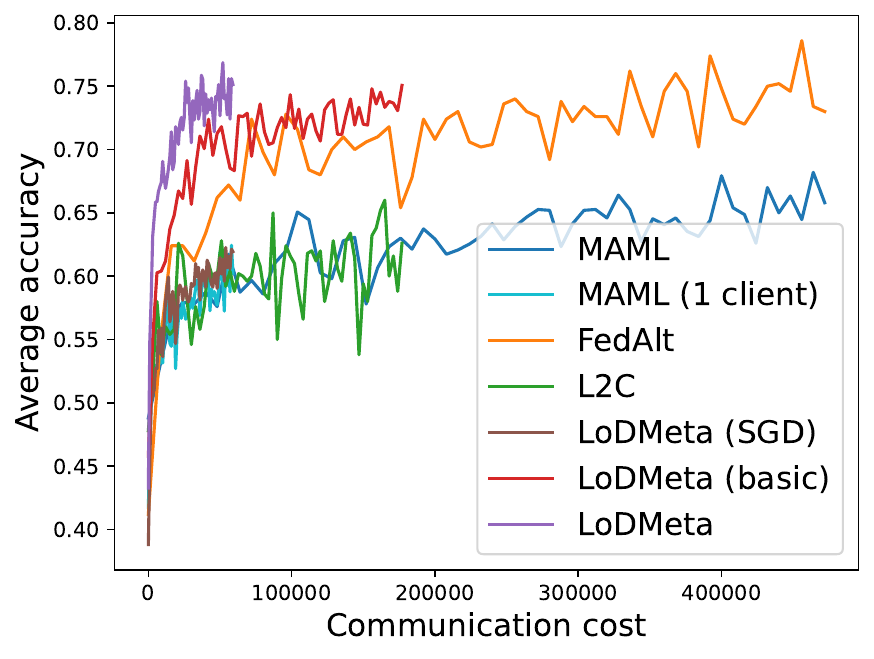}}
 \subfigure[1-shot. 3-regular expander network.]
	{\includegraphics[width=0.23\textwidth]{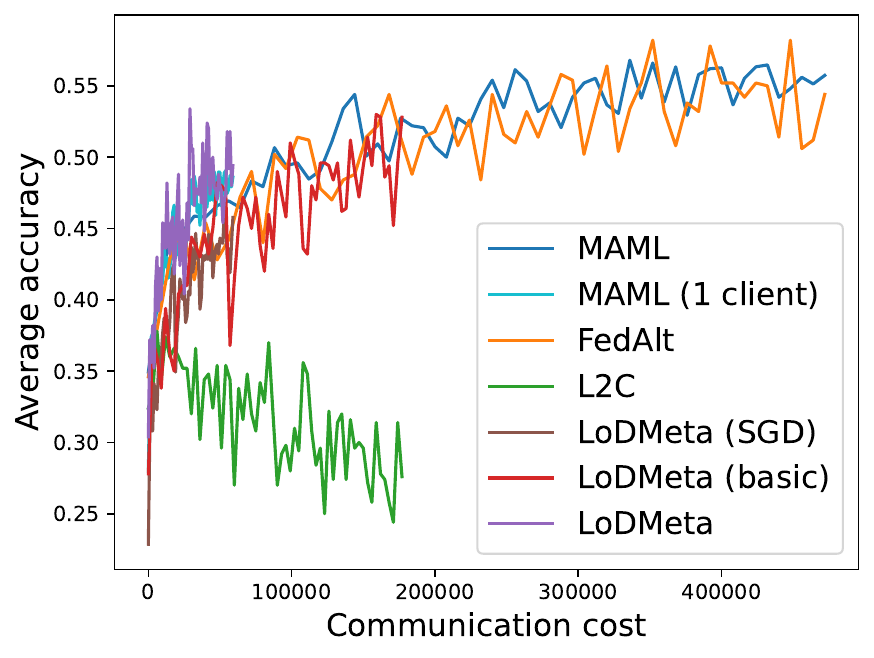}}	
	\subfigure[5-shot. 3-regular expander network.]
	{\includegraphics[width=0.23\textwidth]{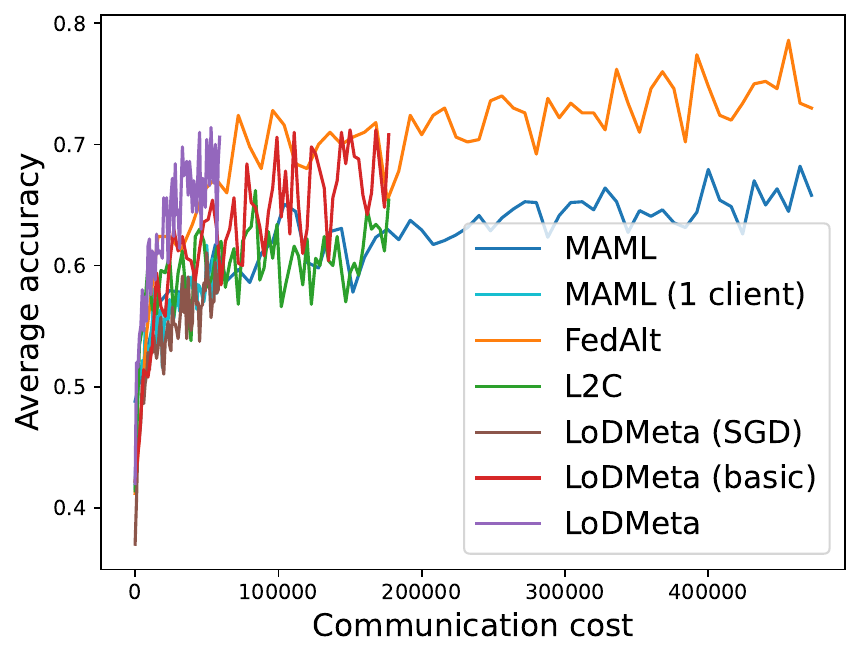}}
\vspace{-.1in}	
	\caption{Average testing accuracies for training clients on \textit{mini-ImageNet}. 
\label{fig:mini_1}}
\vspace{-10px}
\end{figure}

\begin{figure}[h]
	\centering
	\subfigure[1-shot. Small-world network.]
	{\includegraphics[width=0.23\columnwidth]{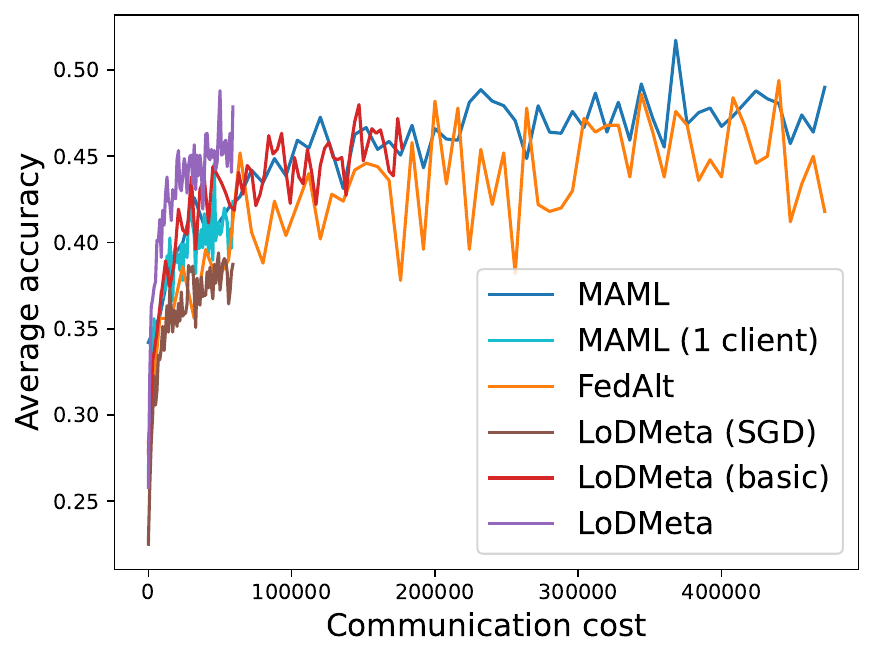}}
	\subfigure[5-shot. Small-world network.]
	{\includegraphics[width=0.23\columnwidth]{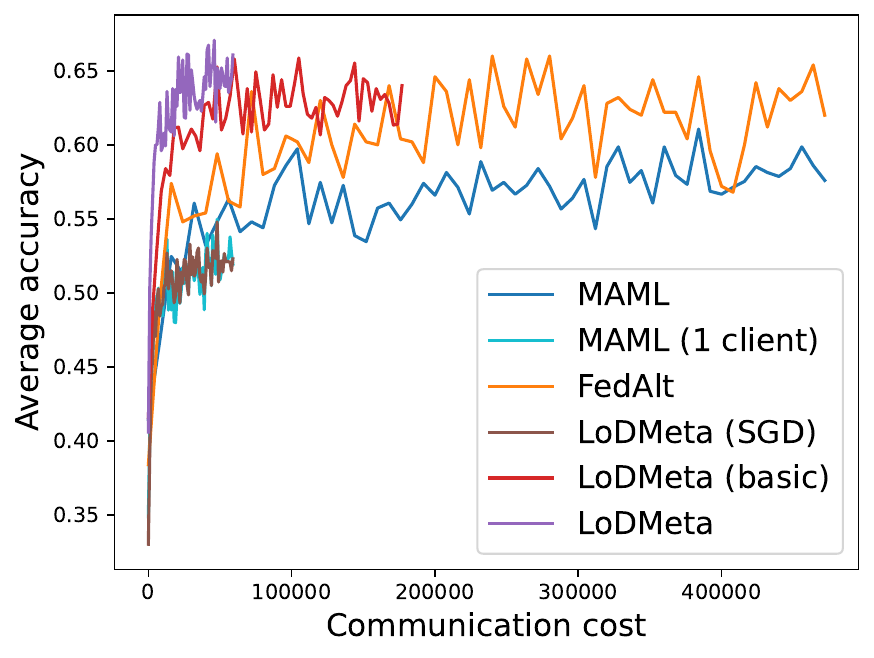}}
	\subfigure[1-shot. 3-regular expander network.]
	{\includegraphics[width=0.23\textwidth]{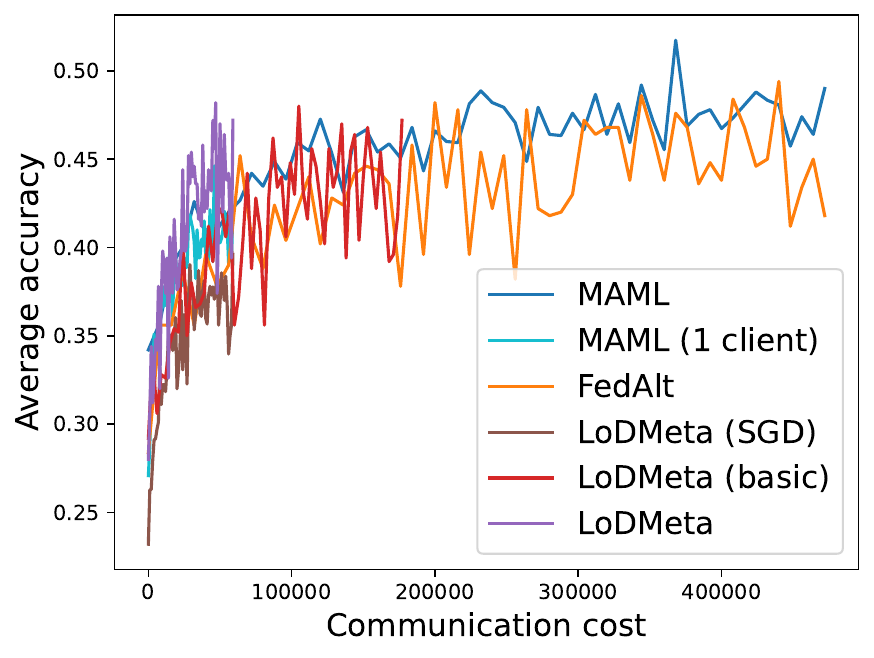}}	
	\subfigure[5-shot. 3-regular expander network.]
	{\includegraphics[width=0.23\textwidth]{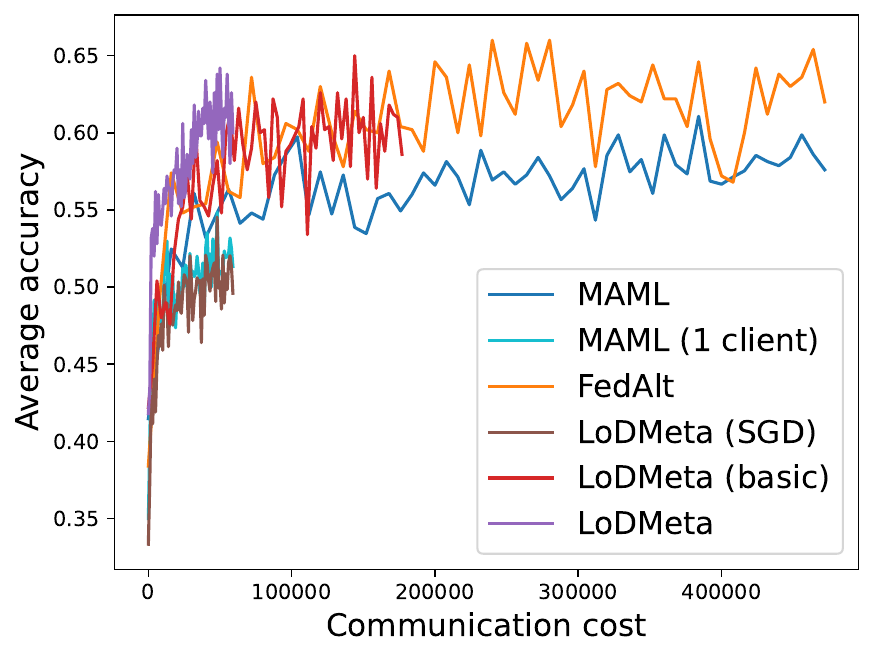}}		
\vspace{-.1in}	
	\caption{Average testing accuracies for unseen clients on \textit{mini-ImageNet}. 
\label{fig:rmini_1}}
\vspace{-10px}
\end{figure}

\begin{figure*}[h]
	\centering
	\subfigure[\textit{Bird}. Small-world network.]
	{\includegraphics[width=0.24\textwidth]{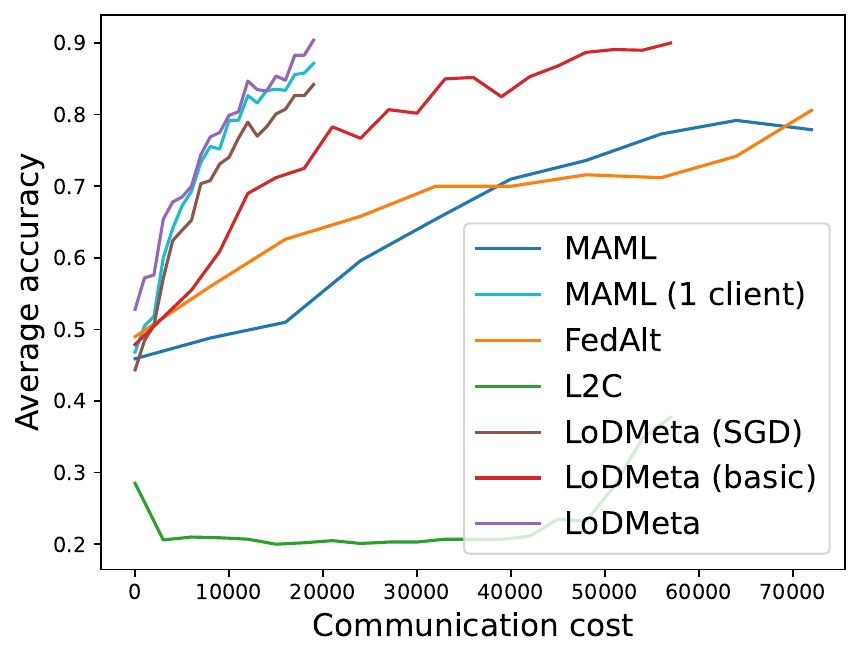}}	
	\subfigure[\textit{Texture}. Small-world network.]
	{\includegraphics[width=0.24\textwidth]{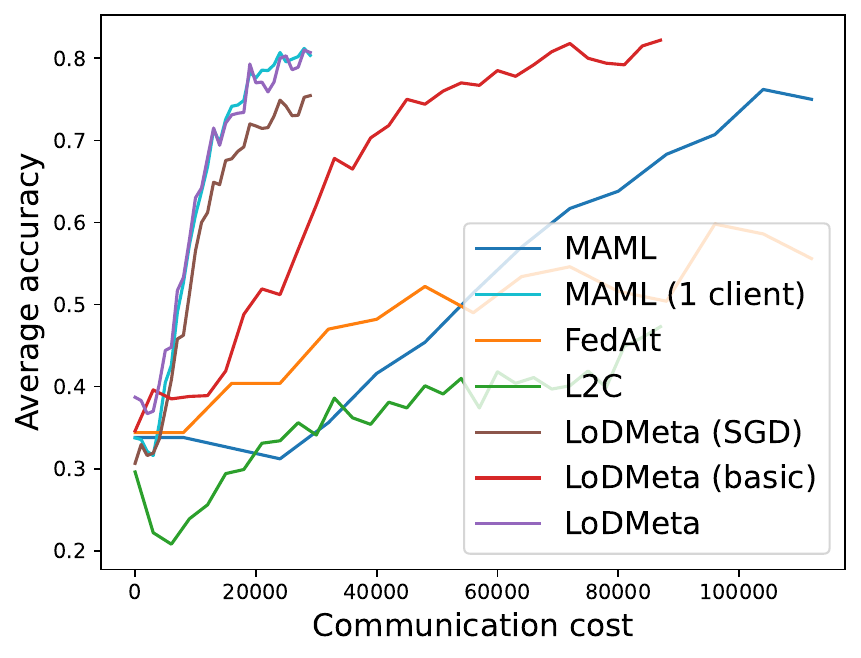}}	
	\subfigure[\textit{Aircraft}. Small-world network.]
	{\includegraphics[width=0.24\textwidth]{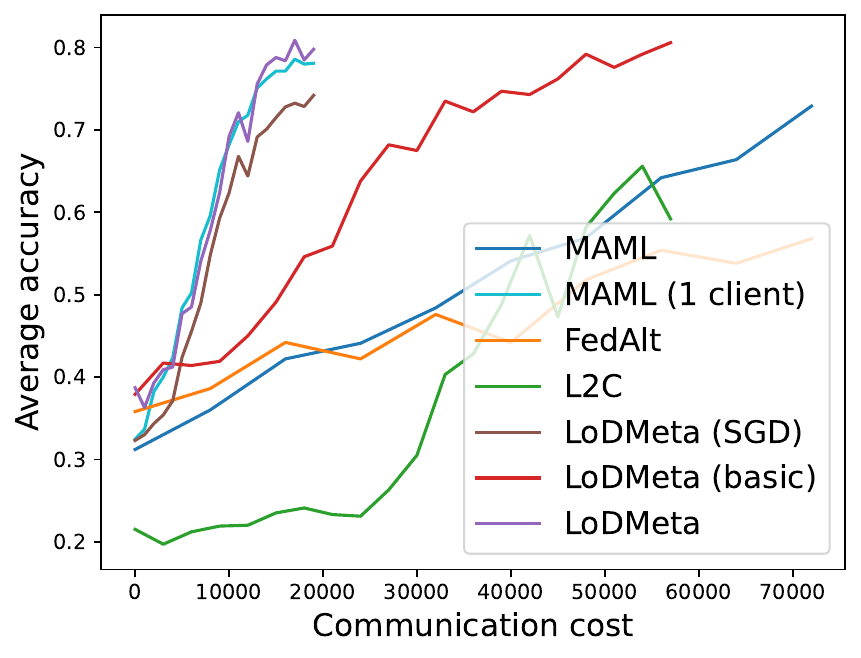}}	
	\subfigure[\textit{Fungi}. Small-world network.]
	{\includegraphics[width=0.24\textwidth]{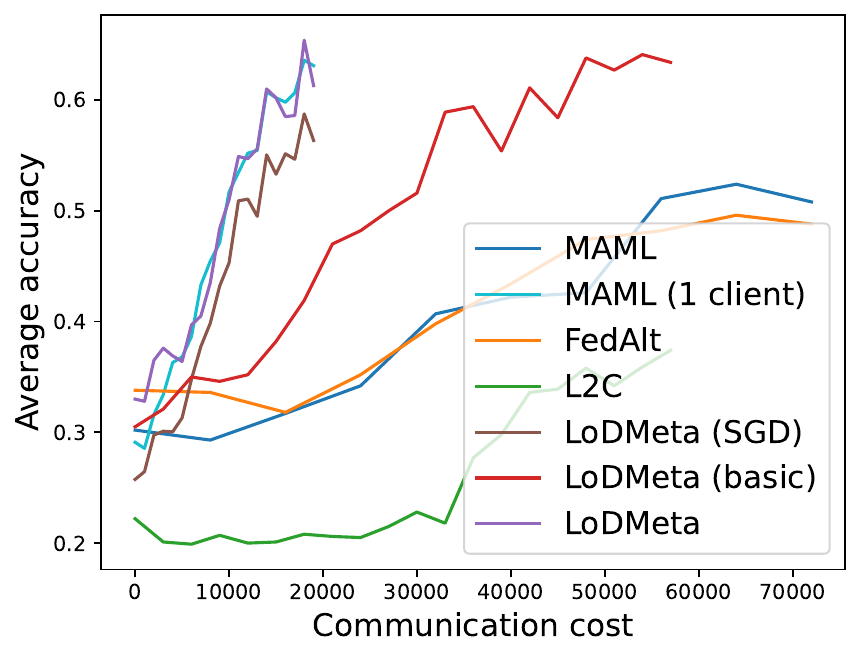}}
 \subfigure[\textit{Bird}. 3-regular expander network.]
	{\includegraphics[width=0.24\textwidth]{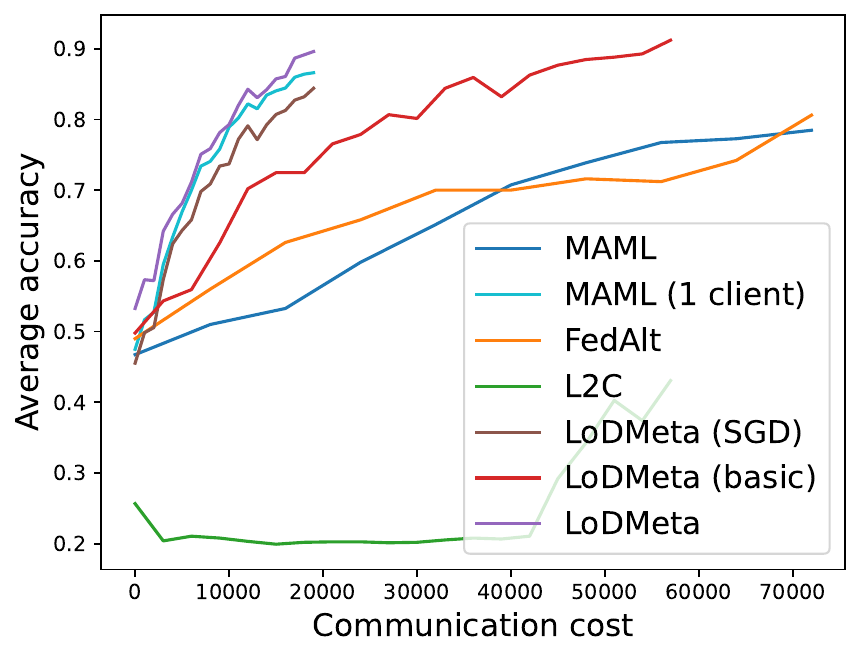}}	
	\subfigure[\textit{Texture}. 3-regular expander network.]
	{\includegraphics[width=0.24\textwidth]{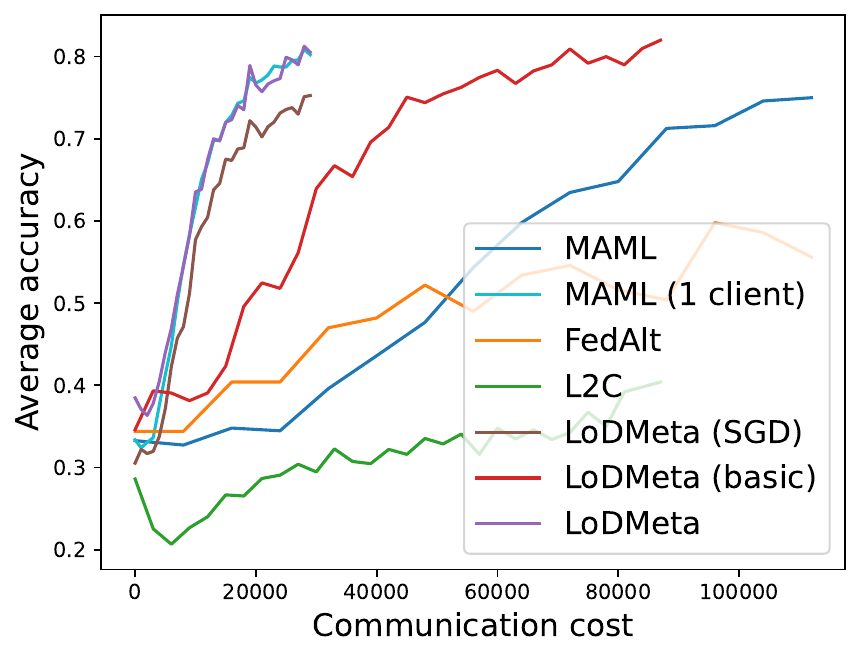}}	
	\subfigure[\textit{Aircraft}. 3-regular expander network.]
	{\includegraphics[width=0.24\textwidth]{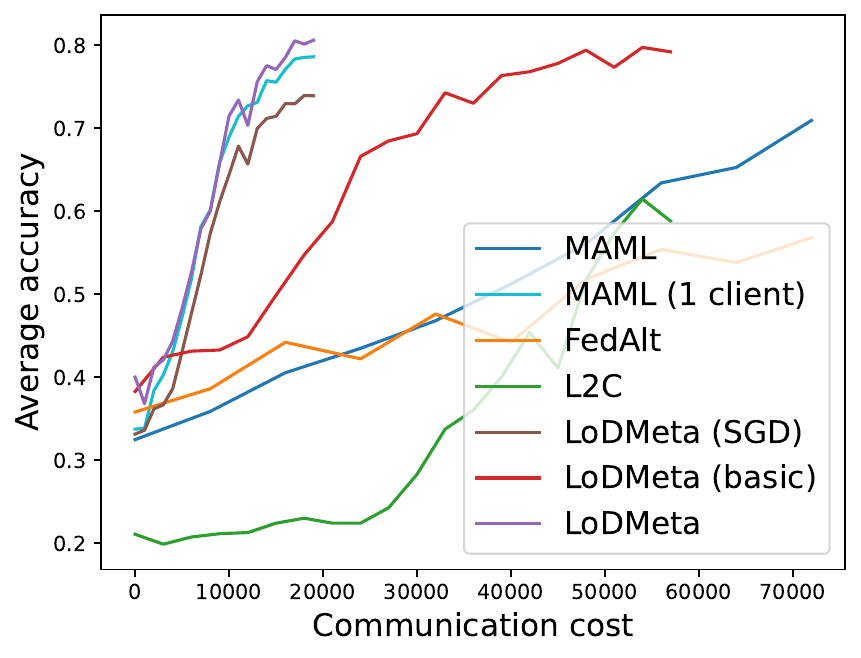}}	
	\subfigure[\textit{Fungi}. 3-regular expander network.]
	{\includegraphics[width=0.24\textwidth]{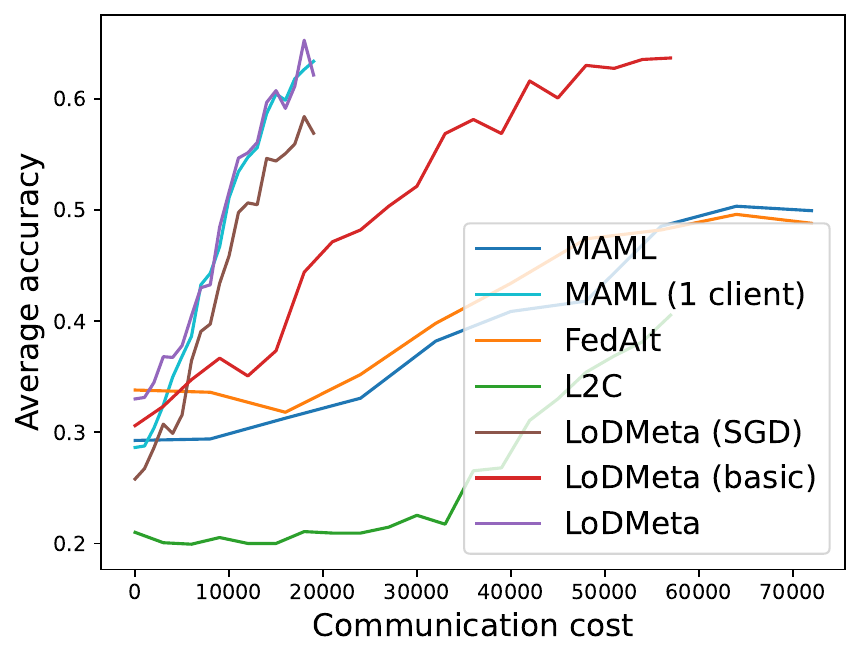}}
\vspace{-.1in}	
	\caption{Average testing accuracy with communication cost for training clients on \textit{Meta-Datasets} under 5-shot setting. 
	\label{fig:meta_5_comm}}
	\vspace{-10px}
\end{figure*}

\begin{figure*}[h]
	\centering
	\subfigure[\textit{Bird}. Small-world network.]
	{\includegraphics[width=0.24\textwidth]{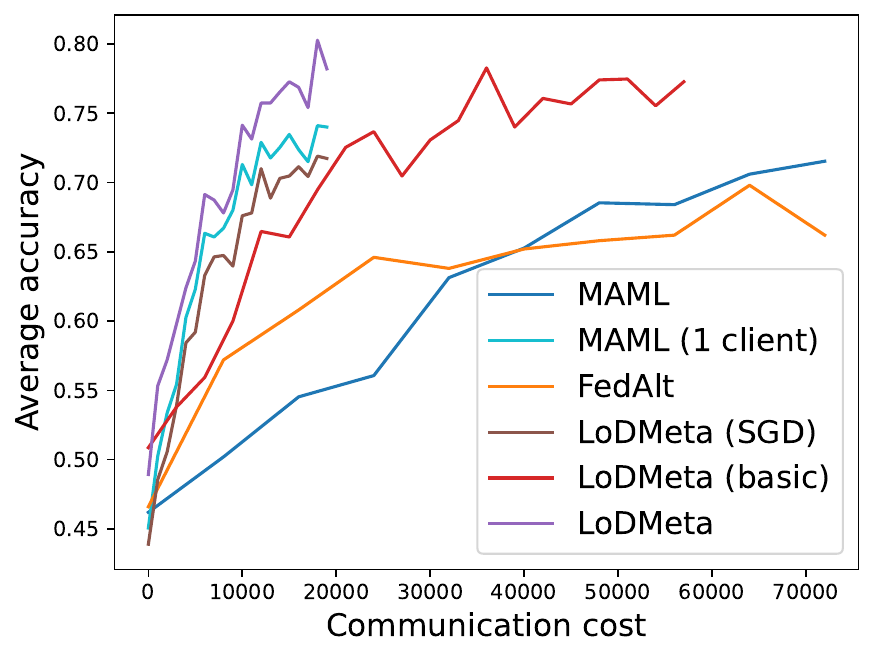}}	
	\subfigure[\textit{Texture}. Small-world network.]
	{\includegraphics[width=0.24\textwidth]{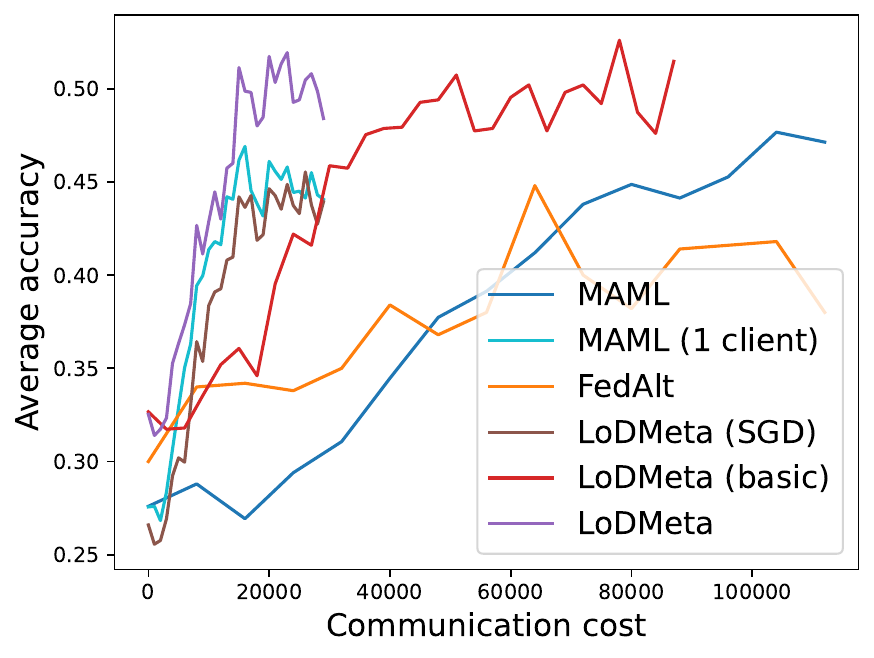}}	
	\subfigure[\textit{Aircraft}. Small-world network.]
	{\includegraphics[width=0.24\textwidth]{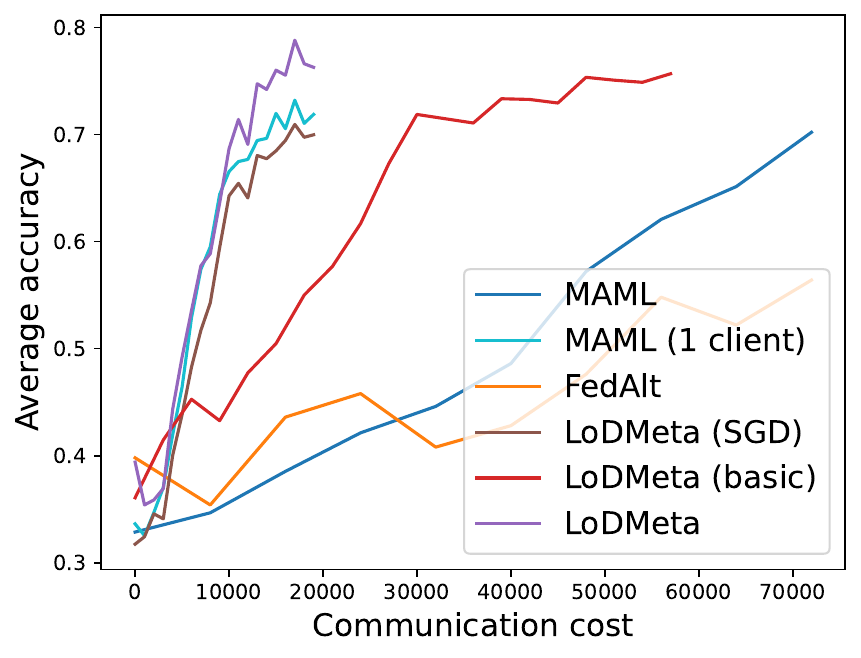}}	
	\subfigure[\textit{Fungi}. Small-world network.]
	{\includegraphics[width=0.24\textwidth]{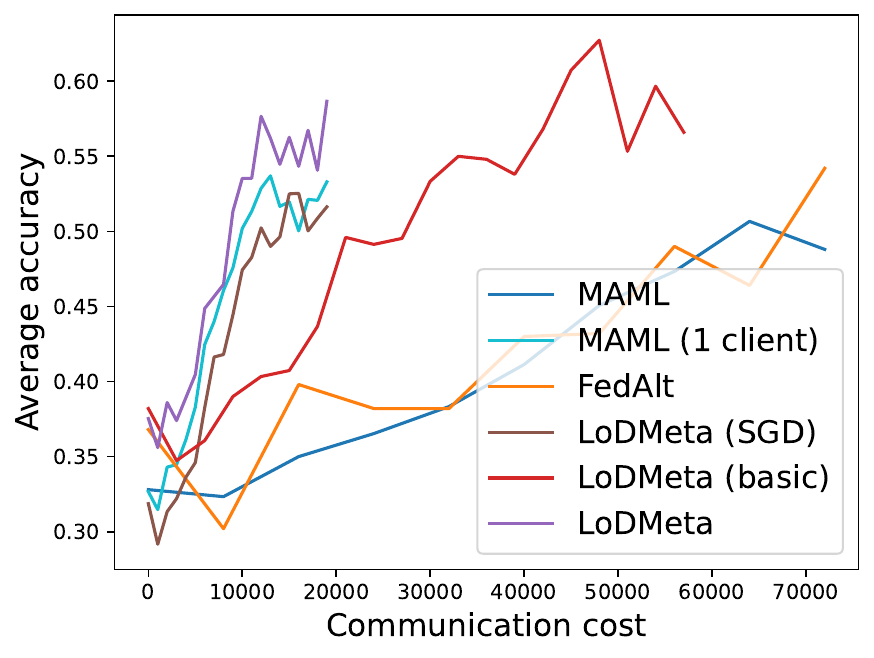}}
 \subfigure[\textit{Bird}. 3-regular expander network.]
	{\includegraphics[width=0.24\textwidth]{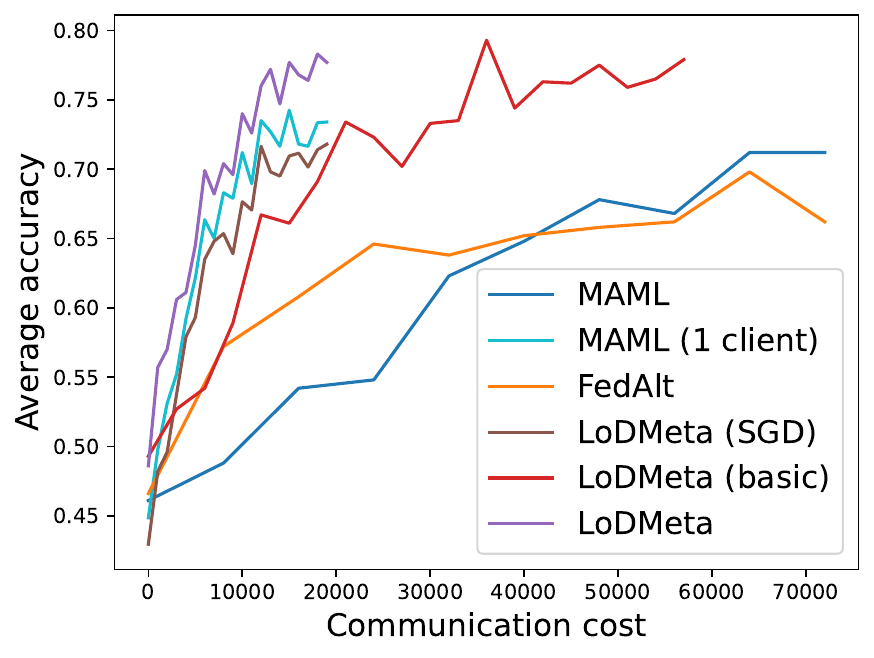}}	
	\subfigure[\textit{Texture}. 3-regular expander network.]
	{\includegraphics[width=0.24\textwidth]{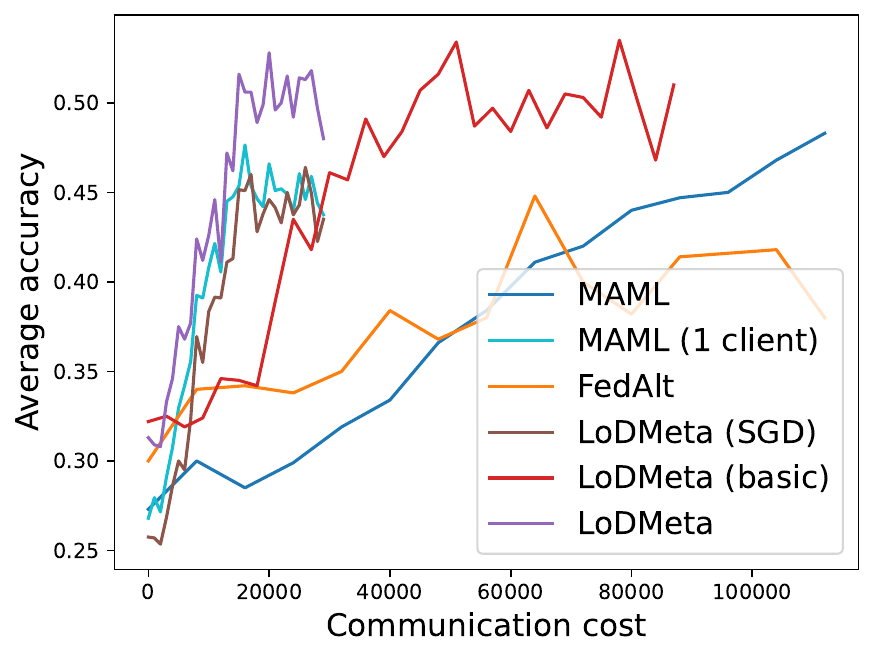}}	
	\subfigure[\textit{Aircraft}. 3-regular expander network.]
	{\includegraphics[width=0.24\textwidth]{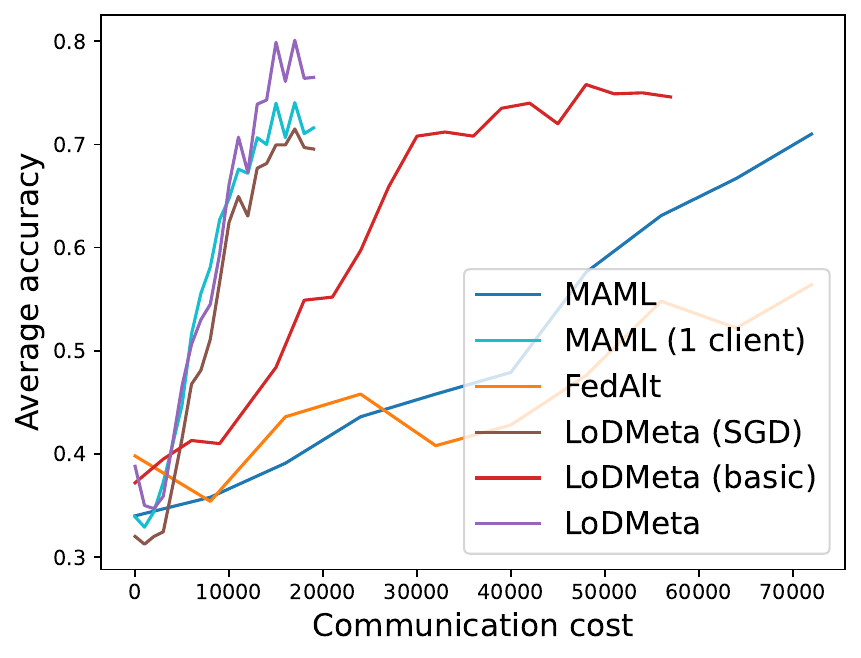}}	
	\subfigure[\textit{Fungi}. 3-regular expander network.]
	{\includegraphics[width=0.24\textwidth]{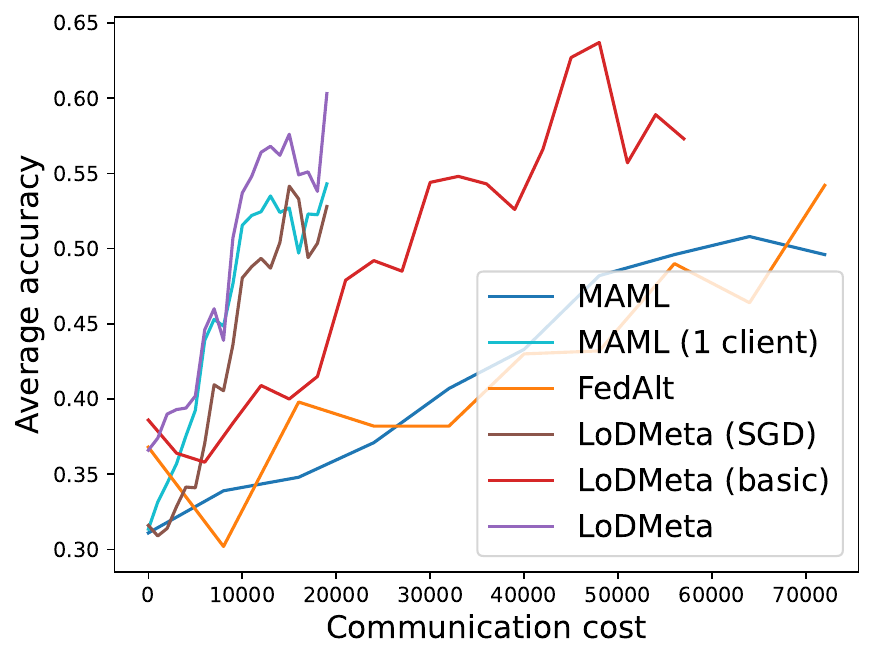}}	
\vspace{-.1in}	
	\caption{Average testing accuracy with communication cost for unseen clients on \textit{Meta-Datasets} under 5-shot setting. 
	\label{fig:rmeta_5_comm}}
	\vspace{-10px}
\end{figure*}

\subsection{Results}
{\bf \textit{Mini-ImageNet}}.
Figure~\ref{fig:mini_1} compares the testing accuracies of training clients with communication cost for different methods. 
We use the relative communication cost as in Table~\ref{tab:comm}. 
Among random-walk methods, while LoDMeta(SGD) performs a bit worse, 
LoDMeta(basic) and LoDMeta
achieve comparable performances with the centralized 
learning methods (MAML and FedAlt), 
and do not need additional central server to coordinate the learning process. 
This agrees with Theorem~\ref{thm:conv}, 
which shows Algorithm~\ref{alg:meta} has the same asymptotic convergence rate as centralized methods. 
It also demonstrates the necessity of using adaptive optimizers for meta-learning problems. 
Moreover, L2C has worse performance than reported in \cite{li2022learning}. This
may be due to that we use fewer
training samples and 
smaller
number of neighbors, and
L2C overfits.\footnote{For the {\it mini-ImageNet} experiment,
\cite{li2022learning}
use 500 samples for each client (50 samples per class), and each client has 10
neighbors. Here, we 
use 100 samples for each client (20 samples per class), and the maximum number of neighbors is 5.}
LoDMeta 
is also more preferable than
LoDMeta(basic)
in the decentralized setting due to its smaller communication cost.  

Figure~\ref{fig:rmini_1} compares the testing accuracy of unseen clients with communication cost for different methods and communication networks. 
L2C is not compared on the unseen clients as it can only produce models for the training clients. 
Similar to the testing accuracies for training clients, LoDMeta (basic) and LoDMeta both
achieve comparable or even better performance than the centralized 
learning methods (MAML and FedAlt), 
and LoDMeta has significantly smaller communication cost 
compared with LoDMeta (basic). 



\noindent
{\bf \textit{Meta-Datasets}}.
Figure~\ref{fig:meta_5_comm} compares the testing accuracies of training clients with communication cost for different methods and communication networks. 
Within limited communication resources, LoDMeta achieves the best performances, 
which comes from its significantly smaller per-iteration communication cost (1/3 of L2C/LoDMeta(basic) as in Table~\ref{tab:comm}). 
Among all the baseline methods, L2C still has poorer performances than both centralized methods (MAML and FedAlt) 
and random-walk methods (LoDMeta(SGD), LoDMeta(basic) and LoDMeta). 

Figure~\ref{fig:rmeta_5_comm} compares the testing accuracy of unseen clients with communication cost for different methods and communication networks. 
LoDMeta(basic) and LoDMeta 
achieve much better performances than the centralized 
learning methods (MAML and FedAlt). 
Compared with LoDMeta(basic), 
LoDMeta further reduces the communication cost, 
and achieves the best performance. 


\noindent
{\bf Effect of Random Perturbations for Privacy}.
Since there are limited works on privacy protection for decentralized meta-learning, 
here we study the performance of LoDMeta at different amounts of privacy perturbation, 
which is controlled by the two hyper-parameters $\epsilon, \delta$ 
used to generate the random perturbation $\bm{\epsilon}_t$.
Table~\ref{tab:prt} (resp. Table~\ref{tab:pru})
compares the testing accuracies on training (resp. unseen)
clients with different $\epsilon$ and $\delta$'s in Algorithm~\ref{alg:meta}. 
As is shown in Theorem~\ref{thm:iteration}, 
a larger perturbation (which corresponds to a smaller $\epsilon$ or $\delta$)
leads to better privacy protection.
From both Tables~\ref{tab:prt} and \ref{tab:pru}, 
a smaller $\epsilon$ or $\delta$ leads to worse testing accuracies. 
Hence, there is a trade-off between privacy protection and model performance, 
which agrees with studies on other settings~\cite{hu2020personalized, cyffers2022privacy}. 

\section{Conclusion}
\label{sec:con}

In this paper, we proposed a novel 
random-walk-based
decentralized meta-learning algorithm (LoDMeta)
in which the learning clients
perform different tasks with
limited data.
It uses local auxiliary parameters to 
remove the communication overhead associated with
adaptive optimizers.
To better protect data privacy for each client,
LoDMeta also introduces random perturbations to the model parameter.
Theoretical analysis demonstrates that LoDMeta
achieves the same convergence rate as centralized
meta-learning algorithms. 
Empirical 
few-shot learning 
results 
demonstrate that LoDMeta has similar
accuracy as centralized meta-learning algorithms, but does not require gathering
data from each client and is able to protect data privacy for each client. 


%
%


\bibliography{dec_meta}
\bibliographystyle{plain}

\newpage
\section*{NeurIPS Paper Checklist}

\begin{enumerate}

\item {\bf Claims}
    \item[] Question: Do the main claims made in the abstract and introduction accurately reflect the paper's contributions and scope?
    \item[] Answer: \answerYes{} 
    \item[] Justification: the claims made in abstract and introduction (section 1) clearly reflects the paper's contributions and scope. 

\item {\bf Limitations}
    \item[] Question: Does the paper discuss the limitations of the work performed by the authors?
    \item[] Answer: \answerYes{} 
    \item[] Justification: we have discussed possible limitations in in Appendix~\ref{app:lim}.  

\item {\bf Theory Assumptions and Proofs}
    \item[] Question: For each theoretical result, does the paper provide the full set of assumptions and a complete (and correct) proof?
    \item[] Answer: \answerYes{} 
    \item[] Justification: the assumptions are all mentioned in section~\ref{sec:ana}, and all proofs can be found in Appendix~\ref{app:proof}. 

    \item {\bf Experimental Result Reproducibility}
    \item[] Question: Does the paper fully disclose all the information needed to reproduce the main experimental results of the paper to the extent that it affects the main claims and/or conclusions of the paper (regardless of whether the code and data are provided or not)?
    \item[] Answer: \answerYes{} 
    \item[] Justification: we have mentioned necessary experimental settings in Appendix~\ref{app:dex} to reproduce our experimental results. 

\item {\bf Open access to data and code}
    \item[] Question: Does the paper provide open access to the data and code, with sufficient instructions to faithfully reproduce the main experimental results, as described in supplemental material?
    \item[] Answer: \answerNo{} 
    \item[] Justification: since our code also includes scripts from other open-source packages, it takes more time to arrange our code. We will release our code along with camera-ready version if our submission is accepted.  

\item {\bf Experimental Setting/Details}
    \item[] Question: Does the paper specify all the training and test details (e.g., data splits, hyperparameters, how they were chosen, type of optimizer, etc.) necessary to understand the results?
    \item[] Answer: \answerYes{} 
    \item[] Justification: we have mentioned necessary experimental settings (including data splits, hyper-parameters and how they were chosen, type of optimizer) in Appendix~\ref{app:dex}.  

\item {\bf Experiment Statistical Significance}
    \item[] Question: Does the paper report error bars suitably and correctly defined or other appropriate information about the statistical significance of the experiments?
    \item[] Answer: \answerNo{} 
    \item[] Justification: we do not report error bars in our experimental results. 

\item {\bf Experiments Compute Resources}
    \item[] Question: For each experiment, does the paper provide sufficient information on the computer resources (type of compute workers, memory, time of execution) needed to reproduce the experiments?
    \item[] Answer: \answerYes{} 
    \item[] Justification: we mentioned compute resources used in our experiments in Appendix~\ref{app:dex}.  
    
\item {\bf Code Of Ethics}
    \item[] Question: Does the research conducted in the paper conform, in every respect, with the NeurIPS Code of Ethics \url{https://neurips.cc/public/EthicsGuidelines}?
    \item[] Answer: \answerYes{} 
    \item[] Justification: we have thoroughly checked the code of ethics and found no conflict. 

\item {\bf Broader Impacts}
    \item[] Question: Does the paper discuss both potential positive societal impacts and negative societal impacts of the work performed?
    \item[] Answer: \answerYes{} 
    \item[] Justification: we have discussed possible societal impacts of our proposed method in Appendix~\ref{app:lim}.  
    
\item {\bf Safeguards}
    \item[] Question: Does the paper describe safeguards that have been put in place for responsible release of data or models that have a high risk for misuse (e.g., pretrained language models, image generators, or scraped datasets)?
    \item[] Answer: \answerNA{} 
    \item[] Justification: this paper does not involve releasing any data or models that may have risk for misuse. 

\item {\bf Licenses for existing assets}
    \item[] Question: Are the creators or original owners of assets (e.g., code, data, models), used in the paper, properly credited and are the license and terms of use explicitly mentioned and properly respected?
    \item[] Answer: \answerYes{} 
    \item[] Justification: we mentioned licenses for data sets used in our work in Appendix~\ref{app:dex}.  

\item {\bf New Assets}
    \item[] Question: Are new assets introduced in the paper well documented and is the documentation provided alongside the assets?
    \item[] Answer: \answerNA{} 
    \item[] Justification: this paper does not release any new assets. 

\item {\bf Crowdsourcing and Research with Human Subjects}
    \item[] Question: For crowdsourcing experiments and research with human subjects, does the paper include the full text of instructions given to participants and screenshots, if applicable, as well as details about compensation (if any)? 
    \item[] Answer: \answerNA{} 
    \item[] Justification: this paper does not involve crowdsourcing nor research with human subjects.

\item {\bf Institutional Review Board (IRB) Approvals or Equivalent for Research with Human Subjects}
    \item[] Question: Does the paper describe potential risks incurred by study participants, whether such risks were disclosed to the subjects, and whether Institutional Review Board (IRB) approvals (or an equivalent approval/review based on the requirements of your country or institution) were obtained?
    \item[] Answer: \answerNA{} 
    \item[] Justification: this paper does not involve crowdsourcing nor research with human subjects.

\end{enumerate}

\newpage
\appendix

\section{Possible Limitations and Broader Impacts}
\label{app:lim}

{\bf Limitations.} One possible limitation of this work is that we only consider network DP for privacy protection. 
We will consider other privacy metric as future works. 

{\bf Broader Impacts.} As a paper on pure machine learning algorithms, there should be no direct societal impact of this work. 
Our proposed algorithm is not about generative models and there is no concern on generating fake contents. 

\section{Algorithms}
\label{app:alg}

\begin{algorithm}[ht]
\caption{Model adaptation on unseen clients.} 
\label{alg:adp}
\begin{algorithmic}[1]
\STATE {\bfseries Input:} meta-trained model $\vw_T$. 
\STATE initialize $\vu_0 = \vw_T$
   \FOR{$k=0$ {\bfseries to} $K-1$}
   \STATE compute $\vg_{k} = \nabla \ell(\vu_k;\xi^s_{i_t})$ with support data $\xi^s_{i_t}$ of client $i_t$;
   \STATE update $\vu_{k+1} = \vu_k - \alpha \vg_k$;
\ENDFOR
\STATE obtain the final model $\vu_K$ for testing
\end{algorithmic}
\end{algorithm}

\section{Details for Experiments}
\label{app:dex}

\begin{figure}[ht]
\begin{center}
\subfigure[Small-world network. \label{fig:small}]
	{\includegraphics[width=0.2\columnwidth]{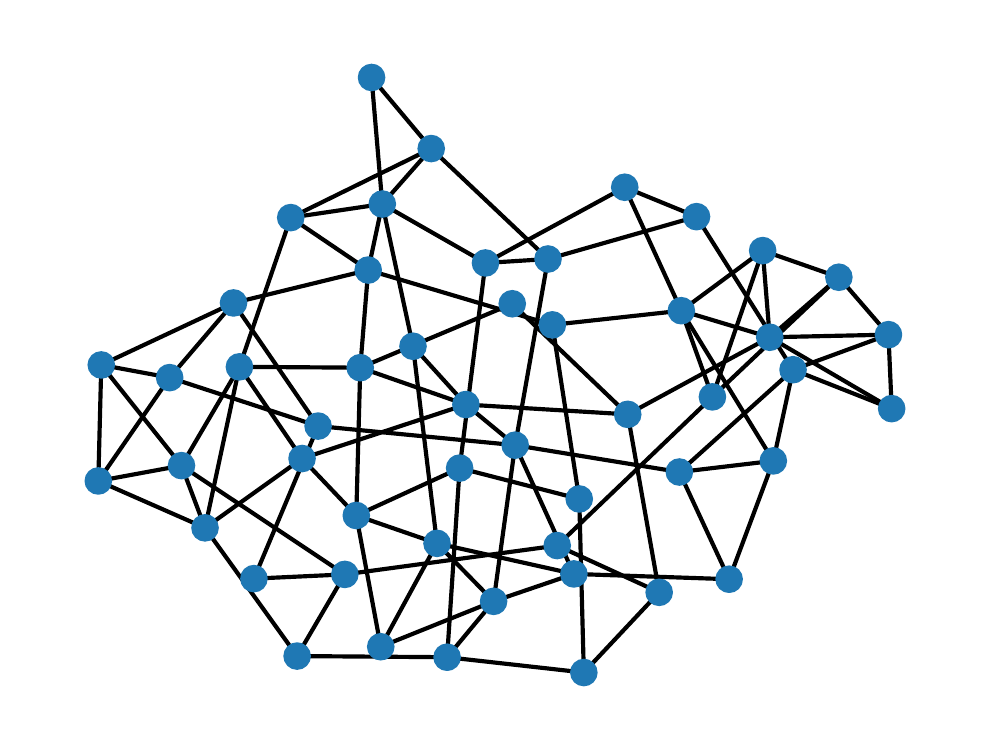}}	
	\subfigure[3-regular expander network. \label{fig:regular}]
	{\includegraphics[width=0.3\columnwidth]{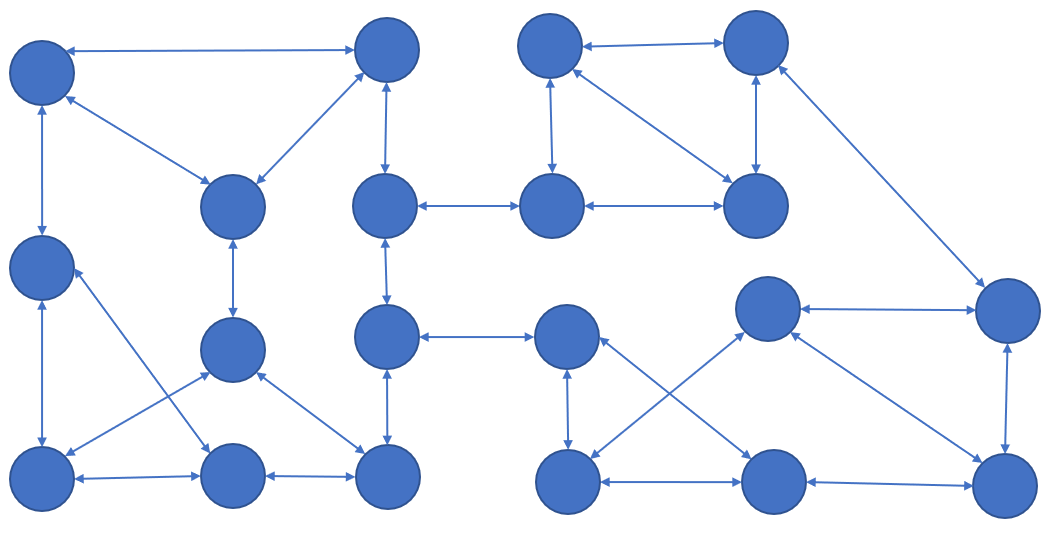}}	
 \vspace{-.1in}
\caption{Example 
communication networks used in the experiments.}
\end{center}
\end{figure}

Some statistics for data sets used in experiments are in Table~\ref{tab:stats}. 
All data sets used in our experiments are released under Apache 2.0 license. 
Figure~\ref{fig:small} gives an example for small-world network, 
while Figure~\ref{fig:regular} gives an example for 3-regular expander network. 
These two network types are used in our experiments (with different number of clients). 

\begin{table*}[ht]
\caption{Statistics for the data sets used in experiments.}
\label{tab:stats}
\begin{center}
\begin{tabular}{c c c cc c c c}
\toprule
& & \multicolumn{2}{c}{number of classes} &  & \multicolumn{2}{c}{\#clients for 1-shot/5-shot setting}\\ 
& & meta-training & meta-testing &  \#samples per class & training clients &  unseen clients \\ \midrule
\multirow{4}{*}{Meta-Dataset} & 
\textit{Bird} & 80 & 20 & 60 & 38/38 & 12/12 \\
 & \textit{Texture} & 37 & 10 & 120 & 42/36 & 14/12\\
 & \textit{Aircraft} & 80 & 20 & 100 & 76/64 & 24/20 \\
 & \textit{Fungi} & 80 & 20 & 150 & 115/89 & 36/28 \\
\midrule
\multicolumn{2}{c}{\textit{mini-Imagenet}} & 80 & 20 & 600 & 380/380 & 120/120 \\
\bottomrule
\end{tabular}
\end{center}
\end{table*}

All experiments are run on a single RTX2080 Ti GPU. 
Following~\cite{finn2017maml, nichol2018reptile}, we use the CONV4\footnote{The CONV4 model is a 4-layer CNN. Each layer contains 64  $3 \times 3$ convolutional filters, followed by batch normalization, ReLU activation, and $2 \times 2$ max-pooling. 
} as base learner.
The hyper-parameter settings for all data sets also
follow MAML~\cite{finn2017maml}: 
learning rate $\eta$ 
is $0.001$, 
first-order momentum weight $\theta$ is $0$, 
and the second-order momentum weight $\beta$ is $0.99$.
The number of gradient descent steps 
($K$)
in the inner loop is
5.
Unless otherwise specified, we set $\epsilon=0.5$ and $\delta=0.3$ for the privacy perturbation. 

\section{Additional Experimental Results}

\subsection{Experiments on 1-shot Meta-Datasets with small-world network}

\begin{figure*}[h]
	\centering
	\subfigure[\textit{Bird}.]
	{\includegraphics[width=0.24\textwidth]{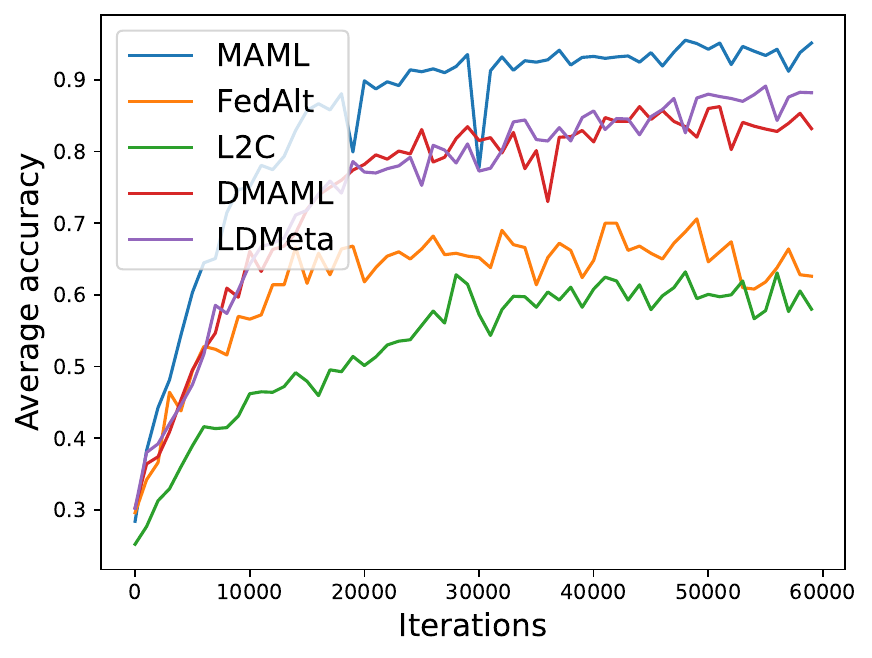}}	
	\subfigure[\textit{Texture}.]
	{\includegraphics[width=0.24\textwidth]{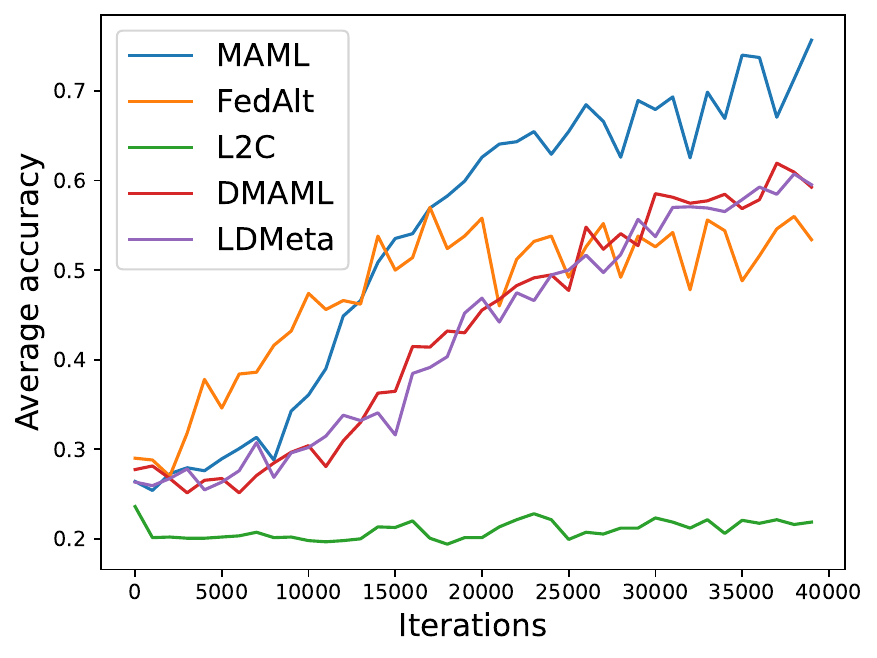}}	
	\subfigure[\textit{Aircraft}.]
	{\includegraphics[width=0.24\textwidth]{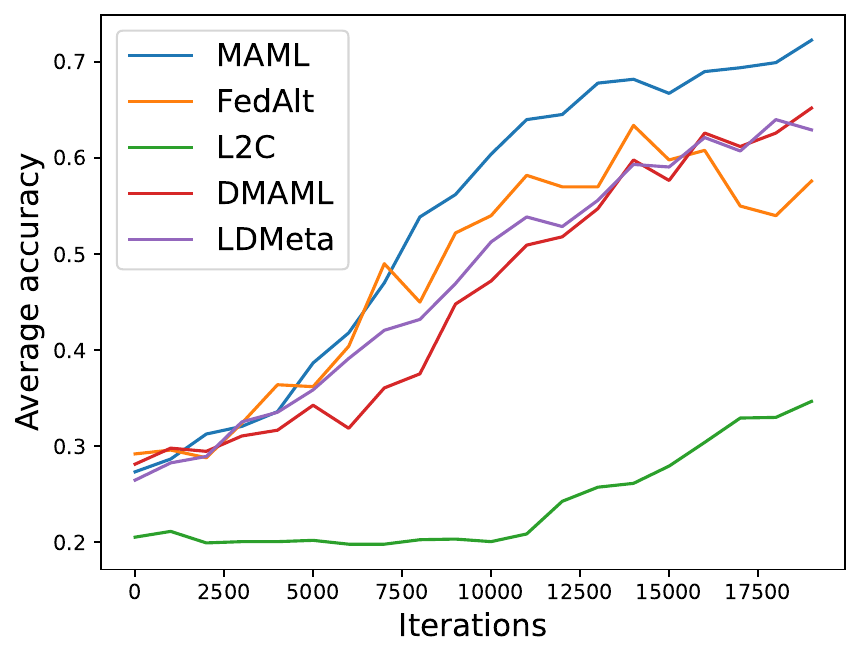}}	
	\subfigure[\textit{Fungi}.]
	{\includegraphics[width=0.24\textwidth]{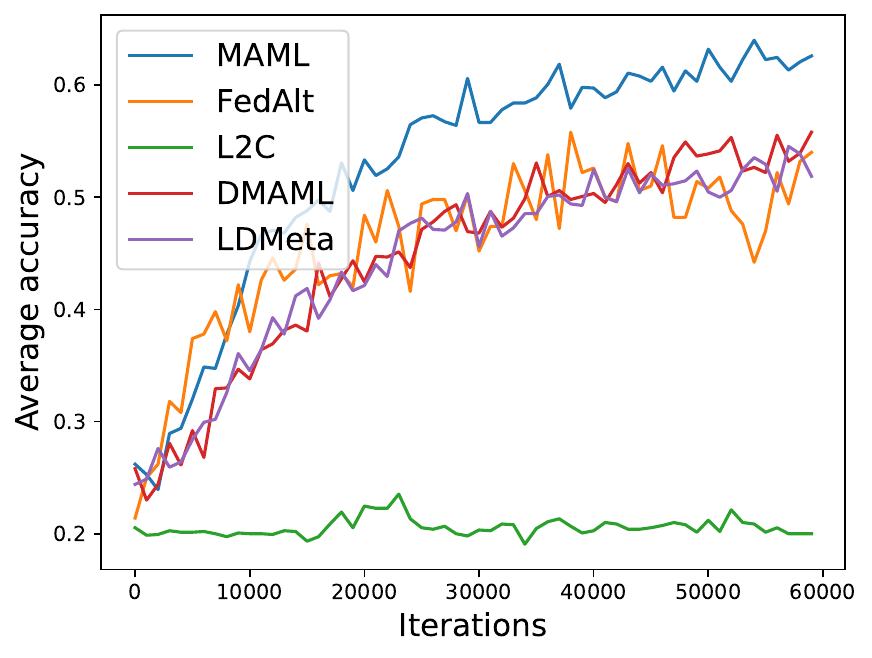}}	
	\subfigure[\textit{Bird}.]
	{\includegraphics[width=0.24\textwidth]{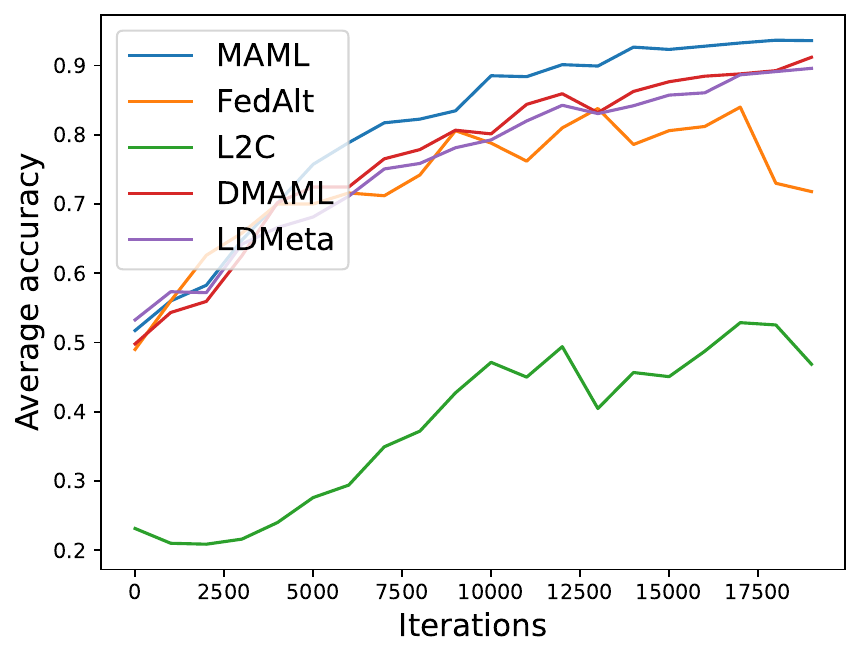}}	
	\subfigure[\textit{Texture}.]
	{\includegraphics[width=0.24\textwidth]{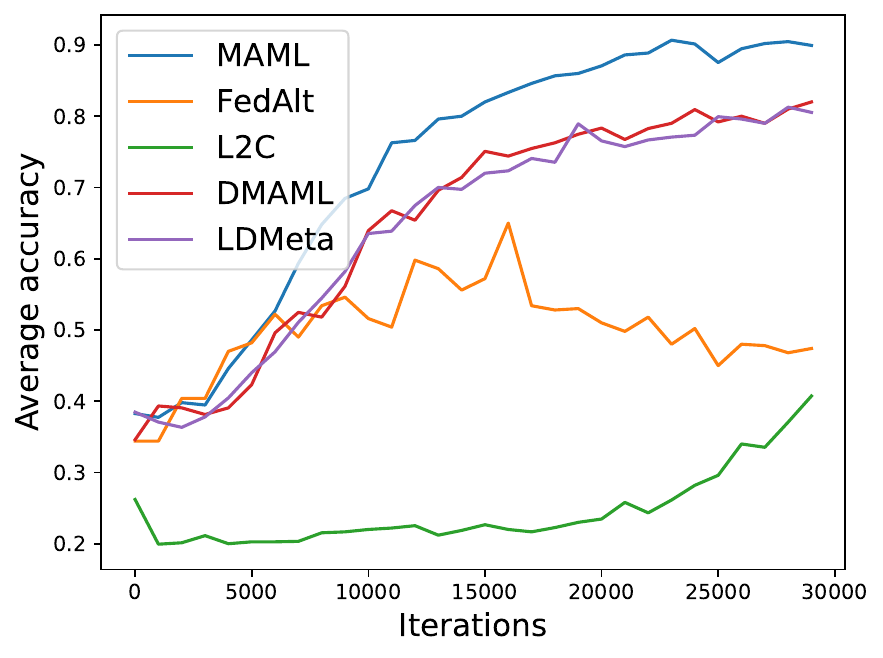}}	
	\subfigure[\textit{Aircraft}.]
	{\includegraphics[width=0.24\textwidth]{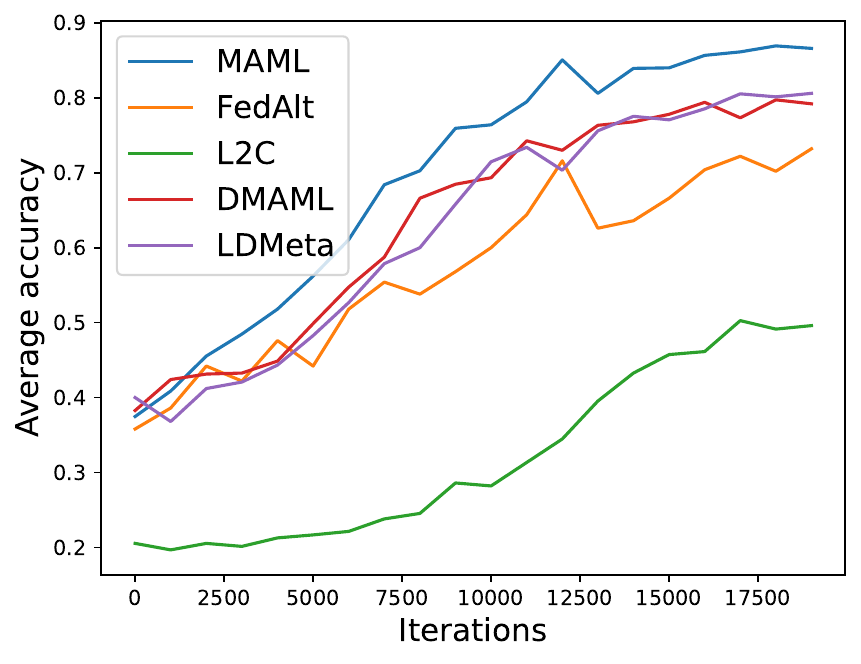}}	
	\subfigure[\textit{Fungi}.]
	{\includegraphics[width=0.24\textwidth]{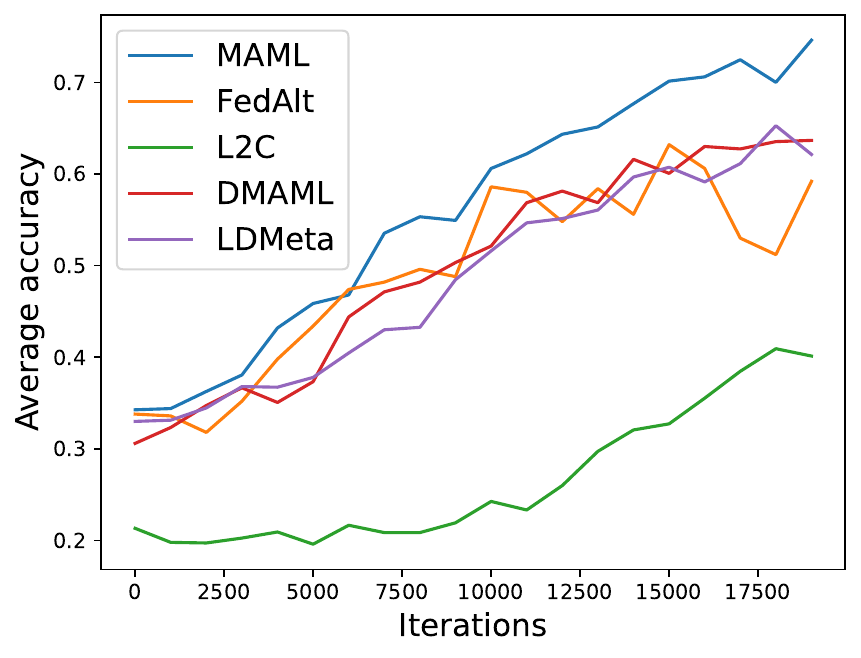}}	
	
	\caption{Average testing accuracy with iterations on \textit{Meta-Datasets} under 1-shot setting. Top: training clients; Bottom: unseen clients.\label{fig:meta_1}}
\end{figure*}

\begin{figure*}[h]
	\centering
	\subfigure[\textit{Bird}.]
	{\includegraphics[width=0.24\textwidth]{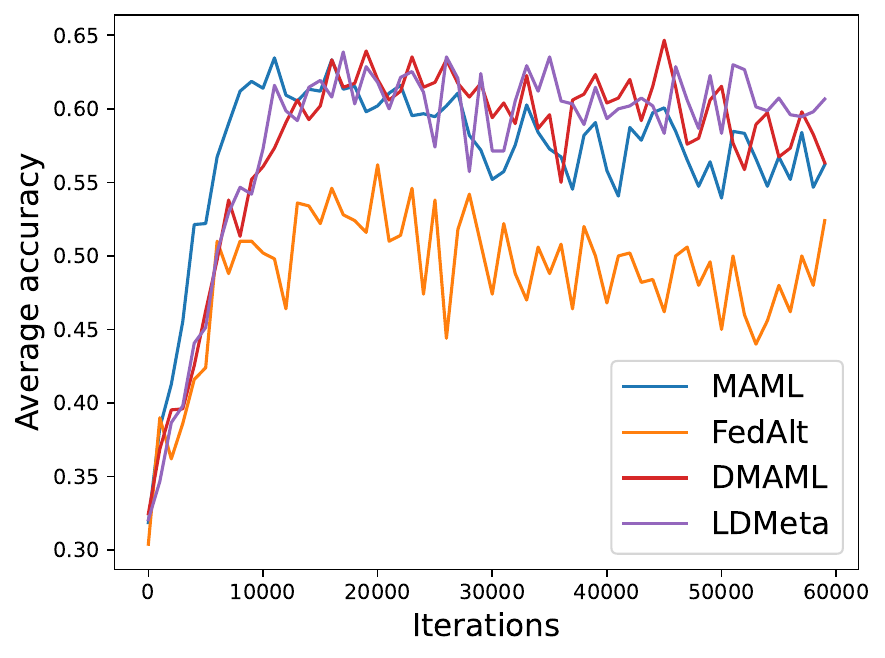}}	
	\subfigure[\textit{Texture}.]
	{\includegraphics[width=0.24\textwidth]{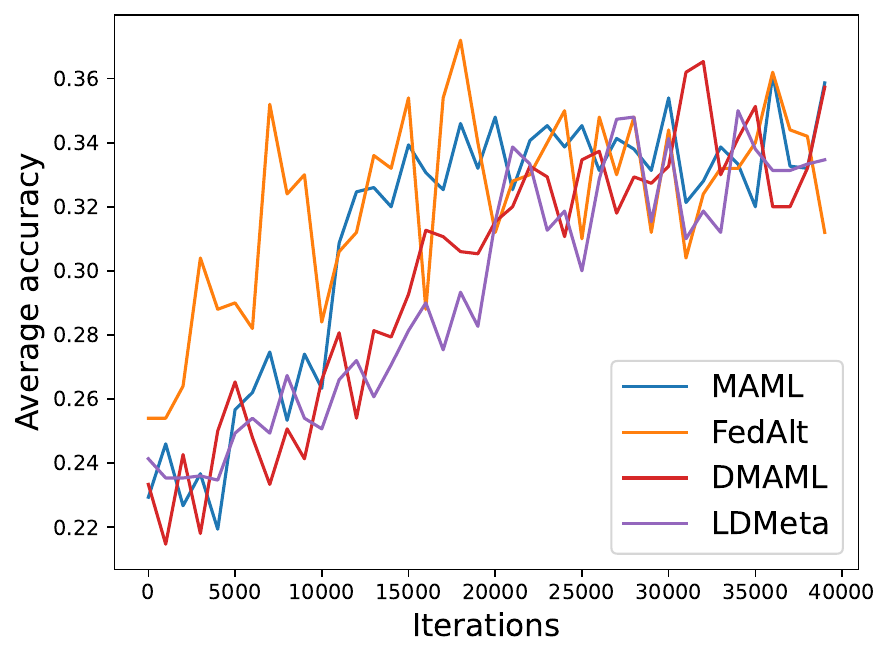}}	
	\subfigure[\textit{Aircraft}.]
	{\includegraphics[width=0.24\textwidth]{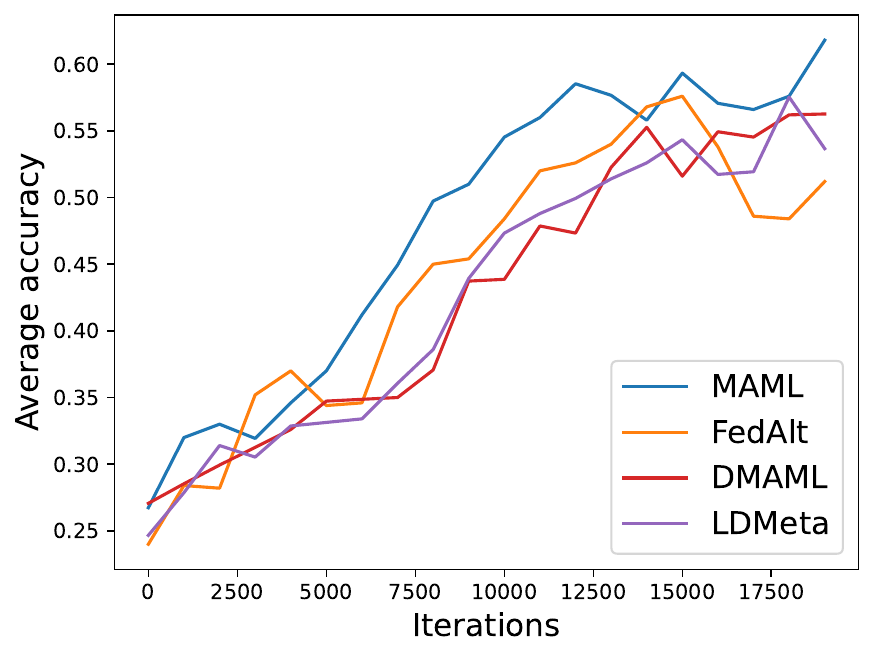}}	
	\subfigure[\textit{Fungi}.]
	{\includegraphics[width=0.24\textwidth]{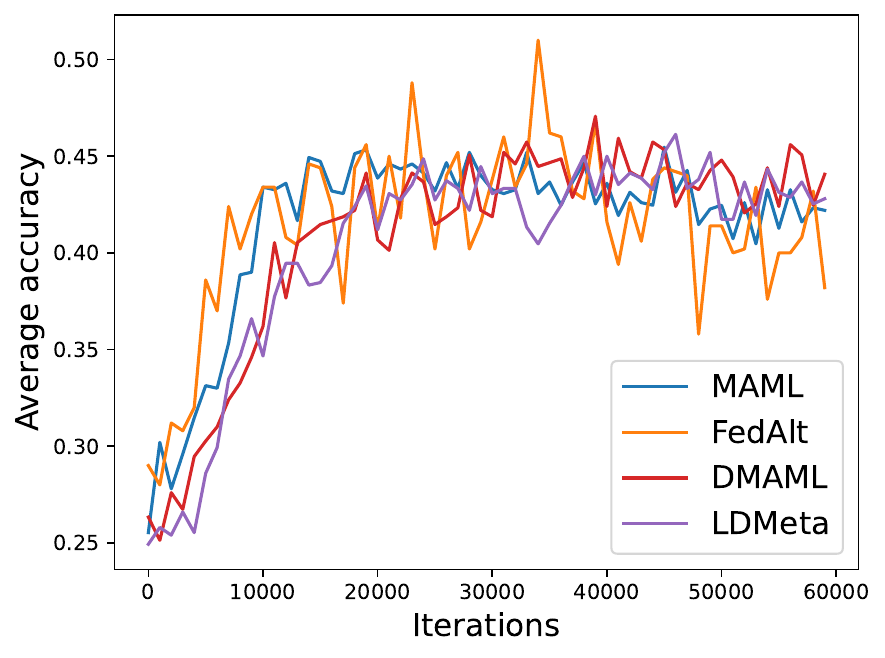}}	
	\subfigure[\textit{Bird}.]
	{\includegraphics[width=0.24\textwidth]{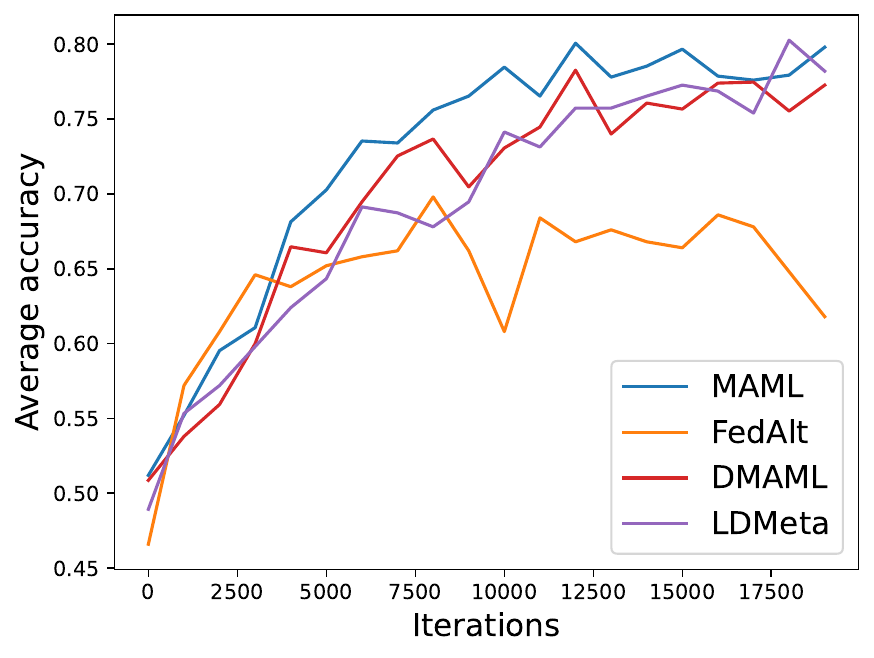}}	
	\subfigure[\textit{Texture}.]
	{\includegraphics[width=0.24\textwidth]{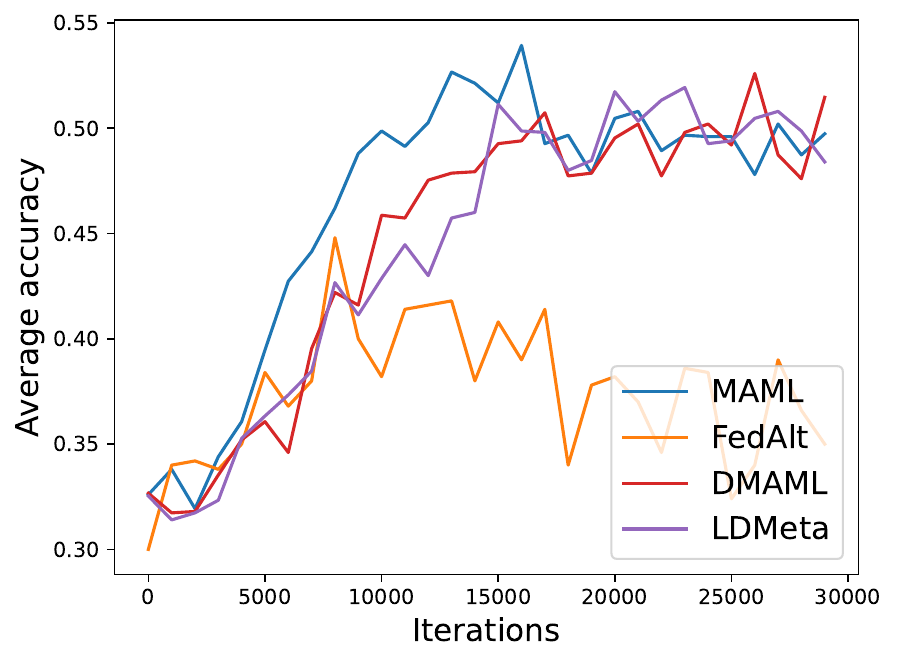}}	
	\subfigure[\textit{Aircraft}.]
	{\includegraphics[width=0.24\textwidth]{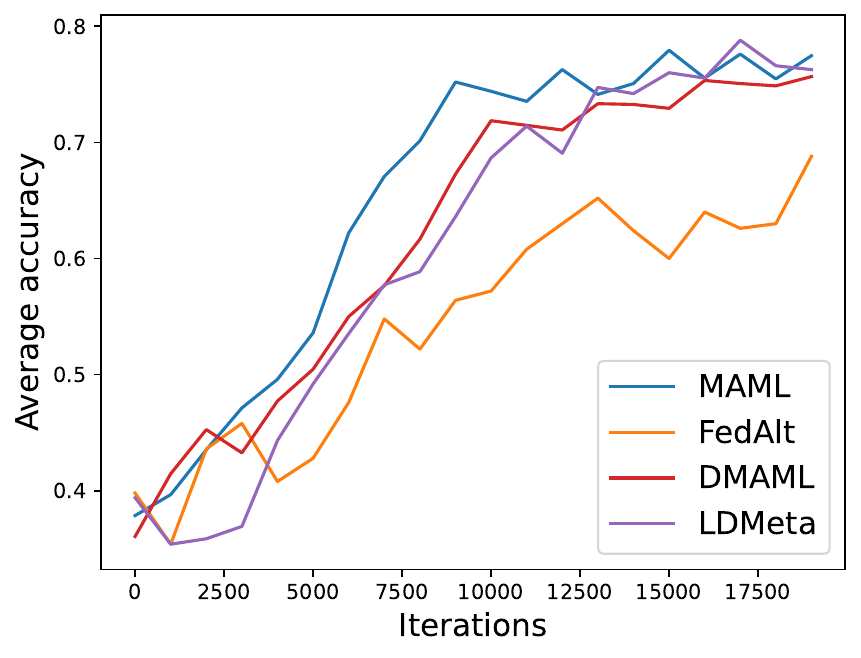}}	
	\subfigure[\textit{Fungi}.]
	{\includegraphics[width=0.24\textwidth]{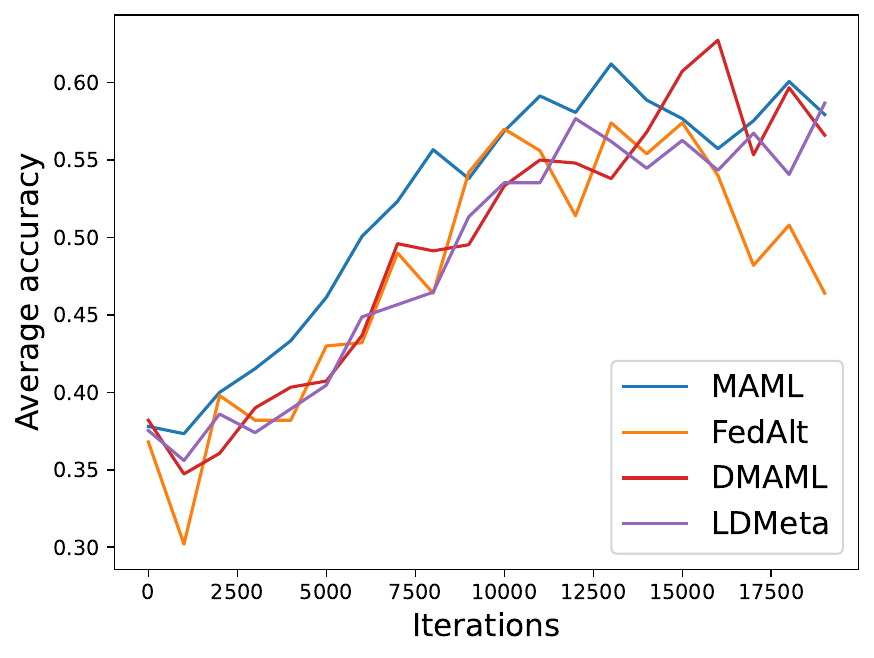}}	
	
\caption{Average testing accuracy with number of iterations on
\textit{Meta-Datasets} under the 5-shot setting. 
Top: training clients; Bottom: unseen clients.
\label{fig:meta_5}}
\end{figure*}

Figure~\ref{fig:meta_1} compares
the average testing 
accuracy across different clients
during training 
on these four data sets under the 1-shot setting
with the number of training iterations.
Similar to the 5-shot learning setting, the two random-walk algorithms
(DMAML and LDMeta)
achieve slightly worse performance than MAML, 
but better performance than FedAlt. 
Compared to the 5-shot setting (Figure~\ref{fig:meta_5}),
the gossip-based algorithm L2C performs even worse in this
1-shot setting because
each client has 
even fewer samples.

\begin{figure*}[h]
	\centering
	\subfigure[\textit{Bird}. Small-world network.]
	{\includegraphics[width=0.24\textwidth]{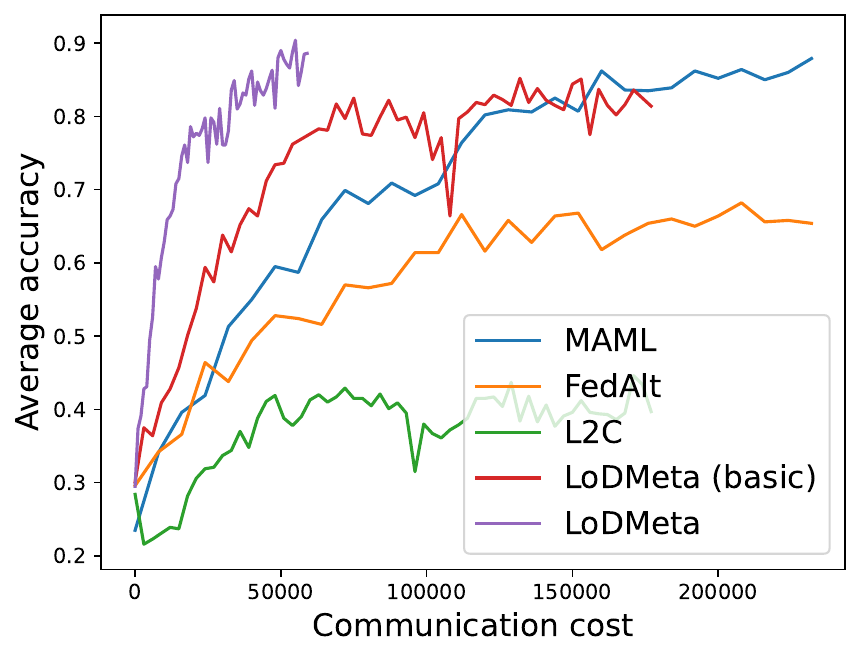}}	
	\subfigure[\textit{Texture}. Small-world network.]
	{\includegraphics[width=0.24\textwidth]{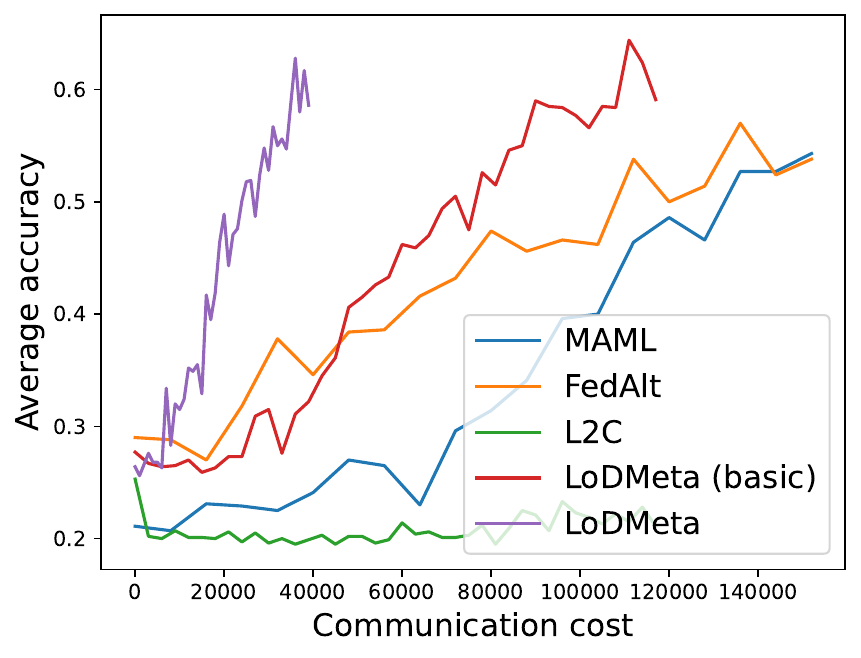}}	
	\subfigure[\textit{Aircraft}. Small-world network.]
	{\includegraphics[width=0.24\textwidth]{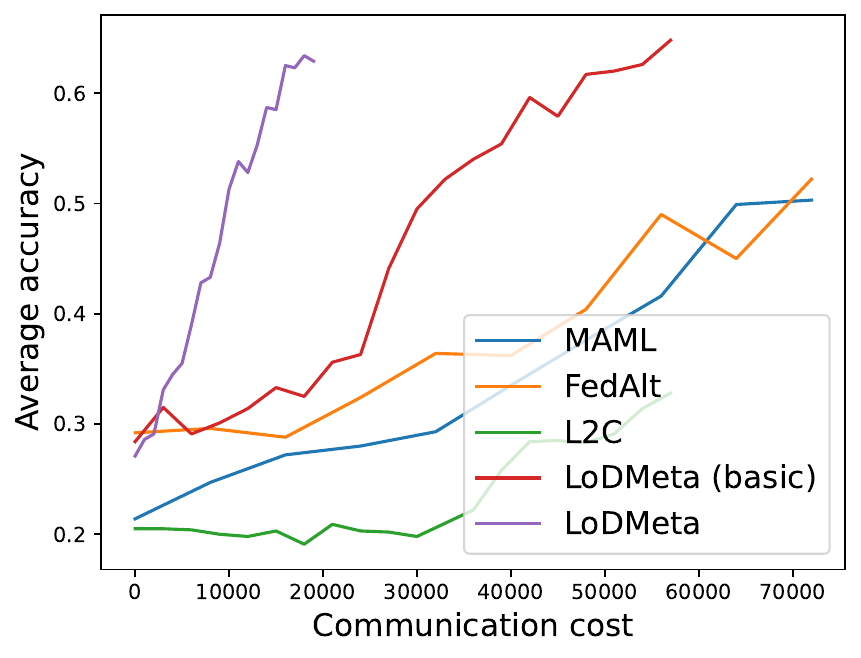}}	
	\subfigure[\textit{Fungi}. Small-world network.]
	{\includegraphics[width=0.24\textwidth]{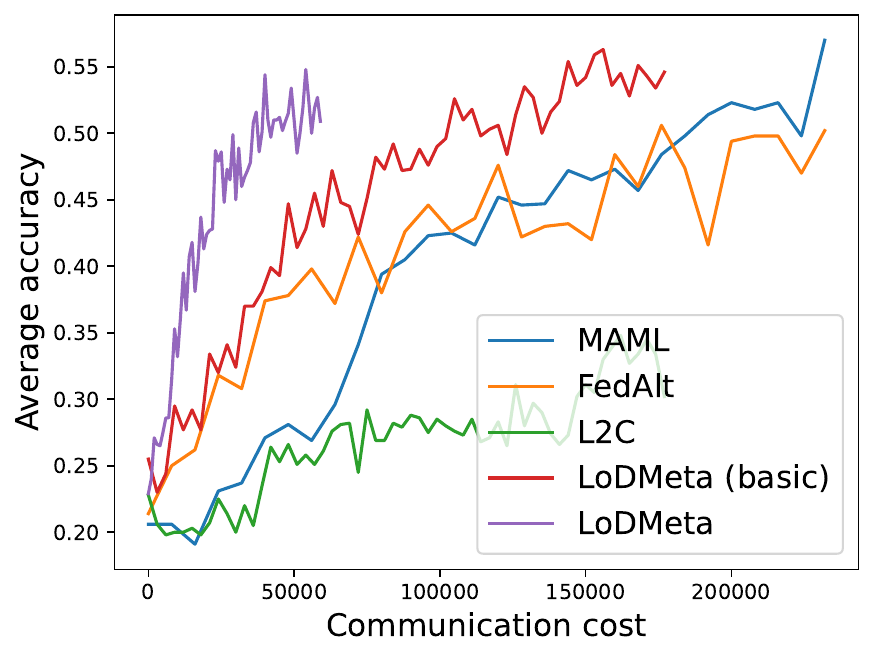}}
 \subfigure[\textit{Bird}. 3-regular expander network.]
	{\includegraphics[width=0.24\textwidth]{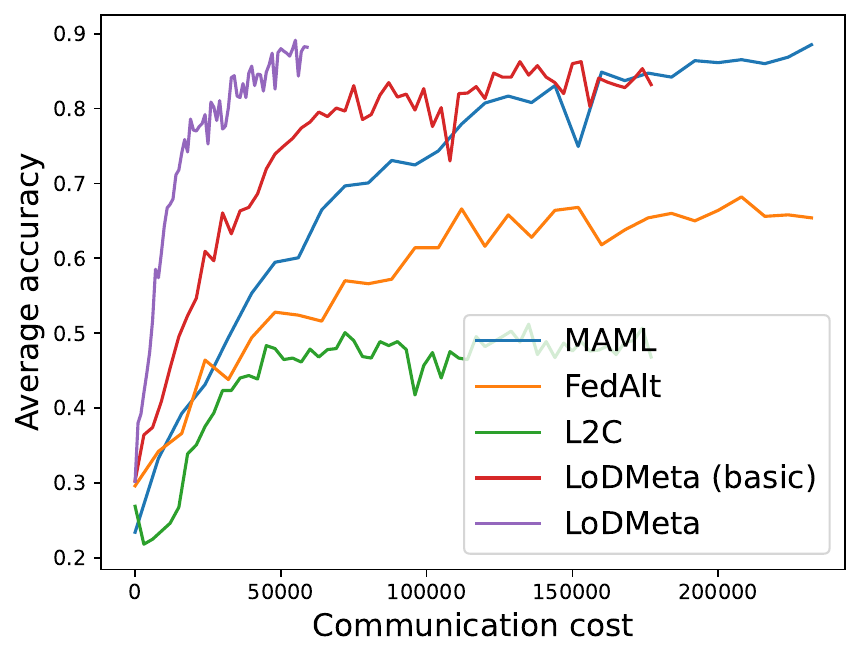}}	
	\subfigure[\textit{Texture}. 3-regular expander network.]
	{\includegraphics[width=0.24\textwidth]{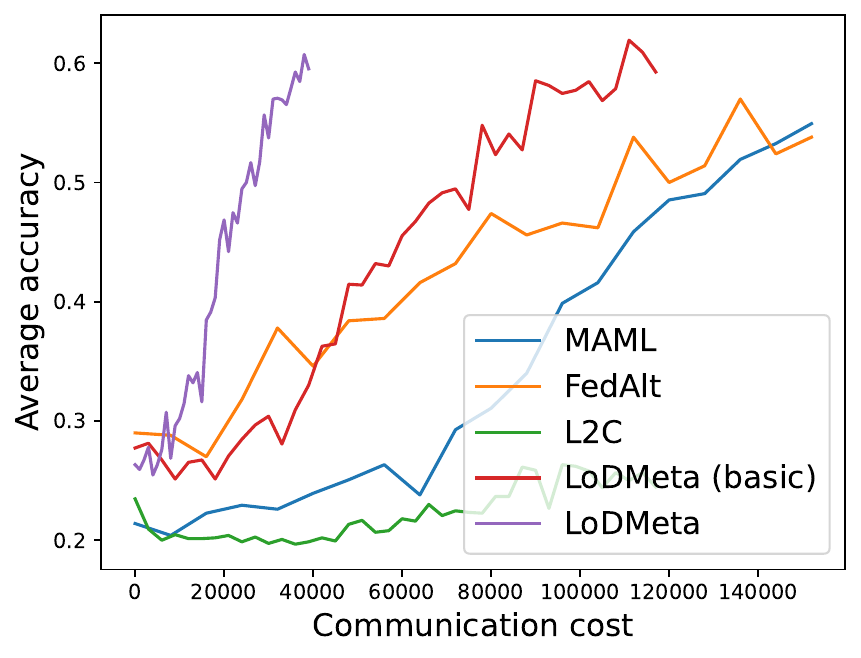}}	
	\subfigure[\textit{Aircraft}. 3-regular expander network.]
	{\includegraphics[width=0.24\textwidth]{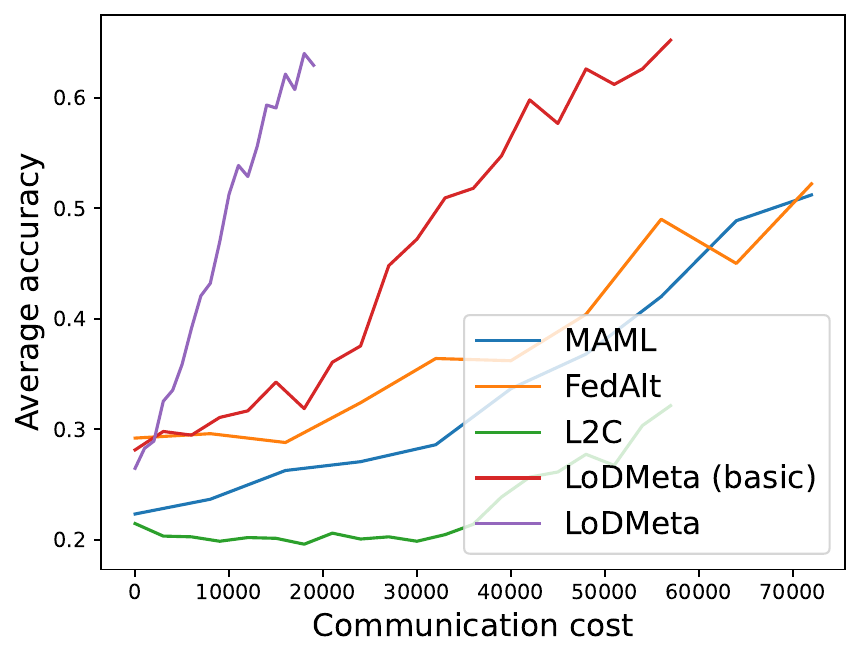}}	
	\subfigure[\textit{Fungi}. 3-regular expander network.]
	{\includegraphics[width=0.24\textwidth]{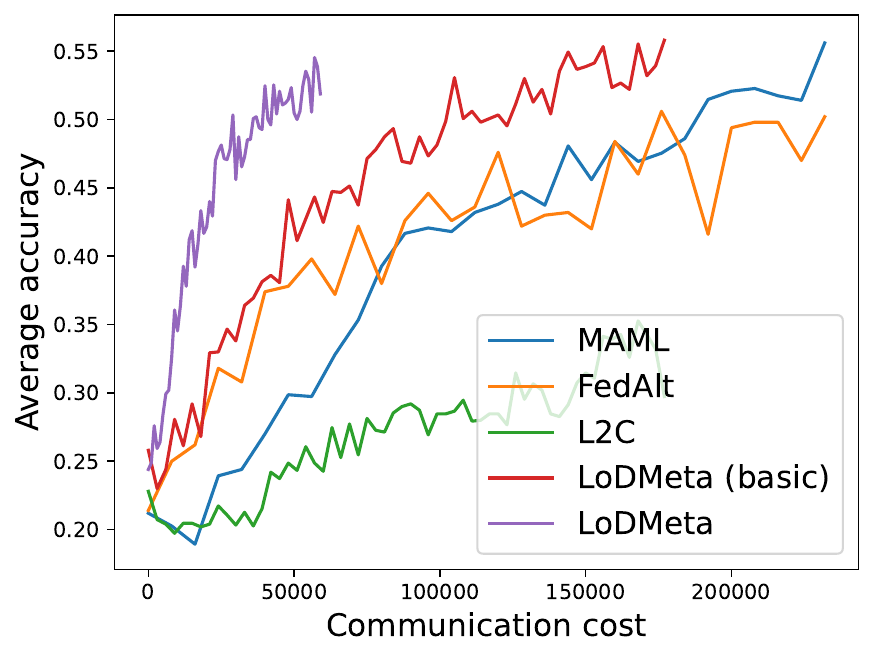}}
\vspace{-.1in}	
	\caption{Average testing accuracy with communication cost for training clients on \textit{Meta-Datasets} under 5-shot setting. 
	\label{fig:meta_1_comm}}
\end{figure*}

Figure~\ref{fig:meta_1_comm} compares
the average testing
accuracy 
across different clients during training 
on these four data sets under the 1-shot setting
with communication cost.
Similar to the 5-shot learning setting (Figure~\ref{fig:meta_5_comm}),
LDMeta has a much smaller communication cost than DMAML, 
and is more preferable when we require communication to be efficient. 

\subsection{Experiments with 3-regular network}

Here, we perform experiments on the 3-regular expander graph,  in which
all clients have 3 neighbors. 
The other settings are the same as experiments in the main text. 





\begin{figure*}[h]
	\centering
	\subfigure[\textit{Bird}. Small-world network.]
	{\includegraphics[width=0.24\textwidth]{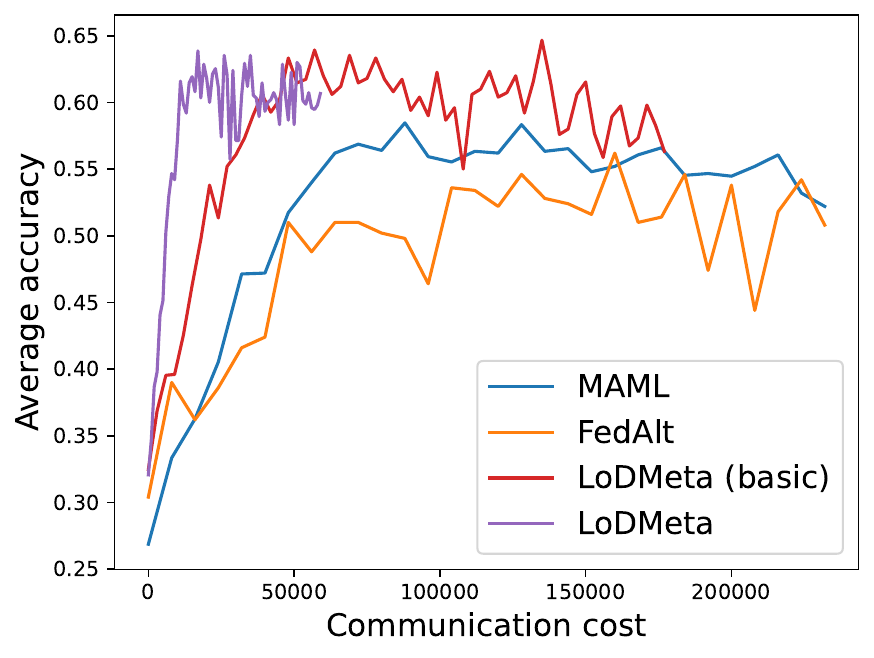}}	
	\subfigure[\textit{Texture}. Small-world network.]
	{\includegraphics[width=0.24\textwidth]{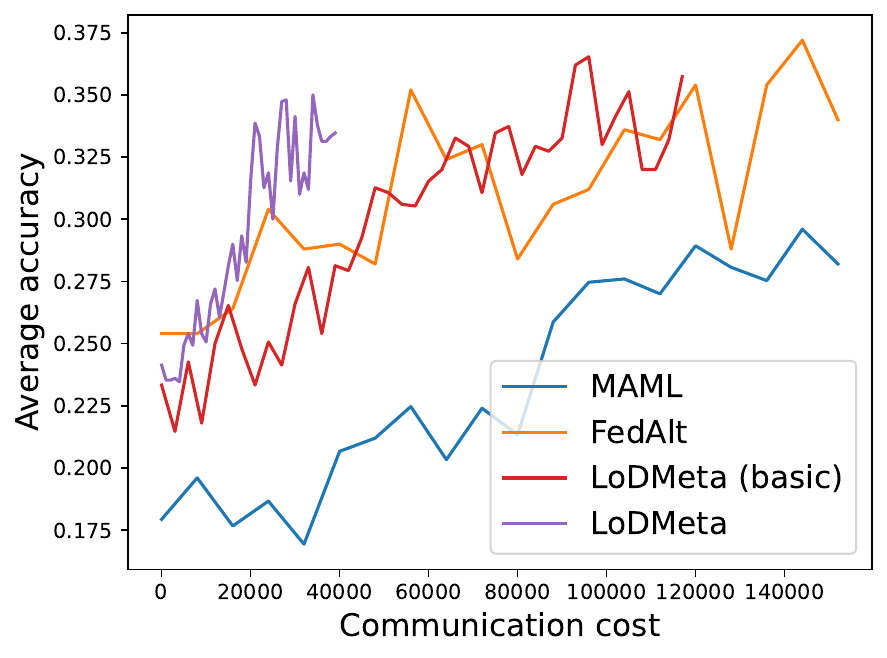}}	
	\subfigure[\textit{Aircraft}. Small-world network.]
	{\includegraphics[width=0.24\textwidth]{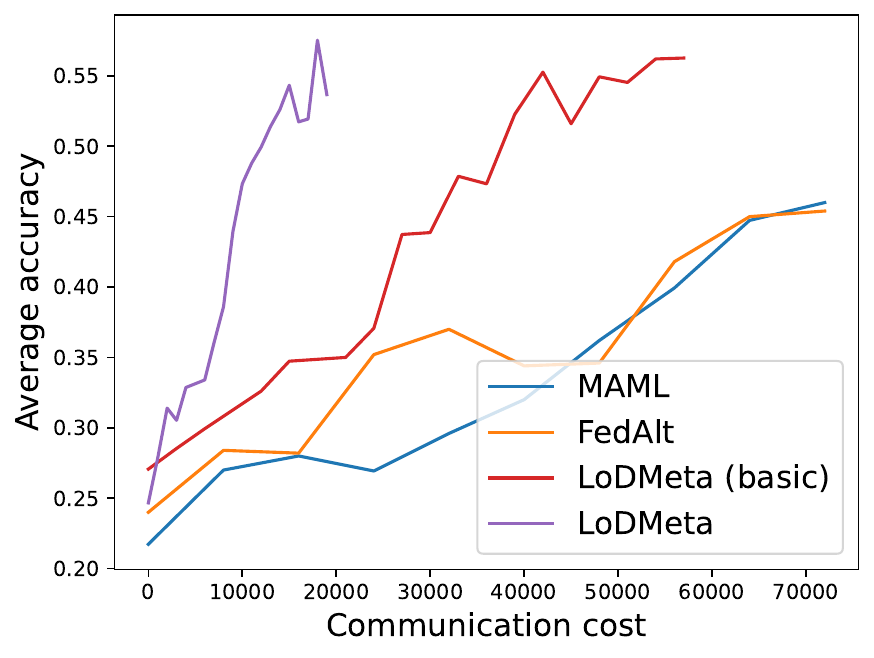}}	
	\subfigure[\textit{Fungi}. Small-world network.]
	{\includegraphics[width=0.24\textwidth]{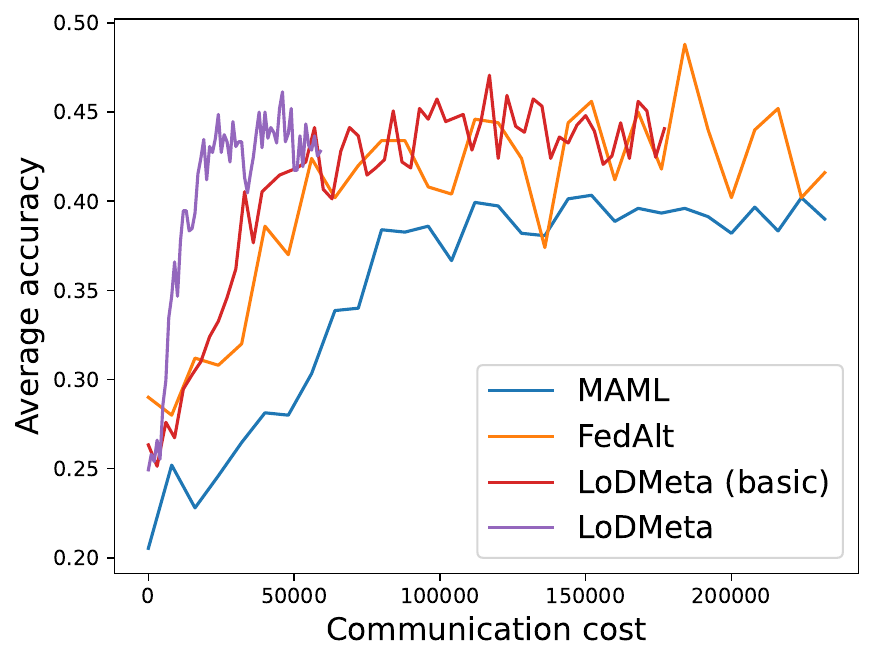}}
 \subfigure[\textit{Bird}. 3-regular expander network.]
	{\includegraphics[width=0.24\textwidth]{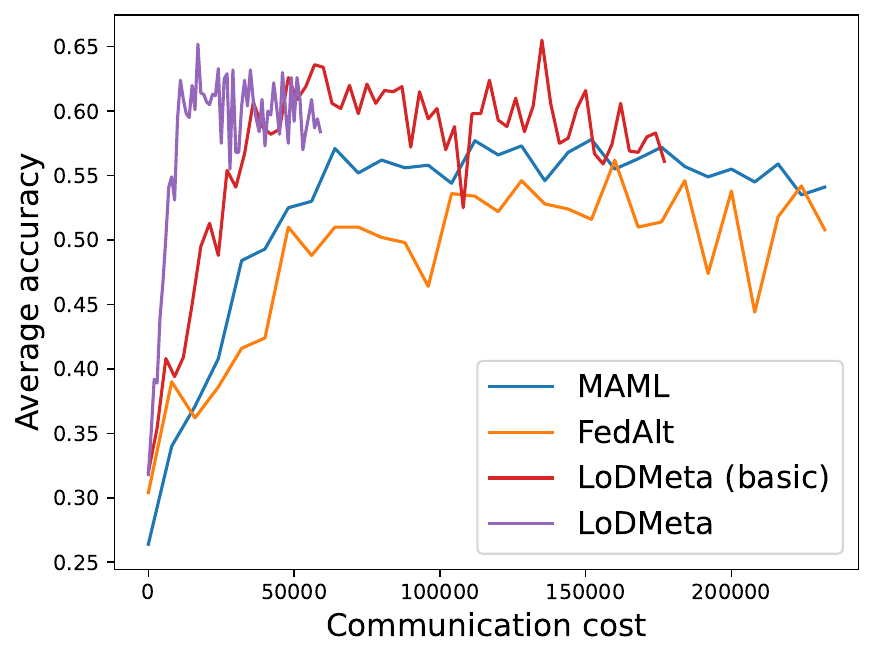}}	
	\subfigure[\textit{Texture}. 3-regular expander network.]
	{\includegraphics[width=0.24\textwidth]{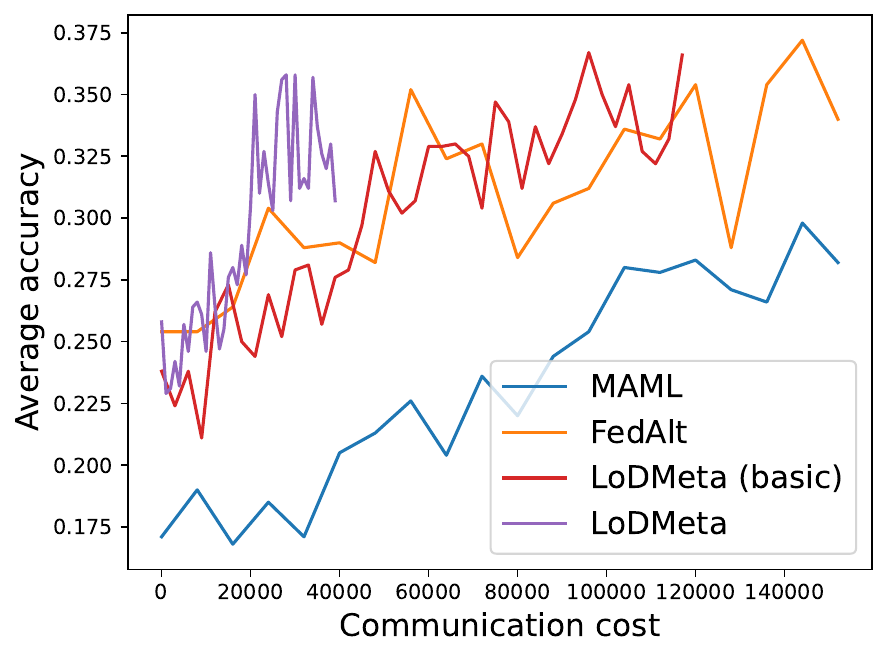}}	
	\subfigure[\textit{Aircraft}. 3-regular expander network.]
	{\includegraphics[width=0.24\textwidth]{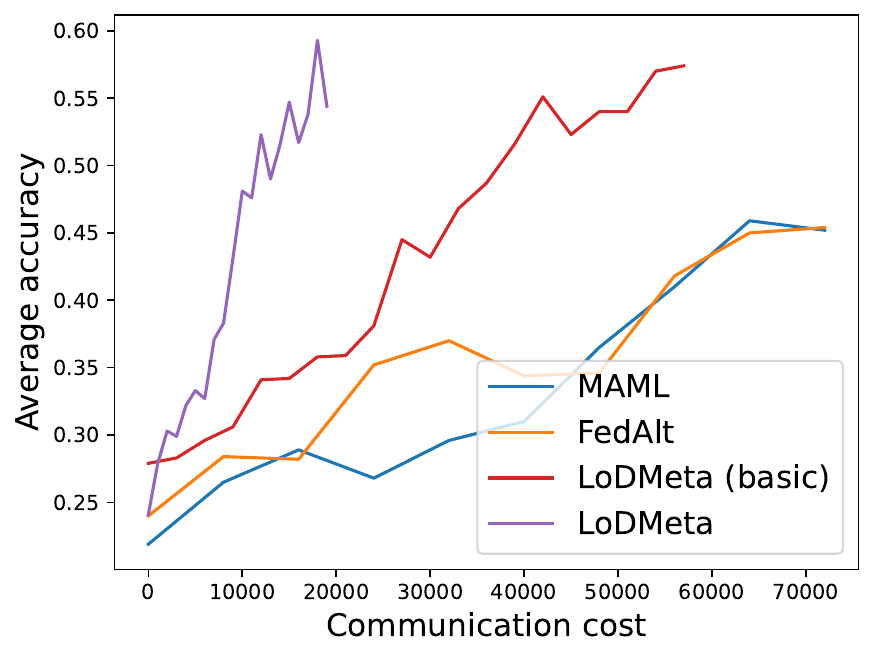}}	
	\subfigure[\textit{Fungi}. 3-regular expander network.]
	{\includegraphics[width=0.24\textwidth]{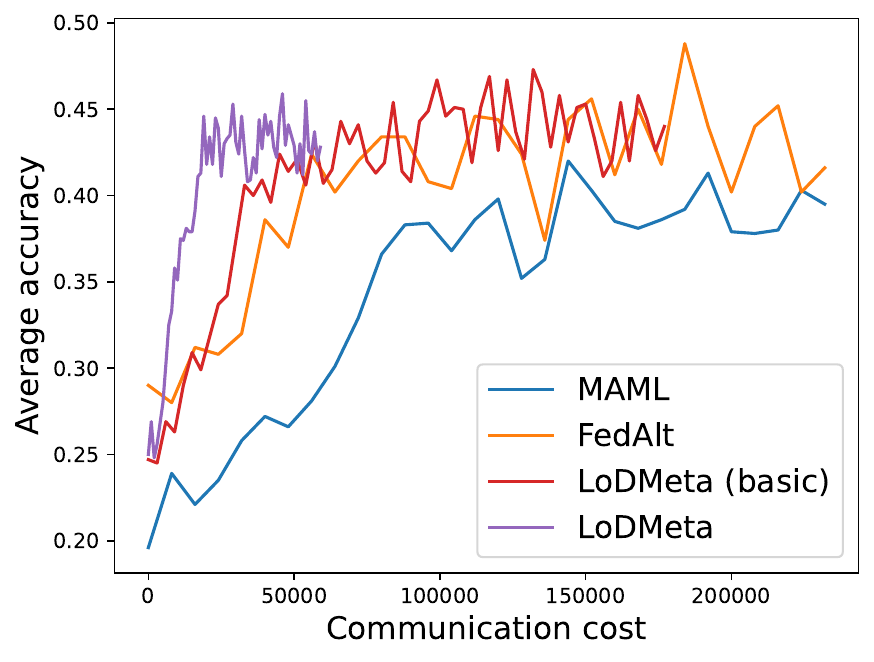}}	
\vspace{-.1in}	
	\caption{Average testing accuracy with communication cost for unseen clients on \textit{Meta-Datasets} under 5-shot setting. 
	\label{fig:rmeta_1_comm}}
\end{figure*}

Figure~\ref{fig:rmeta_1_comm} compares
the average testing
accuracy across different clients
during training 
on four data sets in Meta-Datasets
with the number of training iterations.
As can be seen, the two random-walk algorithms
(DMAML and LDMeta)
have slightly worse performance than MAML, 
but better performance than FedAlt, 
and significantly outperform the gossip-based algorithm L2C. 
This is because in the random-walk setting, only one client needs to update the meta-model in each iteration, 
while personalized federated learning methods require multiple clients to update the meta-model. 


%

\section{Proofs}
\label{app:proof}

\subsection{Proof of Proposition~\ref{prop:lip}}
\label{app:lip}

By the definition of $G_i(\cdot)$, we have 
	\begin{align}\label{lopasv}
	\|G_i(\vw)  - G_i(\vu) \| 
	\leq &\Big\|\prod_{k=0}^{K-1}(\mI - \alpha \nabla^2 \ell(\vw_k; \xi^s_i))\nabla
\ell(\vw_K; \xi^q_i) -\prod_{k=0}^{K-1}(\mI - \alpha \nabla^2 \ell(\vu_k; \xi^s_i))\nabla
\ell(\vw_K; \xi^q_i)\Big\|  \nonumber
	\\ & + \Big\|\prod_{k=0}^{K-1}(\mI - \alpha \nabla^2 \ell(\vu_k; \xi^s_i))\nabla
\ell(\vw_K; \xi^q_i) -\prod_{k=0}^{K-1}(\mI - \alpha \nabla^2 \ell(\vu_k; \xi^s_i))\nabla
\ell(\vu_K; \xi^q_i)\Big\| \nonumber
	\\ \leq &\underbrace{ \Big\|\prod_{k=0}^{K-1}(\mI - \alpha \nabla^2 \ell(\vw_k; \xi^s_i)) -\prod_{k=0}^{K-1}(\mI - \alpha \nabla^2 \ell(\vu_k; \xi^s_i))\Big\|}_{A}  \| \nabla
\ell(\vw_K; \xi^q_i) \|  \nonumber
	\\& + (1+\alpha M)^K \|\nabla
\ell(\vw_K; \xi^q_i) - \nabla
\ell(\vu_K; \xi^q_i)\|.
	\end{align}
	We next upper-bound $A$ in the above inequality. Specifically,  we have
	\begin{align}\label{alegeq}
	A \leq &  \Big\|\prod_{k=0}^{K-1}(\mI - \alpha \nabla^2 \ell(\vw_k; \xi^s_i)) -\prod_{k=0}^{K-2}(\mI - \alpha \nabla^2 \ell(\vw_k; \xi^s_i))(\mI - \alpha \nabla^2 \ell(\vu_{K-1}; \xi^s_i))\Big\| \nonumber
	\\ &+\Big\| \prod_{k=0}^{K-2}(\mI - \alpha \nabla^2 \ell(\vw_k; \xi^s_i))(\mI - \alpha \nabla^2 \ell(\vu_{K-1}; \xi^s_i))-\prod_{k=0}^{K-1}(\mI - \alpha \nabla^2 \ell(\vu_k; \xi^s_i)))\Big\|\nonumber
	\\ \leq &\Big(   (1+\alpha   M)^{K-1}\alpha \rho  + \frac{\rho}{M} (1+\alpha M)^K \big( (1+\alpha M)^{K-1} -1 \big)\Big)\|\vw- \vu\|,
	\end{align}
	Combining (\ref{lopasv}) and (\ref{alegeq}) yields
	\begin{align}\label{inops}
	\|G_i(\vw) - G_i(\vu) \| \leq& \big(   (1+\alpha  M)^{K-1}\alpha \rho  + \frac{\rho}{M} (1+\alpha M)^K \big( (1+\alpha M)^{K-1} -1 \big)\big)\|\vw - \vu \|  \|\nabla
\ell(\vw_K; \xi^q_i)\| + (1+\alpha M)^K  M \| \vw_K-\vu_K \|.
	\end{align}
	To upper-bound $ \|\nabla
\ell(\vw_K; \xi^q_i)\| $ in (\ref{inops}),  using the mean value theorem, we have
	\begin{align}\label{lowni}
	\|\nabla
\ell(\vw_K; \xi^q_i)\|  = &  \Big\|\nabla \ell(\vw-\sum_{k=0}^{K-1}\alpha \nabla \ell(\vw_k; \xi^s_i); \xi^q_i)\Big\|\nonumber
	\\ \leq & \|\nabla \ell(\vw; \xi^q_i) \| + \alpha M \sum_{k=0}^{K-1} (1+\alpha L)^k\big\| \nabla \ell(\vw_k; \xi^s_i)  \big\| \nonumber
	\\ \leq & (1+\alpha M)^K  \|\nabla \ell(\vw; \xi^q_i)\|  + \big( (1+\alpha M)^K-1 \big)b_i,
	\end{align}
For $\|\vw_K - \vu_K\|$, we have:
	\begin{align}\label{wnos}
	\|\vw_K - \vu_K\| \leq (1+\alpha M)^K \|\vw - \vu\|.
	\end{align}
	Combining (\ref{inops}), (\ref{lowni}) and (\ref{wnos}) yields
	\begin{align*}
	\|G_i(\vw) & - G_i(\vu) \| 
	\\ \leq& \Big(   (1+\alpha  M)^{K-1}\alpha \rho  + \frac{\rho}{M} (1+\alpha M)^K \big( (1+\alpha M)^{K-1} -1 \big)\Big)(1+\alpha M)^K \|\nabla l_{T_i} (w)\|\|w-u\| \nonumber
	\\&+ \Big(   (1+\alpha M)^{K-1}\alpha \rho  + \frac{\rho}{M} (1+\alpha M)^K \big( (1+\alpha M)^{K-1} -1 \big)\Big)\big( (1+\alpha M)^K-1 \big)b_i\|w-u\| \nonumber
	\\ &+ (1+\alpha M)^{2K} M \| \vw - \vu\|,
	\end{align*}
	which yields 
	\begin{align*}
	\|G_i(\vw)  - G_i(\vu) \| \leq \big( (1+\alpha M)^{2K}M + C (b + \|\nabla
\ell(\vw; \xi^q_i)\|) \big) \|\vw - \vu\|.
	\end{align*}
	Based on the above inequality and  Jensen's inequality, we 
	finish the proof.

\subsection{Proof for Theorem~\ref{thm:conv}}
Despite Proposition~\ref{prop:lip}, 
we also need to upper-bound the expectation of 
$\mathbb{E}\| G_i(\vw)\|^2$, as follows:
\begin{lemma}\label{prop:2nd} 
Set $\alpha = \frac{1}{8KM}$.
we have
for any $\vw$,
	\begin{align*}
	\mathbb{E}\| G_i(\vw)\|^2 \leq A_{\text{\normalfont squ}_1} \|\nabla \mathcal{L}(\vw)\|^2 + A_{\text{\normalfont squ}_2},
	\end{align*}
where
$A_{\text{\normalfont squ}_1} = \frac{4(1+\alpha M)^{4K}}{(2-(1+\alpha
M)^{2K})^2}, 
A_{\text{\normalfont squ}_2} = \frac{4(1+\alpha M)^{8K}}{(2-(1+\alpha
M)^{2K})^2}(\sigma+b)^2 
+ 2(1+\alpha)^{4K}(\sigma^2+\tilde{b}^2)$, and
$\tilde{b}^2 = \frac{1}{|\gI|} \sum_{i \in \gI} b_i^2$. 
\end{lemma}

\begin{proof}
	Conditioning on $\vw$, we have 
	\begin{align*}
	\mathbb{E}\| G_i(\vw)\|^2 = &\mathbb{E} \Big\| \prod_{k=0}^{K-1}(\mI - \alpha \nabla^2 \ell(\vw_k; \xi^s_i))\nabla
\ell(\vw_K; \xi^q_i)  \Big\|^2 \leq  (1+\alpha M)^{2K} \mathbb{E} \|\nabla \ell(\vw_K; \xi^q_i) \|^2,
	\end{align*}
	Using an approach similar to (\ref{lowni}), we have:
	\begin{align}\label{gwkopo}
	\mathbb{E}\| G_i(\vw)\|^2 \leq&  (1+\alpha M)^{2K} 2(1+\alpha M)^{2K} \mathbb{E} \|\nabla \ell(\vw_K; \xi^q_i) \|^2 + 2(1+\alpha M)^{2K} \big( (1+\alpha M)^K -1\big)^2 \mathbb{E}_i b_i^2 \nonumber
	\\ \leq  & 2(1+\alpha M)^{4K} (\|\nabla \mathcal{L}(\vw)\|^2 + \sigma^2)+ 2(1+\alpha M)^{2K} \big( (1+\alpha M)^K -1\big)^2 \widetilde b \nonumber
	\\ \leq & 2(1+\alpha M)^{4K} \Big(  \frac{2}{C_1^2} \|\nabla \mathcal{L}(\vw)\|^2 + \frac{2C_2^2}{C_1^2} + \sigma^2 \Big) + 2(1+\alpha M)^{2K} \big( (1+\alpha M)^K -1\big)^2 \widetilde b^2 \nonumber
	\\ \leq & \frac{4(1+\alpha M)^{4K}}{C_1^2}\|\nabla \mathcal{L}(\vw)\|^2 + \frac{4(1+\alpha M)^{4K}C_2^2}{C_1^2} + 2(1+\alpha M)^{4K}(\sigma^2 + \widetilde b^2),
	\end{align}
	Noting that $C_2=\big( (1+\alpha M)^{2K}-1  \big)\sigma + (1+\alpha M)^K \big((1+\alpha M)^K -1 \big) b < \big( (1+\alpha M)^{2K}-1  \big)(\sigma +b)$ and using the definitions of $A_{\text{\normalfont squ}_1}, A_{\text{\normalfont squ}_2}$, we finish the proof. 
\end{proof}


Apart from these propositions, 
we also need some auxiliary lemmas to prove Theorem~\ref{thm:conv}. 
In the sequel, for any vector $\vv$, define $[\vv]^2$ as the vector whose elements are the squares of elements in $\vv$. 
\begin{lemma}
\label{lem:aux}
Suppose function $f: [0, +\infty) \to [0, +\infty)$ is a non-increasing function. 
Then for any sequence $a_0, \dots, a_T \ge 0$, we have:
\begin{align*}
\sum_{t=1}^T a_t \cdot f(a_0+\sum_{t=1}^T a_t) \le \int_{a_0}^{\sum_{t=0}^T a_t}
f(x) dx.
\end{align*}
\end{lemma}
\begin{proof}
Let $s_t = \sum_{u=0}^t a_u$. Since any $a_t \ge 0$ for $t = 0, \dots, T$, 
obviously we have $s_{t-1} \le s_t$, and $f(s_0) \ge f(s_1) \ge \dots \ge f(s_T)$. 
Therefore, we have:
\begin{align*}
a_t \cdot f(s_t) = \int_{s_{t-1}}^{s_t} f(s_t) dx \le \int_{s_{t-1}}^{s_t} f(x)
dx.
\end{align*}
Summing from $t=1$ to $T$ gives the result. 
\end{proof}


\begin{lemma}
\label{lem:n1}
Let $\{ \vw_t \}$ be the sequence of model weights generated from Algorithm~\ref{alg:meta} with $K=1$. 
Then for
$A_t = \E \| \frac{[\vm_t]^2}{(\vv_t + \delta \vone)} \|_1$, 
we have:
\begin{align*}
\sum_{t=1}^T A_t \le \sum_{t=1}^T \E \Big\| \frac{[\vg_t]^2}{(\vv_t + \delta \vone)}
\Big\|_1.
\end{align*}
\end{lemma}
\begin{proof}[Proof for Lemma~\ref{lem:n1}]
We first define $\gT_i = \{t:i_t = i \}$ for each client $i$. 
Intuitively, this set counts the iterations where client $i$ is visited, 
and obviously we have 
\begin{align*}
\gT_i \cap \gT_j = \Phi, i \ne j, \\
\cup_{i \in \gI} \gT_i = \{0, \dots, T-1 \}.
\end{align*}
For each iteration $t$, consider set $\gT_{i_t} = \{ t_0, t_1, \dots \}$. We can
express $\vm_t$ as $\vm_t = (1-\theta) \sum_{j: t_j \in \gT_{i_t}} \theta^{j}
\vg_{t_j}$, and we have:
\begin{align*}
\Big\| \frac{[\vm_t]^2}{(\vv_t + \delta \vone)} \Big\|_1 
= \sum_{k=1}^d \Big\| \frac{\vm_{t,k}}{(\vv_{t,k} + \delta)^{1/2}} \Big \|^2 \le \sum_{k=1}^d (1-\theta)^2 \Big\|
\sum_{j: t_j \in \gT_{i_t}} \frac{\theta^{j} \vg_{t_j, k}}{(\vv_{t,k} +
\delta)^{1/2}} \Big\|^2.
\end{align*}
Using Cauchy's inequality $(\sum_{j=1}^k a_j b_j)^2 \le (\sum_{j=1}^k a_j^2)(\sum_{j=1}^k b_j^2)$, 
from 
\[ \frac{\theta^{j} \vg_{t_j, k}}{(\vv_{t,k} + \delta)^{1/2}} =
\frac{\theta^{j/2} \vg_{t_j, k}}{(\vv_{t,k} + \delta)} \cdot
\theta^{j/2}(\vv_{t,k} + \delta)^{1/2}, \]
we can bound it as: 
\begin{align*}
\Big\| \frac{[\vm_t]^2}{(\vv_t + \delta \vone)} \Big\|_1 \le & \sum_{k=1}^d (1-\theta)^2
\left(\sum_{j: t_j \in \gT_{i_t}} \theta^{j} (\vv_{t,k} + \delta) \right)
\left(\sum_{j: t_j \in \gT_{i_t}} \frac{\theta^{j} \vg^2_{t_j, k}}{(\vv_{t,k} +
\delta)^2} \right).
\end{align*}
Since $\theta \in (0,1)$, we always have $\sum_{t=0}^{T} \theta^t < \frac{1}{1-\theta}$ for any $T \ge 0$. 
Then we have:
\begin{align*}
\Big\| \frac{[\vm_t]^2}{(\vv_t + \delta \vone)} \Big\|_1 \le & \sum_{k=1}^d (1-\theta)
\sum_{j: t_j \in \gT_{i_t}} \frac{\theta^{j} \vg^2_{t_j, k}}{(\vv_{t,k} + \delta)} =
(1-\theta) \sum_{j: t_j \in \gT_{i_t}}\theta^{j} \Big\| \frac{ [\vg_{t_j}]^2}{(\vv_{t} +
\delta)} \Big\|_1.
\end{align*}
Note that $t_j \le t$ by definition, and each element of $\vv_t$ is non-decreasing with $t$ since $\vv_{t_{j+1}} - \vv_{t_{j}} = [\vg_{t_{j}}]^2 \ge 0$ for all $t_j$. 
As such, we have:
\begin{align*}
\| \frac{[\vm_t]^2}{(\vv_t + \delta \vone)} \|_1 \le (1-\theta) \sum_{t_j \in
\gT_{i_t}}\theta^{j} \| \frac{ [\vg_{t_j}]^2}{(\vv_{t} + \delta)} \|_1 \le
(1-\theta) \sum_{j: t_j \in \gT_{i_t}}\theta^{j} \| \frac{[\vg_{t_j}]^2}{(\vv_{t_j} +
\delta)} \|_1.
\end{align*}

\begin{table*}[t]
    \centering
    \caption{Computation procedure for the local momentum. Entries indicate the coefficient on historical gradients to compute the local momentum. }
    \label{tab:con}
    \begin{tabular}{c | c c c c c}
    \hline
     & $\vg_{T-1}$ & $\vg_{T-2}$ & $\dots$ & $\vg_{T'}$ \\
    \hline
    $\vm_{T-1}$ & $1-\theta$ & 0 & $0 \dots 0$ & $(1-\theta)\theta$ \\
    $\vm_{T-2}$ & 0 & $1-\theta$ & $\dots$ & 0 & \\
    $\cdots$ & 0 & 0 & $\dots$ & 0 \\
    $\vm_{T'}$ & 0 & 0 & $\dots$ & $1-\theta$ \\
    \hline
    \end{tabular}
\end{table*}

Then sum from $t=0$ to $T-1$, and from Table~\ref{tab:con},
we obtain:
\begin{align*}
\sum_{t=0}^{T-1} \Big\| \frac{[\vm_t]^2}{(\vv_t + \delta \vone)} \Big\|_1 \le &  (1-\theta)
\sum_{t=0}^{T-1} \sum_{j: t_j \in \gT_{i_t}} \theta^{j} \Big\| \frac{
[\vg_{t_j}]^2}{(\vv_{t_j} + \delta)} \Big\|_1 \le \sum_{t=0}^{T-1} \Big\| \frac{
[\vg_{t}]^2}{(\vv_{t} + \delta)} \Big\|_1,
\end{align*}
which concludes the proof. 

%

\end{proof}

\begin{lemma}
\label{lem:n2}
Let $\{ \vw_t \}$ be generated from Algorithm~\ref{alg:meta}. Define
\begin{eqnarray*}
A_t & = & \begin{cases}
\E \Big\| \frac{[\vm_t]^2}{(\vv_t + \delta \vone)^{1/2}} \Big\|_1 & t \ge -1 \\
0 & t < -1
\end{cases}, \\
B_t & = & -\E \langle \nabla \Ls(\vw_t), \frac{\vm_t}{(\vv_t+\delta \vone)^{1/2}}
\rangle, \\
C_t & = & \eta \theta A_{t-1} + (1-\theta)\eta^2 M_{meta} N \sum_{h=1}^N A_{t-h} +
2(1-\theta)M_{meta}^2\beta(N).
\end{eqnarray*}
Further, define $\tau(t, i)$ to be the last iteration before iteration $t$ when worker $i$ is visited. 
Specifically, $\tau(0, i)=-1$ for all $i$. 
We have:
\begin{align*}
B_t + (1-\theta)\E(\Ls(\vw_t) - \Ls^*) \le \theta B_{\tau(t, i_t)}+C_t.
\end{align*}
\end{lemma}

\begin{proof}[Proof for Lemma~\ref{lem:n2}]
We first consider bounding a related term $\E \langle - \frac{\nabla \Ls(\vw_t)}{(\vv_t+\delta \vone)^{1/2}}, \vg_t \rangle = -\E \langle \frac{\nabla \Ls(\vw_t)}{(\vv_t+\delta \vone)^{1/2}}, \nabla \ell(\vw_t, \xi_t) \rangle$. We have:
\begin{eqnarray*}
\vg_t & = & \nabla \Ls(\vw_t) - \nabla \Ls(\vw_t) + \nabla \Ls(\vw_{t-N}) - \nabla
\Ls(\vw_{t-N}) + \nabla \ell(\vw_{t-N}, \xi_t) - \nabla \ell(\vw_{t-N}, \xi_t) \\
& & + \nabla \ell(\vw_t, \xi_t).
\end{eqnarray*}
Then,
\begin{align*}
\E \langle - \frac{\nabla \Ls(\vw_t)}{(\vv_t+\delta \vone)^{1/2}}, \vg_t \rangle = & - \E \| \frac{[\nabla \Ls(\vw_t)]^2}{(\vv_t+\delta \vone)^{1/2}} \|_1 + \E \langle \frac{\nabla \Ls(\vw_t)}{(\vv_t+\delta \vone)^{1/2}}, \nabla \Ls(\vw_t) - \nabla \Ls(\vw_{t-N}) \rangle \\
& + \E \langle  \frac{\nabla \Ls(\vw_t)}{(\vv_t+\delta \vone)^{1/2}}, \nabla \Ls(\vw_{t-N}) - \nabla \Ls(\vw_{t-N}, \xi_t) \rangle \\
& + \E \langle  \frac{\nabla \Ls(\vw_t)}{(\vv_t+\delta \vone)^{1/2}}, \nabla
\ell(\vw_{t-N}, \xi_t) - \nabla \ell(\vw_t, \xi_t) \rangle.
\end{align*}
The second term can be bounded using Proposition~\ref{prop:lip} as:
\begin{align}
\E \langle \frac{\nabla \Ls(\vw_t)}{(\vv_t+\delta \vone)^{1/2}}, \nabla \Ls(\vw_t) - \nabla \Ls(\vw_{t-N}) \rangle \le & \frac{M_{meta}}{\delta^{1/4}} \E \| \frac{\nabla \Ls(\vw_t)}{(\vv_t+\delta \vone)^{1/4}} \| \| \vw_{t-N} - \vw_t \| \notag \\
\le & \frac{M_{meta}}{\delta^{1/4}} \sum_{h=1}^N \E \| \frac{\nabla \Ls(\vw_t)}{(\vv_t+\delta \vone)^{1/4}} \| \| \vw_{t-h+1} - \vw_{t-h} \| \notag \\
\le & \frac{\eta M_{meta}}{\delta^{1/2}} \sum_{h=1}^N \E \| \frac{\nabla
\Ls(\vw_t)}{(\vv_t+\delta \vone)^{1/4}} \| \|  \frac{\vm_{t-h}}{(\vv_{t-h}+\delta
\vone)^{1/4}} \|. \label{eq:n1}
\end{align}
With Cauchy's inequality, we have
\begin{align*}
\| \frac{\nabla \Ls(\vw_t)}{(\vv_t+\delta \vone)^{1/4}} \| \|  \frac{\vm_{t-h}}{(\vv_{t-h}+\delta \vone)^{1/4}} \| \le \frac{1}{2} ( \alpha \| \frac{[\nabla \Ls(\vw_t)]^2}{(\vv_t+\delta \vone)^{1/2}} \| +  \frac{1}{\alpha} \| \frac{[\vm_{t-h}]^2}{(\vv_{t-h}+\delta \vone)^{1/2}} \| ),
\end{align*}
where $\alpha > 0$ is arbitrary. 
Combining it with (\ref{eq:n1}), we have:
\begin{eqnarray*}
\lefteqn{\E \left\langle \frac{\nabla \Ls(\vw_t)}{(\vv_t+\delta \vone)^{1/2}}, \nabla
\Ls(\vw_t) - \nabla \Ls(\vw_{t-N}\right\rangle } \\
& \le & \frac{\eta M_{meta}}{\delta^{1/2}} \sum_{h=1}^N \E \| \frac{\nabla \Ls(\vw_t)}{(\vv_t+\delta \vone)^{1/4}} \| \|  \frac{\vm_{t-h}}{(\vv_{t-h}+\delta \vone)^{1/4}} \| \\
& \le & \frac{\eta M_{meta}}{2 \alpha \delta^{1/2}} \sum_{h=1}^N \E \|  \frac{[\vm_{t-h}]^2}{(\vv_{t-h}+\delta \vone)^{1/2}} \| + \frac{\alpha \eta M_{meta} T}{2 \delta^{1/2}} \E \| \frac{[\nabla \Ls(\vw_t)]^2}{(\vv_t+\delta \vone)^{1/2}} \|.
\end{eqnarray*}
Now choose $\alpha=\frac{\delta^{1/2}}{2\eta MT}$, we have:
\begin{align*}
\E \langle \frac{\nabla \Ls(\vw_t)}{(\vv_t+\delta \vone)^{1/2}}, \nabla \Ls(\vw_t) - \nabla \Ls(\vw_{t-N} \rangle 
\le & \frac{\eta^2 M_{meta}^2 T}{\delta} \sum_{h=1}^N \E \|  \frac{[\vm_{t-h}]^2}{(\vv_{t-h}+\delta \vone)^{1/2}} \| + \frac{1}{4} \E \| \frac{[\nabla \Ls(\vw_t)]^2}{(\vv_t+\delta \vone)^{1/2}} \|.
\end{align*}
The third term can be bounded as:
\begin{align*}
\E \langle \frac{\nabla \Ls(\vw_t)}{(\vv_t+\delta \vone)^{1/2}}, \nabla
\Ls(\vw_{t-N}) - \nabla \Ls(\vw_{t-N}, \xi_t) \rangle \le G^2 \beta(N).
\end{align*}
The bound for the last term is very similar to the second term, and we have:
\begin{align*}
\E \langle \frac{\nabla \Ls(\vw_t)}{(\vv_t+\delta \vone)^{1/2}}, \nabla \ell(\vw_{t-N}, \xi_t) - \nabla \ell(\vw_t, \xi_t) \rangle \le & \frac{M_{meta}}{\delta^{1/4}} \E \| \frac{\nabla \Ls(\vw_t)}{(\vv_t+\delta \vone)^{1/4}} \| \| \vw_{t-N} - \vw_t \| \\
\le & \frac{M_{meta}}{\delta^{1/4}} \sum_{h=1}^N \E \| \frac{\nabla \Ls(\vw_t)}{(\vv_t+\delta \vone)^{1/4}} \| \| \vw_{t-h+1} - \vw_{t-h} \| \\
\le & \frac{\eta M_{meta}}{\delta^{1/2}} \sum_{h=1}^N \E \| \frac{\nabla \Ls(\vw_t)}{(\vv_t+\delta \vone)^{1/4}} \| \|  \frac{\vm_{t-h}}{(\vv_{t-h}+\delta \vone)^{1/4}} \|,
\end{align*}
which is exactly the same as (\ref{eq:n1}). Hence, we have:
\begin{eqnarray*}
\lefteqn{\E \langle \frac{\nabla \Ls(\vw_t)}{(\vv_t+\delta \vone)^{1/2}}, \nabla
\ell(\vw_{t-N}, \xi_t) - \nabla \ell(\vw_t, \xi_t) \rangle } \\
& \le & \frac{\eta^2 M_{meta}^2 T}{\delta} \sum_{h=1}^N \E \|  \frac{[\vm_{t-h}]^2}{(\vv_{t-h}+\delta \vone)^{1/2}} \| + \frac{1}{4} \E \| \frac{[\nabla \Ls(\vw_t)]^2}{(\vv_t+\delta \vone)^{1/2}} \|.
\end{eqnarray*}
Combining them together gives:
\begin{align}
\E \langle - \frac{\nabla \Ls(\vw_t)}{(\vv_t+\delta \vone)^{1/2}}, \vg_t \rangle
\le - \frac{1}{2} \E \| \frac{[\nabla \Ls(\vw_t)]^2}{(\vv_t+\delta \vone)^{1/2}}
\|_1 + \frac{2\eta^2 M_{meta}^2 T}{\delta} \sum_{h=1}^N \E \|
\frac{[\vm_{t-h}]^2}{(\vv_{t-h}+\delta \vone)^{1/2}} \| + G^2\beta(N). \label{eq:n2}
\end{align}
Now for $B_t = -\E \langle \nabla \Ls(\vw_t), \frac{\vm_t}{(\vv_t+\delta \vone)^{1/2}} \rangle$, consider the expectation conditioned on $\chi_{t}$, we have:
\begin{eqnarray*}
\lefteqn{\E[\langle - \nabla \Ls(\vw_t), \frac{\vm_t}{(\vv_t+\delta \vone)^{1/2}}
\rangle|\chi_{t'}]} \\
& = & \E[\langle \frac{-\nabla \Ls(\vw_t)}{(\vv_t+\delta \vone)^{1/2}}, \theta \vm^{i_t}_{t-1} + (1-\theta)\vg_t \rangle|\chi_{t'}] \\
&= & (1-\theta) \E[\langle - \nabla \Ls(\vw_t), \frac{\vg_t}{(\vv_t+\delta \vone)^{1/2}} \rangle|\chi_{t'}] + \theta \E[\langle \frac{-\nabla \Ls(\vw_t)}{(\vv_t+\delta \vone)^{1/2}}, \vm^{i_t}_{t-1} \rangle|\chi_{t'}] \\
&= & (1-\theta) \E[\langle - \nabla \Ls(\vw_t), \frac{\vg_t}{(\vv_t+\delta \vone)^{1/2}} \rangle|\chi_{t'}] + \theta \langle \frac{- \nabla \Ls(\vw_{\tau(t, i)})}{(\vv_{\tau(t, i)}+\delta \vone)^{1/2}}, \vm_{\tau(t, i)} \rangle \\
&& + \theta \langle \frac{\nabla \Ls(\vw_{\tau(t, i)}) - \nabla \Ls(\vw_t)}{(\vv_{\tau(t, i)}+\delta \vone)^{1/2}}, \vm_{\tau(t, i)} \rangle \\
&& + \theta \langle \frac{\nabla \Ls(\vw_t)}{(\vv_{\tau(t, i)}+\delta
\vone)^{1/2}} - \frac{\nabla \Ls(\vw_t)}{(\vv_{t}+\delta \vone)^{1/2}},
\vm_{\tau(t, i)} \rangle.
\end{eqnarray*}
The first term has been bounded by (\ref{eq:n2}), 
and the third term can be bounded from Proposition~\ref{prop:lip} as:
\begin{align*}
\langle \frac{\nabla \Ls(\vw_{\tau(t, i)}) - \nabla \Ls(\vw_t)}{(\vv_{\tau(t, i)}+\delta \vone)^{1/2}}, \vm_{\tau(t, i)} \rangle \le &  \| {\nabla \Ls(\vw_{\tau(t, i)}) - \nabla \Ls(\vw_t)} \| \|\frac{\vm_{\tau(t, i)}}{(\vv_{\tau(t, i)}+\delta \vone)^{1/2}} \| \\
= & \eta M_{meta} \| \sum_{u = \tau(t, i)}^t \frac{\vm_{u}}{(\vv_{u}+\delta \vone)^{1/2}} \| \| \frac{\vm_{\tau(t, i)}}{(\vv_{\tau(t, i)}+\delta \vone)^{1/2}} \| \\
\le & \eta M_{meta} \sum_{u = \tau(t, i)}^t \|  \frac{\vm_{u}}{(\vv_{u}+\delta \vone)^{1/2}} \| \| \frac{\vm_{\tau(t, i)}}{(\vv_{\tau(t, i)}+\delta \vone)^{1/2}} \| \\
\le & \frac{\eta M_{meta} }{2} \sum_{u = \tau(t, i)}^t \left( \|  \frac{\vm_{u}}{(\vv_{u}+\delta \vone)^{1/2}} \|^2 + \| \frac{\vm_{\tau(t, i)}}{(\vv_{\tau(t, i)}+\delta \vone)^{1/2}} \|^2 \right).
\end{align*}
Finally, the last term can be bounded by Proposition~\ref{prop:2nd} as:
\begin{align*}
\langle \frac{\nabla \Ls(\vw_t)}{(\vv_{\tau(t, i)}+\delta \vone)^{1/2}} -
\frac{\nabla \Ls(\vw_t)}{(\vv_{t}+\delta \vone)^{1/2}}, \vm_{\tau(t, i)} \rangle
\le G^2 (\| \frac{\vone}{(\vv_{\tau(t, i)}+\delta \vone)^{1/2}} \| - \|
\frac{\vone}{(\vv_{t}+\delta \vone)^{1/2}} \|).
\end{align*}
Now taking expectation on both sides gives:
\begin{eqnarray*}
B_t & \le & (1-\theta)(- \frac{1}{2} \E \| \frac{[\nabla \Ls(\vw_t)]^2}{(\vv_t+\delta \vone)^{1/2}} \|_1 + \frac{2\eta^2 M^2 T}{\delta} \sum_{h=1}^N \E \|  \frac{[\vm_{t-h}]^2}{(\vv_{t-h}+\delta \vone)^{1/2}} \| + G^2\beta(N)) + \theta B_{\tau(t, i_t)} \\
&& + \frac{\eta M_{meta} \theta}{2} \sum_{u = \tau(t, i_t)}^t \left( \E \|
\frac{\vm_{u}}{(\vv_{u}+\delta \vone)^{1/2}} \|^2 + \E \| \frac{\vm_{\tau(t,
i)}}{(\vv_{\tau(t, i)}+\delta \vone)^{1/2}} \|^2 \right) \\
&& + \theta G^2 (\E \| \frac{\vone}{(\vv_{\tau(t, i_t)}+\delta \vone)^{1/2}} \| - \E \| \frac{\vone}{(\vv_{t}+\delta \vone)^{1/2}} \|) \\
&= & (1-\theta)(- \frac{1}{2} \E \| \frac{[\nabla \Ls(\vw_t)]^2}{(\vv_t+\delta \vone)^{1/2}} \|_1 + \frac{2\eta^2 M_{meta}^2 T}{\delta} \sum_{h=1}^N A_{t-h} + G^2\beta(N)) + \theta B_{\tau(t, i_t)} \\
&& + \frac{\eta M_{meta} \theta}{2} \sum_{u = \tau(t, i_t)}^t \left( A_u + A_{\tau(t,
i_t)} \right) + \theta G^2 (\E \| \frac{\vone}{(\vv_{\tau(t, i)}+\delta
\vone)^{1/2}} \| - \E \| \frac{\vone}{(\vv_{t}+\delta \vone)^{1/2}} \|).
\end{eqnarray*}
Rearranging these terms gives:
\begin{eqnarray*}
\lefteqn{B_t +  \frac{1-\theta}{2} \E \| \frac{[\nabla
\Ls(\vw_t)]^2}{(\vv_t+\delta \vone)^{1/2}} \|_1 } \\
& \le & (1-\theta)(\frac{2\eta^2 M_{meta}^2 T}{\delta} \sum_{h=1}^N A_{t-h} + G^2\beta(N)) + \theta B_{\tau(t, i)} \\
& & + \frac{\eta M_{meta} \theta}{2} \sum_{u = \tau(t, i_t)}^t \left( A_u + A_{\tau(t,
i_t)} \right) + \theta G^2 (\| \frac{\vone}{(\vv_{\tau(t, i)}+\delta \vone)^{1/2}}
\| - \| \frac{\vone}{(\vv_{t}+\delta \vone)^{1/2}} \|),
\end{eqnarray*}
which is exactly Lemma~\ref{lem:n2}. 
\end{proof}
Now we are ready to present the proof for the main theorem:
\begin{proof}[Proof for Theorem~\ref{thm:conv}]
First from Lemma~\ref{lem:n2}, we have:
\begin{eqnarray*}
B_T + \frac{(1-\theta)}{2} \E \| \frac{[\nabla \Ls(\vw_T)]^2}{(\vv_T+\delta
\vone)^{1/2}} \|_1 & \le & \theta B_{\tau(T, i_T)}+C_T, \\
B_{T-1} - \frac{(1-\theta)}{2} \E \| \frac{[\nabla
\Ls(\vw_{T-1})]^2}{(\vv_{T-1}+\delta \vone)^{1/2}} \|_1 & \le & \theta B_{\tau(T-1,
i_{T-1})}+C_{T-1}, \\
\vdots \\
B_{1}- \frac{(1-\theta)}{2} \E \| \frac{[\nabla \Ls(\vw_1)]^2}{(\vv_1+\delta
\vone)^{1/2}} \|_1 & \le & \theta B_{\tau(1, i_{1})}+C_{1}.
\end{eqnarray*}
Summing all the above gives us:
\begin{align}
\frac{(1-\theta)}{2}\sum_{t=1}^T \E \Big\| \frac{\nabla \Ls(\vw_t)}{(\vv_t+\delta \vone)^{1/4}} \Big\|^2 \le &  -\sum_{i=1}^n B_{\tau(T, i)}+(\theta-1) \sum_{t=0}^{T-1}
B_{t}+\sum_{t=1}^{T} C_t,
\label{eq:n3}
\end{align}
where we note that all $B$ terms on the right hand side must have a correspondence on left hand. 
Then from Proposition~\ref{prop:2nd}, for any client $i$, we have:
\begin{align*}
-B_T = \E \langle \nabla \Ls(\vw_t), \frac{\vm_t}{(\vv_t+\delta \vone)^{1/2}}
\rangle \le \E \| \nabla \Ls(\vw_t) \| \| \frac{\vm_t}{(\vv_t+\delta \vone)^{1/2}}
\| \le \frac{G^2}{\sqrt{\delta}}.
\end{align*}
Since the loss function is smooth (Proposition~\ref{prop:lip}), we have:
\begin{eqnarray*}
\Ls(\vw_{t+1}) - \Ls(\vw_t) & \le & \langle \nabla \Ls(\vw_t), \vw_{t+1}-\vw_t \rangle + \frac{M_{meta}}{2} \| \vw_{t+1}-\vw_t \|^2 \\
& = & \eta \langle \nabla \Ls(\vw_t),  \frac{\vm_t}{(\vv_t+\delta \vone)^{1/2}}
\rangle + \frac{\eta^2 M_{meta}}{2} \|  \frac{\vm_t}{(\vv_t+\delta \vone)^{1/2}} \|^2.
\end{eqnarray*}
Taking expectation on both sides gives $\E[\Ls(\vw_{t+1}) - \Ls(\vw_t)] \le \eta B_t + \frac{\eta^2 M_{meta}}{2} A_t$. 
Then summing from $t=0$ to $T-1$ gives:
\begin{eqnarray*}
\E \Ls(\vw_T) - \E \Ls(\vw_0) & \le & \eta \sum_{t=0}^{T-1} B_t + \frac{\eta^2
M_{meta}}{2} \sum_{t=0}^{T-1} A_t, \\
- \sum_{t=0}^{T-1} B_t & \le & \frac{1}{\eta}(\E \Ls(\vw_0) - \E \Ls(\vw_T)) +
  \frac{\eta M_{meta}}{2} \sum_{t=0}^{T-1} A_t \le \frac{1}{\eta} \E \Ls(\vw_0) +
  \frac{\eta M_{meta}}{2} \sum_{t=0}^{T-1} A_t.
\end{eqnarray*}
For the last term, we have:
\begin{align*}
\sum_{t=1}^{T} C_t \le & (1-\theta) \sum_{t=1}^T (\frac{2\eta^2 K M_{meta}^2}{\delta} \sum_{h=1}^K A_{t-h} + G^2\beta(N)) \\
& + \frac{\eta M_{meta} \theta}{2} \sum_{t=1}^T  \sum_{u = \tau(t, i_t)}^t \left( A_u + A_{\tau(t, i_t)} \right) + \theta G^2 \sum_{t=1}^T  (\| \frac{\vone}{(\vv_{\tau(t, i)}+\delta \vone)^{1/2}} \| - \| \frac{\vone}{(\vv_{t}+\delta \vone)^{1/2}} \|) \\
\le &  \frac{2\eta^2 K^2 M_{meta}^2(1-\theta)}{\delta} \sum_{t=1}^T A_{t} + (1-\theta) G^2 T \beta(N) + \eta M_{meta} \theta n \sum_{t=1}^T  A_t + \frac{n \theta G^2}{\sqrt{\delta}}.
\end{align*}
Combined with (\ref{eq:n3}), 
we have:
\begin{eqnarray*}
\lefteqn{\sum_{t=1}^T \E \Big\| \frac{\nabla \Ls(\vw_t)}{(\vv_t+\delta \vone)^{1/4}} \Big\|^2} \\
& \le &  \frac{2n (1+\theta) G^2}{(1-\theta)\sqrt{\delta}}+ \frac{2}{\eta} \E \Ls(\vw_0) + (\eta M_{meta} +\frac{4\eta^2 K^2 M_{meta}^2}{\delta}  + \frac{2\eta M_{meta} \theta n}{1-\theta}) \sum_{t=0}^{T-1} A_t +  2G^2 T \beta(N).
\end{eqnarray*}
For $\sum_{t=0}^{T_1} A_t$, first from Lemma, we have 
\begin{align*}
\sum_{t=0}^{T-1} A_t \le \sum_{t=0}^{T-1} \E \| \frac{[\vg_t]^2}{(\vv_t + \delta \vone)} \|_1.
\end{align*} 
Then using Lemma~\ref{lem:aux} with $f(x) = \frac{1}{x}$, we have:
\begin{align*}
\sum_{j: t_j \in \gT_{i_t}} \| \frac{ \vg^2_{t_j}}{(\vv_{t_j} + \delta)} \|_1 \le \log(\frac{M^2 T +\delta}{\delta})
\end{align*}
Combine it with, we have 
\begin{align*}
\sum_{t=0}^{T-1} A_t \le n \log(\frac{M^2 T +\delta}{\delta})
\end{align*}
Finally, note that
\begin{align*}
\sum_{t=1}^T \E \Big\| \frac{\nabla \Ls(\vw_t)}{(\vv_t+\delta \vone)^{1/4}} \Big\|^2 \ge
(\sum_{t=1}^T \frac{1}{t^{1/2} \sqrt{C}}) \min_{1 \le t \le T} \E \| \nabla
\Ls(\vw_t) \|^2.
\end{align*}
We introduce the following auxiliary variables:
\begin{align*}
c_4(\theta) = &  \frac{2n (1+\theta) G^2 \sqrt{C} }{(1-\theta)\sqrt{\delta}}+
\frac{2 \sqrt{C}}{\eta} \E \Ls(\vw_0), \\
c_5(T, \theta) = &  (\eta M_{meta} n + \frac{2\eta M_{meta} \theta n^2}{1-\theta})
\log(\frac{M_{meta}^2 T +\delta}{\delta}), \\
c_6(T, K) = & \frac{4\eta^2 K^2 M_{meta}^2 n}{\delta} \log(\frac{M^2 T +\delta}{\delta})
+  2G^2 T \beta(N).
\end{align*}
We have:
\begin{align*}
\min_{1 \le t \le T} \E \| \nabla \Ls(\vw_t) \|^2 \le \frac{c_4(\theta) + c_5(T,
\theta) + c_6(T, K)}{T^{1/2}}.
\end{align*}
Let $\eta = \min \{ \frac{1}{nK}, 1 \}$, we obtain:
\begin{align*}
\min_{1 \le t \le T} \E \| \nabla \Ls(\vw_t) \|^2 =
O(\frac{n(K+1)}{T^{1/2}}+T^{1/2}\beta(N)).
\end{align*}
Let $\frac{n(K+1)}{T^{1/2}} = O(\epsilon)$ and $T^{1/2}\beta(N) = O(\epsilon)$ gives:
\begin{align*}
N = & \min \{ \frac{\log(1/\eps)}{\log(1/\sigma_2(\mP))}, 1\}, \\
\eta = & \min \{ \frac{\log(1/\sigma_2(\mP))}{n \log(1/\eps)}, 1\}, \\
T = & O\left(\max \{ \frac{n}{\eps^{2}
[\log(1/\sigma_2(\mP))]^{2}}, \frac{n}{\eps^{2}} \} \right),
\end{align*}
which completes our proof. 
\end{proof}

\subsection{Proof of Theorem~\ref{thm:iteration}}
\label{sgdproof}

\begin{proof}

The proof is similar to the proof in~\cite{cyffers2022privacy}, 
which tracks privacy loss using Rényi
Differential Privacy (RDP)~\cite{RDP} 
and leverages results on amplification by iteration~\cite{amp_iter}. 
We first recall the definition of RDP and the main theorems that we will use. 
Then, we apply these tools to our setting and conclude by translating the resulting RDP bounds into $(\eps, \delta)$-DP.


\begin{definition}[Rényi divergence~\cite{renyi1961measures, RDP}]
Let $1<\alpha<\infty$ and $\mu, \nu$ be measures such that for all measurable set $A$, $\mu(A)=0$ implies $\nu(A)=0$. The Rényi divergence of order $\alpha$ between $\mu$ and $\nu$ is defined as
\[
D_{\alpha}(\mu \| \nu) = \frac{1}{\alpha-1} \ln \int\left(\frac{\mu(z)}{\nu
(z)}\right)^{\alpha} \nu(z) d z.
\]
\end{definition}


\begin{definition}[Rényi DP~\cite{RDP}]
For $1 < \alpha \leq \infty$ and $\eps \geq 0$, a randomized algorithm $
\mathcal{A}$ satisfies $(\alpha, \eps)$-Rényi differential privacy, or $
(\alpha, \eps)$-RDP, if for all neighboring data sets $D$ and $D'$ we
have
\[
D_{\alpha}\left(\mathcal{A}(D) \| \mathcal{A}\left(D'\right)\right)
\leq \eps.
\]
\end{definition}

Similar to network DP, the definition of \emph{Network-RDP}~\cite{cyffers2022privacy} can also be introduced as follows:

\begin{definition}[Network Rényi DP~\cite{cyffers2022privacy}]
For $1 < \alpha \leq \infty$ and $\eps \geq 0$, a randomized algorithm $
\mathcal{A}$ satisfies $(\alpha, \eps)$-network Rényi differential privacy,
or
$(\alpha, \eps)$-NRDP, if for all pairs of
  distinct users $u, v\in V$ and all
  pairs   of neighboring datasets $D \sim_u D'$, we have
\[
D_{\alpha}\left(\mathcal{O}_{v}(\mathcal{A}(D)) \| \mathcal{O}_{v}(\mathcal{A}\left(D')\right)\right)
\leq \eps.
\]
\end{definition}


This proposition will be used in later proofs 
to analyze the privacy properties for the composition of different messages:
\begin{proposition}[Composition of RDP~\cite{RDP}]
\label{prop:comp}
If $\mathcal{A}_{1}, \ldots, \mathcal{A}_{k}$ are randomized algorithms
satisfying $(\alpha, \eps_{1})\text{-RDP}, \ldots,(\alpha, \eps_{k})$-RDP
respectively, then their composition $(\mathcal{A}_{1}(S), \ldots, 
\mathcal{A}_{k}(S))$ satisfies $(\alpha, \sum_{l=1}^k \eps_l)$-RDP. Each
algorithm can be chosen adaptively, i.e., based on the outputs of
algorithms that come before it.
\end{proposition}

We can also translate the result of the RDP by the following proposition~\cite{RDP}.

\begin{proposition}[Conversion from RDP to DP~\cite{RDP}]
\label{prop:convert}
If $\mathcal{A}$ satisfies $(\alpha, \eps)$-Rényi differential privacy, then for all $\delta \in(0,1)$ it also satisfies $\left(\eps+\frac{\ln (1 / \delta)}{\alpha-1}, \delta\right)$ differential privacy.
\end{proposition}


In our context, we aim to leverage this result to capture the
privacy amplification 
since a given user $v$ will only
observe
information about the update of another user $u$
after some steps of the random walk. To account for the fact that this number
of steps will itself be random, we will use the so-called weak convexity
property of the Rényi divergence~\cite{amp_iter}.

\begin{proposition}[Weak convexity of Rényi divergence~\cite{amp_iter}]
\label{prop:convexity}
Let $\mu_{1}, \ldots, \mu_{m}$ and $\nu_{1}, \ldots, \nu_{m}$ be probability
distributions over some domain $\mathcal{Z}$ such that for all $i \in[m], D_
{\alpha}\left(\mu_{i} \| \nu_{i}\right) \leq c /(\alpha-1)$ for some $c \in 
(0,1]$. Let $\rho$ be a probability distribution over $[m]$ and denote by
$\mu_{\rho}$ (resp. $\nu_{\rho}$) the probability distribution over $
\mathcal{Z}$ obtained by sampling i from $\rho$ and then outputting a random
sample from $\mu_{i}$ (resp. $\nu_{i}$). Then we have:
\[
D_{\alpha}\left(\mu_{\rho} \| \nu_{\rho}\right) \leq(1+c) \cdot \underset{i \sim \rho}{\E}\left[D_{\alpha}\left(\mu_{i} \| \nu_{i}\right)\right].
\]
\end{proposition}

We now have all the technical tools needed to prove our result.
Let us denote by $\sigma^2=\frac{8 M_{meta}^2 \ln(1.25/\delta)}{\eps^2}$ the variance
of the Gaussian noise added at each gradient step in Algorithm~\ref{alg:meta}.
Let us fix two distinct users $u$ and $v$. We aim to quantify how much
information about the private data of user $u$ is leaked to $v$ from
the visits of the token. 
Let us fix a contribution of user $u$ at some time $t_1$. 
Note that the token values observed before $t_1$ do not depend on the
contribution of $u$ at time $t_1$. Let $t_2>t_1$ be the first time that $v$
receives the token posterior to $t_1$. It is sufficient to bound the privacy
loss induced by the observation of the token at $t_2$: indeed, by the
post-processing property of DP, no additional privacy loss with respect to
$v$ will occur for observations posterior to $t_2$.
If there is no time
$t_2$ (which can be seen as $t_2>T$), then no privacy loss occurs. Let $Y_v$
and
$Y_v$ be the distribution followed by the token when observed by $v$ at
time $t_2$
for two neighboring datasets $D \sim_u D'$ which only differ in the dataset of
user $u$. For any $t$, let also $X_t$ and $X'_t$ be the
distribution followed by the token at time $t$ for two neighboring datasets $D \sim_u D'$.
Then, we can apply Proposition~\ref{prop:convexity} to $D_{\alpha}(Y_v||
Y'_v)$ with $c=1$, which is
ensured
when $\sigma \geq L\sqrt{2\alpha (\alpha - 1)}$, and we have:

\[ D_{\alpha}(Y_v|| Y'_v) \leq (1+1) \E_{t: i_t=i_0} D_{\alpha}(X_t || X'_t). \]

We can now bound $D_ {\alpha}(X_t || X'_t)$ for
each $t$
and obtain:
\[ \begin{array}{lll}
D_{\alpha}(Y_v|| Y'_v) & \leq & \sum_{t=1}^{T-t_1} P(i_t=i, i_{t-1} \ne i, \dots, i_1 \ne i |i_0=i) \frac{2 \alpha L^2}{\sigma^2 t} \\
    & \leq & \frac{2 \alpha L^2}{\sigma^2} \sum_{t=1}^{\infty} \frac{P(i_t=i, i_{t-1} \ne i, \dots, i_1 \ne i|i_0=i)}{t}\\
    & \leq & \frac{2 \alpha L^2}{\sigma^2}.
\end{array}
\]

Denote $T_u$ as the maximum number of contributions for user $u$. 
Using 
the
composition property of RDP, 
we can then prove that 
Algorithm~\ref{alg:meta} satisfies $(\alpha, \frac{4 T_u \alpha L^2}
{\sigma^2})$-Network Rényi DP, 
which can then be converted into an $(\eps_{\text{c}}, \delta_{\text{c}})$-DP
statement using
Proposition~\ref{prop:convert}. 
This proposition calls for minimizing the
function $\alpha \rightarrow \eps_{\text{c}}(\alpha)$
for $\alpha\in(1,\infty)$. However, recall that from our use of the weak convexity
property we have the additional constraint
on
$\alpha$ requiring that $\sigma
\geq L\sqrt{2 \alpha (\alpha - 1)}$. This creates two regimes: for
small $\eps_{\text{c}}$ (i.e, large $\sigma$ and small $T_u$), the minimum
is not reachable, so we take the best possible $\alpha$ within the interval,
whereas we have an optimal regime for larger $\eps_{\text{c}}$. This
minimization can be done numerically, but for simplicity of
exposition we can derive a suboptimal closed form which is the one given in
Theorem~\ref{thm:iteration}.

To obtain this closed form, we reuse 
Theorem 32 
of \cite{amp_iter}.
In
particular, for $q = \max \big(2 T_u, 2 \ln(1/\delta_{\text{c}}) \big)$, $\alpha = \frac{\sigma \sqrt{\ln(1/\delta_{\text{c}})} }{L 
\sqrt{q}}$ and $\eps_{\text{c}} = \frac{4L \sqrt{q \ln(1/\delta_{\text{c}})}}{\sigma}$, the
conditions $\sigma \geq L\sqrt{2 \alpha (\alpha - 1)}$ and $\alpha > 2$ are satisfied. Thus, we have a bound on the privacy loss which holds the two
regimes thanks to the definition of $q$.

Finally, we bound $T_u$ by $N_u = \frac{T}{n} + \sqrt{\frac{3T}{n} \log(1/
\hat{\delta})}$ with probability $1-\hat{\delta}$ as
done in the previous proofs for real summation and discrete histograms.
Setting $\eps'=\eps_{\text{c}}$ and $\delta'=\delta_{\text{c}} + \hat{\delta}$
concludes the proof.
\end{proof}

%



%
%
%

\end{document}